\documentclass[a4paper,fleqn]{cas-sc}

\usepackage[numbers]{natbib}

\usepackage{graphicx}%
\usepackage{multirow}%
\usepackage{amsmath,amssymb,amsfonts}%
\usepackage{amsthm}%
\usepackage{mathrsfs}%
\usepackage[title]{appendix}%
\usepackage{xcolor}%
\usepackage{textcomp}%
\usepackage{manyfoot}%
\usepackage{booktabs}%
\usepackage{algorithm}%
\usepackage{algorithmicx}%
\usepackage{algpseudocode}%
\usepackage{listings}%
\usepackage{makecell}%
\usepackage{adjustbox}%
\usepackage{float}%
\usepackage{rotating}%

\def\tsc#1{\csdef{#1}{\textsc{\lowercase{#1}}\xspace}}
\tsc{WGM}
\tsc{QE}
\tsc{EP}
\tsc{PMS}
\tsc{BEC}
\tsc{DE}

\begin{document}
\emergencystretch=3em 
\let\WriteBookmarks\relax
\hypersetup{hypertexnames=false}

\shorttitle{PR-Net for physically grounded extrapolation in PEMWE}

\shortauthors{Kim et~al.}

\title [mode = title]{Hard-constraint physics-residual networks for hydrogen crossover prediction and high-pressure extrapolation in PEM water electrolysis}

\author[1,2]{Yong-Woon Kim}
\ead{ywkim@jejunu.ac.kr}
\credit{Conceptualization, Methodology, Software, Data curation, Formal analysis, Visualization, Writing - original draft}
\affiliation[1]{organization={Department of Computer Engineering, Jeju National University},
    city={Jeju},
    postcode={63243},
    country={South Korea}}
\affiliation[2]{organization={Regional Leading Research Center (RLRC) for Green Hydrogen, Jeju National University},
    city={Jeju},
    postcode={63243},
    country={South Korea}}

\author[3]{Jihyeok Lee}
\credit{Visualization, Validation, Formal analysis, Writing - review \& editing}
\affiliation[3]{organization={Faculty of Applied Energy System, Jeju National University},
    city={Jeju},
    postcode={63243},
    country={South Korea}}

\author[4]{Chulung Kang}
\credit{Investigation, Resources, Data curation, Validation, Writing - review \& editing}
\affiliation[4]{organization={Department of Mechanical System Engineering, Jeju National University},
    city={Jeju},
    postcode={63243},
    country={South Korea}}

\author[5]{Yung-Cheol Byun}
\cormark[1]
\ead{ycb@jejunu.ac.kr}
\credit{Supervision, Project administration, Funding acquisition, Writing - review \& editing}
\affiliation[5]{organization={Department of Computer Engineering, Major of Electronic Engineering, Jeju National University},
    city={Jeju},
    postcode={63243},
    country={South Korea}}

\cortext[cor1]{Corresponding author}

\begin{abstract}
Hydrogen crossover is a critical safety and efficiency constraint in
high-pressure polymer electrolyte membrane water electrolysis (PEMWE),
yet accurate prediction remains challenging because available experimental data are
limited, transport physics are strongly coupled, and industrial
operation requires reliable extrapolation beyond observed conditions.
This study develops a hard-constraint physics-residual network
(PR-Net) for hydrogen crossover prediction in PEMWE and evaluates it
against a purely data-driven neural network (NN) and a soft-constraint
physics-informed neural network (PINN) under a common benchmark.
PR-Net embeds Henry's, Fick's, and Faraday's laws as a deterministic
backbone and learns only the residual correction required to capture
unmodelled non-linear effects. The benchmark comprises 184 observations
compiled from eight peer-reviewed sources across six membrane types,
covering 1--200~bar, 25--85$^{\circ}$C, and 0.05--5.0~A~cm$^{-2}$.
PR-Net achieves $R^2 = 99.57 \pm 0.16\%$, representing 9-fold lower prediction variability than the purely data-driven NN and
the soft-constraint PINN. In the pressure-axis extrapolation
protocol, PR-Net attains $R^2 = 94.02 \pm 0.92\%$ at 200~bar---2.5 times beyond
the training pressure range---compared
with $68.06 \pm 5.52\%$ for the soft-constraint PINN and
$58.00 \pm 8.60\%$ for the purely data-driven NN, all differences statistically significant ($p < 0.001$). Residual analysis is consistent with the learned correction
accounting for part of the high-pressure gas-phase non-ideality while the model
recovers a transport-regime transition near 0.23~A~cm$^{-2}$
between Fickian diffusion-dominated and Faradaic
production-dominated transport. With a computation time of
$1.08 \pm 0.34$~ms on low-power embedded hardware, PR-Net
provides a practical framework for real-time crossover monitoring,
adaptive process control, and safer high-pressure green-hydrogen
operation.
\end{abstract}

\begin{keywords}
Physics-residual network \sep Physics-informed neural network \sep
Hard-constraint learning \sep
Hydrogen crossover \sep PEM water electrolysis \sep
Uncertainty quantification
\end{keywords}

\begingroup
\sloppy
\hbadness=10000
\hfuzz=120pt
\maketitle
\endgroup

\section{Introduction}
\label{sec:introduction}

Polymer electrolyte membrane water electrolysis (PEMWE) is a promising
technology for large-scale green hydrogen production because it offers
rapid dynamic response, high current density, and seamless integration
with intermittent renewable energy sources, as detailed in the
comprehensive review by Carmo et al.~\cite{bib1} and the assessment
of current status and challenges by Grigoriev et al.\
\cite{bib2}; Shiva Kumar and Himabindu \cite{bib3} further
summarise PEM electrolysis as a route for clean hydrogen
production.
However, its industrial deployment remains constrained by the
cross-permeation of product gases through the polymer membrane.
Hydrogen crossover into the anodic oxygen stream lowers Faradaic
efficiency and creates a significant safety concern; operation near the
lower flammability limit of 4\% H$_2$ in O$_2$ increases the risk of
system failure, as quantified by Trinke et al.~\cite{bib4} for
current-density-dependent hydrogen permeation, by Schalenbach et al.\
\cite{bib6} for pressurised operation, and by Omrani and Shabani
\cite{bib31} for combined effects of current density, pressure,
temperature, and compression. As PEMWE systems move towards
high-pressure operation to reduce or eliminate
downstream mechanical compression, accurate monitoring and mitigation of
crossover under severe operating transients have become central
chemical-engineering challenges, as discussed by Schalenbach et al.\
\cite{bib6} for high-pressure crossover, by Feng et al.\
\cite{bib7} for degradation mechanisms, and by Schmidt et al.\
\cite{bib9} for future cost and performance projections.
Conventional prediction of hydrogen crossover relies primarily on
mechanistic formulations based on Fick's law of diffusion and Henry's
law of solubility. Representative examples include the two-dimensional
low-pressure electrolyser model by Aubras et al.~\cite{bib10}, the
electrochemical performance model by Han et al.~\cite{bib11}, and the
high-temperature thermal-electrochemical assessment by Toghyani
et al.~\cite{bib12}. Although these
analytical models provide essential thermodynamic insight, their use in
dynamic industrial settings remains limited. In particular, they often
struggle to parameterise highly non-linear and interacting physical
effects---including temperature-induced membrane swelling, localised
gradients, and time-dependent structural degradation---with sufficient
fidelity, as noted in the energy and exergy analysis of Ni et al.\
\cite{bib13}, the simple electrolyser model and experimental
validation by García-Valverde et al.~\cite{bib14}, and the dynamic
modelling and simulation study of Awasthi et al.~\cite{bib15}. In addition, the computational
cost of solving coupled multiphysics differential equations can hinder
deployment in real-time, millisecond-scale adaptive process control.

Data-driven and knowledge-integrated
machine-learning approaches have therefore been increasingly investigated for
PEMWE design, modelling, optimisation, and monitoring
For instance, Mohamed et al.~\cite{bib16} optimised PEM electrolyser
cell design via machine learning, Ding et al.~\cite{bib17} reviewed
machine-learning utilisation in PEM electrolyser development, and
Chen et al.~\cite{bib18} proposed a knowledge-integrated
machine-learning framework for PEMWE. Purely data-driven neural
network (NN) can
capture complex non-linear input-output relationships without explicit
mechanistic programming; however, experimental hydrogen crossover
measurements are inherently limited in volume and diversity, spanning
multiple membrane types and laboratory setups. With limited
experimental data, models trained without physical constraints risk
overfitting to dataset-specific patterns rather than the underlying
transport mechanisms, producing predictions that violate thermodynamic
bounds when applied outside the experimental pressure range---a
critical concern for safe operation at high-pressures. Physics-
informed neural networks (PINNs) address this limitation by embedding
governing equations directly into the training objective, promoting physically consistent predictions and improving
generalisation, as discussed in \cite{bib19} and \cite{bib20}. In the prevailing implementation, governing
equations enter the objective as soft-constraint penalty terms within a
composite loss function.

\paragraph{Related work.}
As summarised in Table~\ref{tab:literature_main}, recent PINN studies
in polymer electrolyte membrane (PEM)-related systems have focused on
temperature prediction in PEM water electrolysis \cite{bib52},
remaining useful life (RUL) estimation in polymer electrolyte membrane
fuel cells \cite{bib51}, and membrane degradation modelling in PEM
electrolysers \cite{bib53}. Across these studies,
physical knowledge is incorporated through \textit{soft-constraint}
formulations, in which governing equations enter the objective as
penalty terms within a composite loss. An expanded comparison is
provided in Supplementary Table~\ref{tab:s_literature}, which places
the representative PEM-related studies in a broader context. However, these prior studies collectively suggest that the specific combination of hydrogen crossover prediction, comparative physics-integration
benchmarking, and hard-constraint reliability
assessment has not yet been addressed in the PEMWE literature.

\begin{table}[htbp]
\caption{Representative PINN studies in PEM-related systems considered in this work. Columns list application domain, system type, constraint-integration strategy, and relevance to the present hydrogen-crossover prediction task. An expanded comparison across electrochemical and related systems is provided in Supplementary Table~\ref{tab:s_literature}.}
\label{tab:literature_main}
\small
\begin{tabular*}{\textwidth}{@{}p{2.2cm}p{1.8cm}p{2.5cm}p{5.4cm}p{2.9cm}@{}}
\toprule
\textbf{Application} & \textbf{System} & \textbf{Constraint integration} & \textbf{Relevance to this work} & \textbf{Reference} \\
\midrule
Temperature prediction & PEMWE & Soft-constraint PINN & PINN for PEM electrolysis, but not H$_2$ crossover or hard constraints & Zerrougui et al.~(2025)~\cite{bib52} \\
RUL estimation & PEM fuel cell & Soft-constraint PINN & PEM-related prognostics, but outside PEMWE crossover & Ko et al.~(2025)~\cite{bib51} \\
Membrane degradation & PEMWE & Soft-constraint PINN & Closest PEMWE PINN study, but on degradation rather than crossover and without hard constraints & Polo-Molina et al.~(2025)$^{a}$~\cite{bib53} \\
\bottomrule
\end{tabular*}
\begin{flushleft}
\footnotesize{$^{a}$~Preprint; not peer-reviewed at the time of writing.}
\end{flushleft}
\end{table}

Despite recent advances, three gaps remain in the literature reviewed
here (Table~\ref{tab:literature_main} and Supplementary
Table~\ref{tab:s_literature}). First, hydrogen crossover
concentration has not been established as a primary machine-learning
prediction target in PEMWE, despite being the key safety variable for
high-pressure operation near the 4\% H$_2$-in-O$_2$ flammability
limit. Second, no study has placed purely data-driven NNs,
soft-constraint PINNs, and hard-constraint physics-integrated models
within a single comparative framework for PEMWE crossover prediction
under a common dataset, validation protocol, and pressure-axis
extrapolation setting. Third, although decomposing predictions into a physics-based component and
a learned residual correction has been explored in other engineering
domains---for example, Chen and Xiu~\cite{chen2021} for generalised
residual networks on unknown dynamical systems and Jia et
al.~\cite{jia2021} for physics-guided learning of lake-temperature
profiles---this approach has not been applied to
electrochemical transport, and whether hard-constraint architectures
structurally improve reliability under
limited experimental data and safety-critical extrapolation conditions of
high-pressure PEMWE remains unverified.

To address these three gaps, we present, to the best of our knowledge,
the first systematic comparison of three physics-integration
strategies---a purely data-driven NN, a soft-constraint
PINN, and a hard-constraint physics-residual network (PR-Net)---for
predicting hydrogen crossover in high-pressure PEMWE. The central
engineering question is whether structurally embedding transport
physics into the model architecture improves prediction reliability
under the limited experimental data and high-pressure extrapolation
conditions characteristic of industrial PEMWE operation. Under this
controlled evaluation, the purely data-driven NN establishes a
no-physics reference, the soft-constraint PINN represents the
prevailing PEM-related approach, and PR-Net is assessed for whether
hard structural physics integration confers a measurable advantage in
safety-critical extrapolation. This framework provides a comparison
not previously available for PEMWE hydrogen-crossover prediction.

Rather than treating physical laws solely as soft regularisation
penalties, PR-Net embeds analytical transport equations---specifically
Henry's, Fick's, and Faraday's laws---as a deterministic computational
backbone. In soft-constraint PINN, governing equations enter the
training objective as penalty terms alongside the data-fidelity loss;
under limited experimental data and tightly coupled non-linear transport
equations, simultaneously minimising the data-fidelity loss and the
physics penalty can cause the model to trade physical consistency for
empirical fit, or vice versa, as discussed in \cite{bib20} and \cite{bib24}. These consequences
are well documented in the PINN literature and manifest as
gradient-direction conflicts between the data and physics objectives
during training, unstable or stalled convergence, sensitivity to the
choice of loss-weighting coefficients, and degraded generalisation
outside the training distribution, as reported in \cite{bib20}, \cite{bib24}, and \cite{Wang2021}.
Dynamic re-weighting schemes---such as neural-tangent-kernel--style
adaptive weighting~\cite{wang2022when} and PCGrad gradient
projection~\cite{Yu2020}---have been proposed to mitigate these
issues, yet soft-constraint formulations retain post-hoc balancing
as their underlying mechanism rather than removing the source of the
conflict. This trade-off is especially pertinent in PEMWE
hydrogen-crossover prediction, where the physics backbone comprises a
coupled system of Henry's, Fick's, and Faraday's equations rather
than a simple analytical prior, and where the extrapolation
requirement demands physical consistency at pressures not covered by
the experimental training range. PR-Net mitigates this trade-off
by assigning known transport structure to the deterministic backbone
and restricting the neural network to modelling systematic residuals,
thereby constraining the model to physically admissible predictions.
To evaluate this design under heterogeneous experimental conditions,
data were consolidated from multiple independent peer-reviewed
studies spanning diverse membrane types and a wide operating range
(Section~\ref{sec:data_compilation}), exposing all competing
architectures to the experimental variability and limited data
availability characteristic of real PEMWE operation.

The proposed PR-Net framework makes five principal contributions to
electrochemical process safety and physics-integrated predictive modelling:

\begin{enumerate}

  \item \textbf{First Systematic Benchmark for PEMWE Hydrogen-Crossover Prediction.}
  This study establishes, to the best of our knowledge, the first
  systematic benchmark for hydrogen-crossover prediction in PEMWE that compares purely data-driven, soft-constraint, and
  hard-constraint physics-integrated approaches under a common dataset,
  validation protocol, and pressure-axis extrapolation setting. The
  benchmark isolates the effect of physics-integration strategy
  and provides quantitative performance tiers
  across these strategies in terms of prediction stability,
  pressure-axis extrapolation, and cross-source transferability.

  \item \textbf{Stable and Physics-Consistent Prediction Under Limited Experimental Data.}
  By embedding Henry's, Fick's, and Faraday's laws as a deterministic
  backbone and learning only the residual component, PR-Net mitigates
  the optimisation trade-off inherent in soft-constraint PINNs. This
  hard-constraint design reduces prediction variability
  by 9-fold relative to the soft-constraint PINN and purely
  data-driven NN while improving convergence stability; near-zero
  prediction residuals ($\mu = 0.0131$\%) indicate that the physics
  backbone limits systematic prediction bias. These
  findings provide quantitative evidence that hard-constraint
  physics embedding achieves stable, low-bias predictions,
  outperforming soft-constraint approaches under the data-limited
  conditions characteristic of PEMWE research.

  \item \textbf{Pressure-Axis Extrapolation Performance.}
  Under Protocol-IEP (Isolated Extrapolation Protocol)---a zero-data-
  leakage setting in which backbone parameters are calibrated
  exclusively from training-domain data---PR-Net achieves
  $R^2 = 94.02 \pm 0.92$\% at 200~bar, corresponding to a 2.5-fold
  extension beyond the training pressure range, compared with
  $68.06 \pm 5.52$\% for the soft-constraint PINN and
  $58.00 \pm 8.60$\% for the purely data-driven NN (all differences
  $p < 0.001$). These results provide evidence that
  hard-constraint physics integration substantially outperforms
  soft-constraint and purely data-driven approaches beyond the
  training pressure range, maintaining higher prediction accuracy where
  alternative approaches show substantial performance degradation.
  Apparatus-bias bounding analyses are presented in Section~\ref{sec:apparatus_bias_caveat}.

  \item \textbf{Residual-Based Mechanistic Insight.}
  Because PR-Net explicitly separates the analytical backbone from the
  learned correction, the residual term provides a transparent window
  into the limitations of conventional electrochemical transport models.
  Residual analysis identifies a near-zero industrial-band residual
  at 80--85$^{\circ}$C (mean $\approx -0.17$\%p, $n=84$)
  and a Fickian-to-Faradaic transport transition near
  0.23~A~cm$^{-2}$, without requiring explicit mechanistic
  programming. These findings establish a residual-based mechanistic
  approach in which structured machine-learning residuals directly
  identify gaps in theoretical electrochemical transport models and
  guide their refinement.

  \item \textbf{Industrial Edge Deployment.}
  Because PR-Net combines a non-iterative analytical physics backbone
  with a lightweight residual network (34,305 parameters), the
  architecture achieves $1.08 \pm 0.34$~ms computation time on low-power
  embedded hardware---a speedup of 5--6 orders of magnitude
  over representative computational fluid dynamics (CFD) runtimes.
  This computational profile satisfies the ISO~26142 $t_{90} \leq
  30$~s response-time requirement for real-time safety monitoring
  and adaptive process control in large-scale hydrogen
  facilities~\cite{iso26142}. This result shows that physics-integrated PEMWE
  hydrogen-crossover prediction can satisfy the ISO~26142
  response-time requirement on constrained edge hardware,
  supporting deployment without centralised computational
  infrastructure.
\end{enumerate}

\begin{figure}
\centering
\includegraphics[width=\textwidth,keepaspectratio]{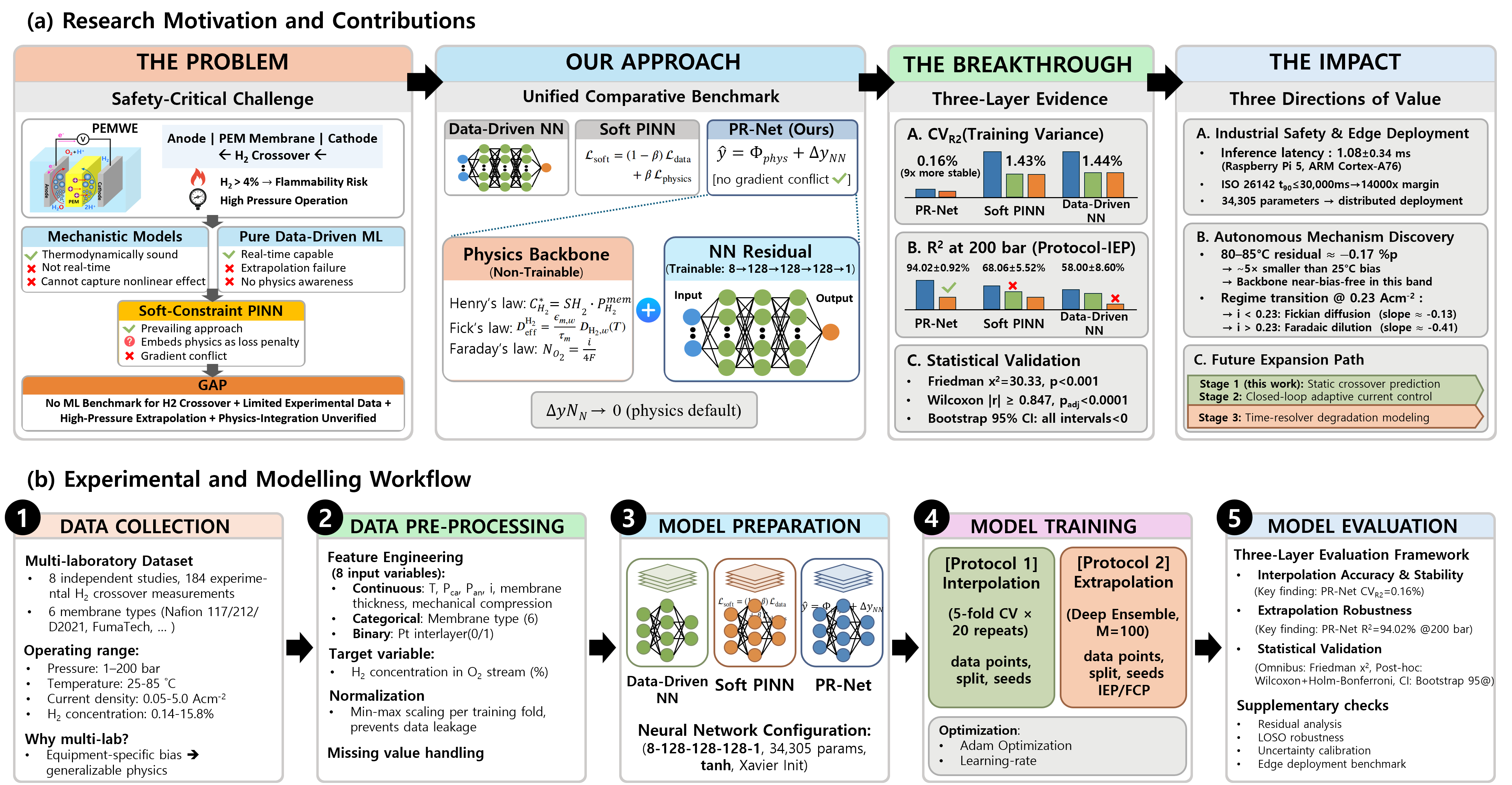}
\caption{Overview of the study design and main findings. The schematic summarises the PEMWE hydrogen-crossover problem, the comparative benchmark across the data-driven NN, soft-constraint PINN, and PR-Net, and the corresponding evaluation and deployment workflow.}
\label{fig:study_overview}
\end{figure}

Fig.~\ref{fig:study_overview} summarises the overall study design,
including the hydrogen-crossover problem setting, the comparative
benchmark across the three physics-integration regimes, and the
associated evaluation and deployment workflow.

\section{Theory and Methodology}
\label{sec:methodology}

The core innovation of PR-Net lies in
its deterministic phenomenological backbone, which tightly constrains
the neural network's hypothesis space. Unlike purely data-driven models
that must implicitly infer thermodynamic laws from limited experimental data, our
approach explicitly embeds the foundational transport mechanisms---gas
solubility, structural swelling, effective diffusion, and
electrochemical production---as a non-trainable computational
foundation.

\subsection{Phenomenological Modelling of Hydrogen Crossover}
\label{subsec:physics_model}

The theoretical hydrogen crossover concentration
$\Phi_{\mathrm{H_2}}^{\mathrm{phys}}$ is calculated by solving a
coupled system of constitutive equations governing mass transport
within the polymer electrolyte membrane (PEM) and the liquid/gas
diffusion layer (LGDL), following \cite{bib6} and \cite{bib4}. This model serves as the
physical anchor for the subsequent residual learning.

\subsubsection{Gas Solubility and Compression-Adjusted Pressure}

The dissolved hydrogen concentration at the membrane--catalyst
interface ($C_{\mathrm{H_2}}^*$) is determined by Henry's law, which
links the interfacial partial pressure to the temperature-dependent
solubility:
\begin{equation}
C_{\mathrm{H_2}}^* = S_{\mathrm{H_2}}(T)\,P_{\mathrm{H_2}}^{\mathrm{mem}}
\label{eq:henrys_law}
\end{equation}
where the solubility $S_{\mathrm{H_2}}(T)$ is derived from the
empirical Bunsen coefficient $\beta_{\mathrm{H_2}}$:
\begin{equation}
\beta_{\mathrm{H_2}} = \exp\!\left(
  -39.9611 + \frac{53.9381}{T/100}
  + 16.3135\ln\!\left(\frac{T}{100}\right)
\right)
\label{eq:bunsen}
\end{equation}

The effective hydrogen pressure at the membrane interface
($P_{\mathrm{H_2}}^{\mathrm{mem}}$) must account for the mechanical
compression of the LGDL during cell assembly. The compressed porosity
($\phi_{\mathrm{comp}}$) and permeability ($K_{\mathrm{LGDL}}$) are
corrected using a cubic relation with porosity:
\begin{equation}
\phi_{\mathrm{comp}} =
  1 - \frac{t^{\mathrm{initial}}}{t^{\mathrm{comp}}}
  \left(1 - \phi_{\mathrm{initial}}\right)
\label{eq:porosity_comp}
\end{equation}
\begin{equation}
K_{\mathrm{LGDL}} = K_{\mathrm{initial}}
  \cdot \left(\frac{\phi_{\mathrm{comp}}}{\phi_{\mathrm{initial}}}\right)^3
\label{eq:permeability_comp}
\end{equation}
Incorporating these structural modifications, the effective interfacial
pressure is determined as a function of cathode pressure ($P_{\mathrm{ca}}$)
and current density ($i$) via the compression-adjusted transport
parameter $K_D$:
\begin{equation}
P_{\mathrm{H_2}}^{\mathrm{mem}} =
  \sqrt{P_{\mathrm{ca}}^2 + K_D \cdot i}
\label{eq:effective_pressure}
\end{equation}

\subsubsection{Effective Diffusion and Membrane Swelling Dynamics}

Diffusive transport through the PEM is governed by Fick's law.
To accurately capture transport dynamics across diverse membrane types,
the effective diffusion coefficient ($D_{\mathrm{eff}}^{\mathrm{H_2}}$)
is computed dynamically by accounting for membrane swelling. The water
volume fraction ($\epsilon_{m,w}$) and tortuosity ($\tau_m$) are
explicitly calculated from the membrane's specific water content
($\lambda_m$) and molar volumes ($V_w$, $V_m$):
\begin{equation}
\epsilon_{m,w} =
  \frac{\lambda_m V_w}{V_m + \lambda_m V_w}
\label{eq:water_volume}
\end{equation}
\begin{equation}
\tau_m = \left(1 - 0.85\left(1 - \epsilon_{m,w}\right)\right)^{-1}
\label{eq:tortuosity}
\end{equation}
The effective diffusion coefficient is then formulated by adjusting
the Arrhenius-type self-diffusion coefficient of hydrogen in water
($D_{\mathrm{H_2},w}(T)$) with these structural parameters:
\begin{equation}
D_{\mathrm{H_2},w}(T) =
  7.734 \times 10^{-6}
  \cdot \exp\!\left(-\frac{2225.4}{T}\right)
  \quad [\mathrm{m^2\,s^{-1}}]
\label{eq:self_diffusion}
\end{equation}
\begin{equation}
D_{\mathrm{eff}}^{\mathrm{H_2}} =
  \frac{\epsilon_{m,w}}{\tau_m}\,D_{\mathrm{H_2},w}(T)
\label{eq:effective_diffusion}
\end{equation}

\subsubsection{Combined Mass Transport and Final Crossover Prediction}

To address high-pressure and high-current-density operation, the
total crossover flux ($N_{\mathrm{H_2}}^{\mathrm{co}}$) integrates
both diffusive and convective/migration contributions.

The membrane-specific mass transport coefficient
($k_\mathrm{MT} = \alpha \cdot i^{\beta}$) employs
power-law relationships whose four scalar coefficients
$(a_{\alpha}, b_{\alpha}, a_{\beta}, b_{\beta})$ are fitted
per membrane by minimising the mean-squared error
between $\Phi_{\mathrm{H_2}}^{\mathrm{phys}}$ and the
measured hydrogen concentration over the calibration
subset. Two calibration protocols are
employed in this study, as detailed in
Section~\ref{sec:training_validation}: (1)~a \textit{Isolated Extrapolation Protocol}
(IEP) in which parameters are estimated exclusively
from the extrapolation training subset ($\leq$80~bar,
$n = 42$) to avoid data leakage; and (2)~a
\textit{Full-data Calibration Protocol} (FCP) utilising all $n = 184$
compiled observations, representing a
industrial scenario in which a globally pre-calibrated
physics model is subsequently deployed with a locally
trained neural residual. Unless otherwise stated, all
extrapolation analyses in this study use
Protocol~(1):

\begin{equation}
\alpha = a_{\alpha} \cdot P^{b_{\alpha}}
  \cdot \left(1 + 0.005\left(T - 60\right)\right)
\label{eq:alpha_param}
\end{equation}
\begin{equation}
\beta = a_{\beta} + b_{\beta} \cdot \ln(P)
  \cdot \left(1 + 0.003\left(T - 60\right)\right)
\label{eq:beta_param}
\end{equation}

The two protocols (IEP and FCP) share an identical
calibration procedure and differ only in the data
subset used to fit the per-membrane parameters
$(a_{\alpha}, b_{\alpha}, a_{\beta}, b_{\beta})_k$.
Because the residual surface of
$\Phi_{\mathrm{H_2}}^{\mathrm{phys}}$ with respect to
these four coefficients is non-convex---the four
coefficients enter through a power law in $P$, a
logarithm in $P$, and a Sherwood-type rational form
that couples $k_{\mathrm{MT}}$ to the
diffusion-convection denominator in
Eq.~\ref{eq:total_flux}---a gradient-based non-linear
least-squares solver is sensitive to initial values
and prone to terminating in shallow local minima. We
therefore adopt a derivative-free global
optimiser, SciPy's \texttt{differential\_evolution}
\cite{storn1997differential}, with box constraints
chosen so that the power-law
$\alpha = a_{\alpha} P^{b_{\alpha}}$ remains positive
and monotonically decreasing in pressure and the
logarithmic form $\beta = a_{\beta} + b_{\beta}\ln P$
stays in the empirically observed range across
membranes. A pooled-fit seed obtained from a
cross-membrane fit is injected into the initial
population and is also retained as the inference-time
fallback (\texttt{alpha\_common},
\texttt{beta\_common}) for any membrane type absent
from the calibration subset.

The two protocols then differ only in the calibration
subset: IEP fits the four-parameter set on the
$\le 80$~bar training partition for each membrane $k$
($n=42$ aggregated across membranes), whereas FCP fits
on the full $n=184$ dataset. All reported PR-Net
extrapolation results use IEP-calibrated backbones to
preserve a strict train/test boundary, and FCP-
calibrated backbones are reported only for
completeness under the industrial pre-calibration
scenario. Exact box-constraint values, population
size, convergence tolerance, generation budget, random
seed, and the differential-evolution pseudocode are
documented in Supplementary
Section~\ref{suppl:de_calibration_details}
(Algorithm~\ref{alg:de_calibration}).

The combined crossover flux across the total diffusion path length
($t_{\mathrm{dif}}$), encompassing the membrane and catalyst layers,
is given by:
\begin{equation}
N_{\mathrm{H_2}}^{\mathrm{co}} =
  \frac{D_{\mathrm{eff}}^{\mathrm{H_2}} \cdot C_{\mathrm{H_2}}^*
        + \dfrac{i}{2F}}
       {1 + k_{\mathrm{MT}} \cdot
        \dfrac{t_{\mathrm{dif}}}{D_{\mathrm{eff}}^{\mathrm{H_2}}}}
\label{eq:total_flux}
\end{equation}
Finally, the theoretical hydrogen concentration in the oxygen stream
($\Phi_{\mathrm{H_2}}^{\mathrm{phys}}$) is derived from the ratio of
the crossover flux to the total gas production rate, the latter
predicted via Faraday's law:
\begin{equation}
N_{O_2} = \frac{i}{4F}
\label{eq:faradays_law}
\end{equation}
\begin{equation}
\Phi_{\mathrm{H_2}}^{\mathrm{phys}} =
  \frac{N_{\mathrm{H_2}}^{\mathrm{co}}}
       {N_{O_2} + N_{\mathrm{H_2}}^{\mathrm{co}}}
  \times 100 \quad [\%]
\label{eq:physics_backbone_final}
\end{equation}
This derived $\Phi_{\mathrm{H_2}}^{\mathrm{phys}}$
provides a thermodynamically grounded reference across the full operational
domain of 0.05--5.0~A~cm$^{-2}$, 1--200~bar, and 25--85$^{\circ}$C,
providing a physically grounded anchor that helps limit unphysical
extrapolation by the neural network residual component.

\subsection{Hard-Constraint PR-Net Architecture}
\label{sec:prnet}

In complex electrochemical systems such as PEMWE, purely data-driven
models may not reliably satisfy fundamental thermodynamic boundaries under extrapolation. Conversely, idealised
mechanistic models struggle to accurately capture non-linear behaviours
such as temperature-induced membrane swelling and variable tortuosity
under high-pressure operation. To bridge this gap, we propose the
hard-constraint PR-Net, which structurally
embeds the phenomenological transport equations derived in
Section~\ref{subsec:physics_model} as a deterministic, non-trainable
computational backbone.

Unlike conventional machine-learning approaches that map operating
conditions directly to the target variable, the PR-Net decomposes the
prediction task into two complementary components:
\begin{equation}
\hat{y}_{\mathrm{final}}(\mathbf{x}) =
  \Phi_{\mathrm{H_2}}^{\mathrm{phys}}(\mathbf{x}) +
  \mathrm{NN}_{\mathrm{residual}}(\mathbf{x})
\label{eq:prnet_formulation}
\end{equation}
where $\Phi_{\mathrm{H_2}}^{\mathrm{phys}}(\mathbf{x})$ represents
the theoretical hydrogen crossover concentration computed analytically
from Henry's law, Fick's diffusion, and Faraday's law
(Section~\ref{subsec:physics_model}). The neural network component
$\mathrm{NN}_{\mathrm{residual}}(\mathbf{x})$ is strictly tasked with
learning the systematic discrepancy between theoretical predictions
and experimental observations.

\paragraph{Network architecture}
The residual network is implemented as a fully connected
feedforward neural network (FFNN) with a \texttt{feature\_extractor}
module comprising three hidden layers (the first projecting
from input to hidden dimension, followed by two same-dimension
hidden layers), and a separate \texttt{output\_layer}:
\begin{equation*}
  \underbrace{
    \mathrm{Linear}(8,128)\text{--}\mathrm{Tanh}
    \text{--}\bigl[\mathrm{Linear}(128,128)
    \text{--}\mathrm{Tanh}\bigr]{\times 2}
  }_{\texttt{feature\_extractor}}
  \;\longrightarrow\;
  \underbrace{
    \mathrm{Linear}(128,1)
  }_{\texttt{output\_layer}}
\end{equation*}
yielding the architecture $8\text{--}128\text{--}128
\text{--}128\text{--}1$ with 34,305 total trainable
parameters. The eight input neurons correspond to the normalised operating
parameters: temperature, cathode pressure, anode pressure, membrane
thickness, current density, membrane type (one-hot encoded, 6
classes), mechanical compression, and platinum interlayer position (binary; 0 = absent, 1 = present).
The tanh activation function is used throughout, providing smooth
gradients essential for physics-constraint enforcement, following
\cite{bib20} and \cite{bib52}. Network weights are initialised using Xavier
uniform initialisation \cite{bib47}:
\begin{equation}
W \sim \mathcal{U}\!\left(
  -\sqrt{\frac{6}{n_{\mathrm{in}} + n_{\mathrm{out}}}},\;
   \sqrt{\frac{6}{n_{\mathrm{in}} + n_{\mathrm{out}}}}
\right)
\label{eq:xavier}
\end{equation}
where $n_{\mathrm{in}}$ and $n_{\mathrm{out}}$ denote the number of
input and output neurons for each layer; biases are initialised to
zero.

\paragraph{Loss formulation.}
The loss function for the PR-Net balances prediction accuracy with
residual magnitude regularisation:
\begin{equation}
\mathcal{L}_{\mathrm{total}} =
  \mathcal{L}_{\mathrm{data}} +
  \lambda \cdot \mathcal{L}_{\mathrm{residual}}
\label{eq:prnet_loss}
\end{equation}
The data loss minimises the mean squared error between the combined
output and experimental measurements:
\begin{equation}
\mathcal{L}_{\mathrm{data}} = \frac{1}{N} \sum_{k=1}^{N}
  \left(
    \Phi_{\mathrm{H_2}}^{\mathrm{phys}}(\mathbf{x}^{(k)}) +
    \mathrm{NN}_{\mathrm{residual}}(\mathbf{x}^{(k)}) -
    y_{\mathrm{true}}^{(k)}
  \right)^2
\label{eq:prnet_ldata}
\end{equation}
The residual loss provides explicit $L_2$ regularisation on the
magnitude of the learned corrections:
\begin{equation}
\mathcal{L}_{\mathrm{residual}} = \frac{1}{N} \sum_{k=1}^{N}
  \left( \mathrm{NN}_{\mathrm{residual}}(\mathbf{x}^{(k)}) \right)^2
\label{eq:prnet_lres}
\end{equation}
The regularisation weight $\lambda = 2.0$ was selected by a
controlled empirical sweep over
$\lambda \in \{0.1,\,0.5,\,1.0,\,2.0,\,5.0,\,10.0,\,20.0\}$
under the two validation protocols described in
Section~\ref{sec:training_validation}: stratified 5-fold
cross-validation with 20 repeats for interpolation accuracy
(100 trained models per $\lambda$, fold-level CV $R^2$), and a
$M = 100$ Deep Ensemble on the 120--200~bar test set for
extrapolation accuracy under Protocol-IEP. The selected
operating point sits at the elbow of the
interpolation--extrapolation trade-off and remains empirically
robust across the neighbourhood $[1.0,\,5.0]$, so the
selection is not the artefact of a sharply tuned weight. By
design, the residual is shrunk---rather
than zeroed---towards the physics anchor, so in extrapolation
regimes lacking empirical support the model degrades
gracefully towards the thermodynamically consistent backbone
prediction. The complete swept range, including the
small-$\lambda$ over-fitting regime (down to $\lambda = 0.1$)
and the large-$\lambda$ asymptotic regime (up to $\lambda =
20$), is documented in Supplementary
Section~\ref{suppl:computational_framework}
(Table~\ref{tab:exp4_lambda},
Fig.~\ref{fig:exp4_lambda_tradeoff}).

\paragraph{Gradient flow analysis.}
During backpropagation, the total gradient with respect to network
parameters $\theta$ is:
\begin{equation}
\frac{\partial \mathcal{L}_{\mathrm{total}}}{\partial \theta} =
  \frac{2}{N} \sum_{k=1}^{N}
  \left[
    \left(\hat{y}_{\mathrm{final}}^{(k)} - y_{\mathrm{true}}^{(k)}\right)
    + \lambda \cdot \mathrm{NN}_{\mathrm{residual}}(\mathbf{x}^{(k)})
  \right]
  \cdot
  \frac{\partial \mathrm{NN}_{\mathrm{residual}}(\mathbf{x}^{(k)})}
       {\partial \theta}
\label{eq:gradient_flow}
\end{equation}
Both gradient terms flow exclusively through
$\mathrm{NN}_{\mathrm{residual}}$, while
$\Phi_{\mathrm{H_2}}^{\mathrm{phys}}$ carries no learnable parameters
and does not participate in gradient-based optimisation. The physics
model acts as a fixed computational anchor, contributing to
predictions without introducing any gradient conflict.

\subsection{Baseline architectures: Soft-Constraint PINN and
Purely Data-Driven Neural Network}
\label{sec:baselines}

To evaluate the efficacy of hard-constraint
integration, two conventional modelling paradigms are
established as baselines. Both baselines share an
\textbf{identical} feedforward neural network (FFNN)
backbone with the PR-Net residual branch:
$8\text{--}128\text{--}128\text{--}128\text{--}1$,
tanh activation, 34,305 parameters, and Xavier uniform
initialisation. This controlled architecture ensures
that any observed performance differences are attributable
solely to the physics-integration strategy rather than
differences in model capacity or initialisation.

\paragraph{Rationale for baseline selection.}
The model set was selected to represent three distinct regimes of
physics integration: none (purely data-driven NN), soft
penalty-based integration (soft-constraint PINN), and hard structural
integration (PR-Net). This taxonomy makes the comparison
hypothesis-driven rather than exhaustive: removing any one of the
three would eliminate either the no-physics lower bound, the current
soft-constraint reference paradigm---which, as shown in
Table~\ref{tab:literature_main} and Supplementary
Table~\ref{tab:s_literature}, represents the integration strategy
adopted across prior PEM-related PINN studies---or the proposed
hard-constraint formulation itself. The selected models therefore
isolate the structural effect of physics integration on interpolation
stability, pressure-axis extrapolation, and uncertainty behaviour
under a common architecture and training protocol.

This interpretation should be read with an important scope condition.
Supplementary Table~\ref{tab:s_literature} also documents that the
broader PEMWE machine-learning literature includes many non-PINN model
families and architecture variants. Accordingly, the purpose of the
present benchmark is not to claim that tree-based methods, support
vector models, recurrent networks, or alternative neural architectures
are universally inferior for all PEMWE tasks. Rather, the benchmark is
deliberately restricted to the architecture-controlled comparison most
relevant to this study's central question: whether the \emph{mode of
physics integration}---none, soft penalty, or hard structural
embedding---changes stability and extrapolation behaviour for hydrogen
crossover prediction.

Alternative machine-learning families were not adopted as primary
comparators because they are less aligned with the present research
question than this physics-integration taxonomy. Tree-based methods
can be effective interpolation baselines but are not designed around
explicit physics integration and are less suitable for controlled
analysis of extrapolation to pressures beyond the training range.
Gaussian processes are attractive for uncertainty estimation but are
less compatible with the present edge-deployment objective and the
structural comparison of physics-integration mechanisms, especially as
dataset size and ensemble-style evaluation expand. Recurrent models
such as LSTMs are not well matched to the cross-sectional steady-state
dataset used here, because they are primarily designed for sequential
time-series dynamics rather than source-aggregated operating-state
snapshots.

\paragraph{Purely data-driven NN.}
The purely data-driven NN is implemented as
a fully connected FFNN with structure
$8\text{--}128\text{--}128\text{--}128\text{--}1$
(tanh, Xavier uniform; 34,305 parameters), built as a
single \texttt{nn.Sequential} module. It maps the input
state space directly to hydrogen concentration without
any physical priors:
\begin{equation}
\hat{y}_{\mathrm{pure}} = \mathrm{NN}_{\mathrm{direct}}(\mathbf{x})
\label{eq:pure_nn}
\end{equation}
While data-driven NNs possess universal approximation capability within the
interpolation domain, they inherently lack thermodynamic awareness,
which can yield predictions that are difficult to reconcile with known
physical trends (e.g., saturation behaviour inconsistent with Henry's
law) when extrapolating beyond limited experimental conditions.

\paragraph{Soft-constraint PINN.}
The soft-constraint PINN uses the same FFNN backbone
($8\text{--}128\text{--}128\text{--}128\text{--}1$,
tanh, Xavier uniform; 34,305 parameters) implemented
as a single \texttt{nn.Sequential} module and introduces
physical knowledge through a composite loss function:
\begin{equation}
\mathcal{L}_{\mathrm{soft}} =
  (1 - \beta)\,\mathcal{L}_{\mathrm{data}} +
  \beta\,\mathcal{L}_{\mathrm{physics}}
\label{eq:soft_loss}
\end{equation}
where $\beta$ follows an adaptive linear decay schedule
($0.7 \rightarrow 0.01$ over training), balancing empirical data
fidelity and physical compliance. The physics loss penalises
deviations of the direct network prediction from the analytical
crossover model:
\begin{equation}
\mathcal{L}_{\mathrm{physics}} = \frac{1}{N} \sum_{k=1}^{N}
  \left(
    \hat{y}_{\mathrm{NN}}^{(k)} -
    \Phi_{\mathrm{H_2}}^{\mathrm{phys}}(\mathbf{x}^{(k)})
  \right)^2
\label{eq:soft_lphys}
\end{equation}
From a chemical engineering optimisation perspective, this formulation
can create a gradient trade-off between data fitting and physics
regularisation. Because phenomenological models generally contain
idealised assumptions (e.g., simplified Arrhenius temperature
dependencies), the backbone $\Phi_{\mathrm{H_2}}^{\mathrm{phys}}$ may
systematically deviate from empirical data at low temperatures and
high pressures, as discussed in Section~\ref{subsec:physics_model}.
Consequently, simultaneously minimising $\mathcal{L}_{\mathrm{data}}$
and $\mathcal{L}_{\mathrm{physics}}$ may produce competing
optimisation directions: reducing the physics loss can limit fitting
of complex empirical phenomena, whereas over-prioritising data
fidelity can reduce adherence to physical transport laws. In our
evaluation, this trade-off is associated with less stable convergence,
higher training variance, and weaker extrapolation performance, as
quantified in Section~\ref{sec:optimization_resolution}.

\subsection{Experimental Data Compilation and Operational Domain}
\label{sec:data_compilation}

To evaluate the proposed PR-Net across heterogeneous PEM water
electrolysis conditions, we did not rely on single-laboratory
experimental datasets. Single-setup measurements are inherently prone
to systematic equipment biases and localised measurement artefacts,
which can artificially inflate a machine-learning model's apparent
accuracy while severely degrading its generalisation capability in
real-world deployment. Instead, we compiled a comprehensive and
heterogeneous experimental dataset comprising 184 independent hydrogen
crossover measurements extracted from eight distinct peer-reviewed
studies. These sources comprise high-pressure PEM electrolysis
datasets reported by Grigoriev et al.~\cite{bib32}, Schalenbach
et al.~\cite{bib6}, Bernt et al.~\cite{bib38}, and Wu et al.\
\cite{bib39}; current-density and porous-transport-layer
compression effects characterised by Trinke et al.~\cite{bib4}
and Stähler et al.~\cite{bib34}; and Pt-based
catalyst- and membrane-modification studies by Garbe et al.\
\cite{bib37} and Martin et al.~\cite{bib36}.

By consolidating data across multiple independent experimental setups,
the dataset spans inter-laboratory variance and operational
variability. This compilation allows the modelling architectures to be
evaluated under a broader range of conditions and reduces reliance on
laboratory-specific measurement patterns.

The compiled operational domain spans a broad thermodynamic and
electrochemical spectrum representative of both current commercial
electrolysers and next-generation high-pressure systems. As detailed
in Table~\ref{tab:operational_domain}, the dataset encompasses
operating temperatures of 25--85~$^{\circ}$C, cathode pressures
ranging from ambient conditions up to extreme high-pressure regimes
of 200~bar, and current densities from 0.05 to
5.0~A~cm$^{-2}$.

\begin{table}[htbp]
\caption{Membrane-specific operating conditions and structural
parameters in the consolidated experimental dataset ($n = 184$).}
\label{tab:operational_domain}
\centering
\resizebox{\textwidth}{!}{%
\begin{tabular}{lccccc}
\toprule
\textbf{Membrane Type}
  & \textbf{Temperature}
  & \textbf{Pressure Range}
  & \textbf{Current Density}
  & \textbf{Thickness}
  & \textbf{Data Points} \\
 & ($^{\circ}$C) & (bar) & (A~cm$^{-2}$) & ($\mu$m) & ($n$) \\
\midrule
Nafion\textsuperscript{\textregistered} 117
  & 25, 80, 85 & 1--200 & 0.07--1.80 & 209 & 66 \\
Nafion\textsuperscript{\textregistered} 212
  & 60, 80     & 1--10  & 0.17--2.98 & 58  & 32 \\
Nafion\textsuperscript{\textregistered} D2021
  & 80         & 1--10  & 0.50--3.50 & 110 & 21 \\
FumaTech E-730
  & 60         & 1--21  & 0.05--1.04 & 260 & 28 \\
Nafion\textsuperscript{\textregistered} 117 (178~$\mu$m)
  & 80         & 10--30 & 0.20--5.00 & 178 & 20 \\
Nafion\textsuperscript{\textregistered} 212 (51~$\mu$m)
  & 80         & 10--30 & 0.50--5.00 & 51  & 17 \\
\bottomrule
\end{tabular}%
}
\end{table}

The dataset incorporates six distinct membrane categories,
spanning diverse thicknesses (51--260~$\mu$m) and different ionomer
structures, including standard perfluorosulfonic acid (PFSA)-based Nafion variants and
FumaTech membranes. This structural diversity is essential for
validating the physical backbone's ability to accurately compute
effective diffusivity via membrane-specific water volume fractions
($\epsilon_{m,w}$) and tortuosity factors ($\tau_m$) under varying
hydration states.

For each experimental observation, the target variable was the
measured hydrogen concentration in the oxygen stream (\%, ranging
from 0.14\% to 15.8\%).

The feature space mapped to the PR-Net inputs consists of eight
parameters explicitly linked to phenomenological transport
mechanisms: temperature, cathode pressure, anode pressure
(assumed 1~bar when not explicitly reported, consistent with
standard differential-pressure operation \cite{bib1}),
nominal membrane thickness, operational current density,
categorical membrane type (one-hot encoded, 6 classes), mechanical
compression of the porous transport layer, and platinum interlayer
position (binary; 0 = absent for all 184 observations, retained for
forward compatibility).

\subsection{Model Training, Ensembling, and Statistical Validation
Protocols}
\label{sec:training_validation}

To standardise optimisation across the diverse electrochemical
operating space, all modelling
architectures were optimised using the Adam optimiser \cite{bib48}
with a batch
size of 32. Model-specific learning-rate optima were selected from a
shared search grid: $1.5 \times 10^{-3}$ for PR-Net, and
$2.5 \times 10^{-3}$ for both the soft-constraint PINN and the
purely data-driven NN. To prevent overfitting to the limited experimental dataset,
an early stopping mechanism was enforced with a patience of 250 epochs
and a minimum delta of $10^{-6}$ (maximum 700 epochs). For the
baseline soft-constraint PINN, the physics regularisation weight ($\beta$)
followed the selected initial-to-final linear decay schedule
$(0.7 \rightarrow 0.01)$ over training to balance physical guidance and data fidelity. In contrast,
the proposed PR-Net maintained the selected fixed residual regularisation weight
($\lambda = 2.0$), so that the deterministic physics backbone
consistently anchored the predictions. The full hyperparameter grid, the soft-constraint PINN
physics-weight sensitivity check, and hardware-specific computational
benchmarking are detailed in Supplementary Tables~\ref{tab:s_hyperparams_search},
\ref{tab:s_soft_pinn_beta_sensitivity}, and~\ref{tab:s_hardware}.

Because chemical process safety requires uncertainty
quantification---especially when operating near the explosive
threshold of 4\% H$_2$ in O$_2$---we implemented two
validation frameworks to assess interpolation (prediction accuracy within the experimental operating range) stability and
extrapolation reliability:

\begin{enumerate}

  \item \textbf{Interpolation and Stochastic Stability.}
  To evaluate predictive accuracy and training stability within the
  full experimental domain, a stratified 5-fold cross-validation
  scheme was employed, with stratification by membrane type to ensure
  balanced representation of all six membrane categories across folds.
  All 184 experimental points were used for training (stratified
  80/20 per fold). To account for stochastic initialisation effects,
  each fold was repeated with 20 independent runs using sequential
  random seeds ($42 + \mathrm{rep\_idx} \times 5$ per repetition),
  yielding 100 trained model instances per architecture. This
  repeated sampling was used to estimate both mean
  performance and training variance. Scaling parameters for input
  normalisation were computed exclusively on each training fold and
  applied consistently to the corresponding validation fold to
  prevent data leakage.

  \item \textbf{Extrapolation and Epistemic Uncertainty.}
  To evaluate model behaviour under unobserved
  extreme high-pressure conditions (120, 160, and 200~bar),
  a Deep Ensemble approach~\cite{bib49}
  comprising 100 independently trained models was employed.
  These ensembles were trained exclusively on observations
  with cathode pressures $\leq 80$~bar ($n = 42$),
  isolating the high-pressure extrapolation regime as a
  pressure-domain-unseen test set.
  Two backbone calibration protocols were evaluated to
  assess sensitivity to parameterisation scope. Under the
  \textit{Protocol-IEP},
  the physics backbone parameters ($a_\alpha$, $b_\alpha$,
  $a_\beta$, $b_\beta$) were estimated exclusively from
  the extrapolation training subset ($\leq$80~bar, $n=42$),
  maintaining isolation of the high-pressure test
  domain from all calibration steps. Under the
  \textit{Protocol-FCP} (Full-data Calibration Protocol),
  backbone parameters were pre-calibrated on the entire
  compiled dataset ($n = 184$), representing a
  industrial deployment scenario in which a globally
  calibrated physics model is combined with a residual
  network trained on locally available operational data.
  Both protocols used identical neural network training
  configurations.
  It is further noted that the extrapolation training
  subset contains data from three independent sources:
  Wu et al.~\cite{bib39} ($n = 24$; 20, 40, 80~bar;
  25$^{\circ}$C),
  Grigoriev et al.~\cite{bib32} ($n = 9$; 1~bar;
  85$^{\circ}$C),
  and Schalenbach et al.~\cite{bib6} ($n = 9$;
  6~bar; 80$^{\circ}$C). The test set comprises
  exclusively Wu et al.~observations at 120--200~bar,
  25$^{\circ}$C ($n = 24$). This design isolates the
  extrapolation challenge to the pressure domain.
  Each ensemble member was initialised with a
  sequential random seed (base seed $42 +$ model index, where model
  index ranges from 0 to 99) to ensure reproducibility while
  capturing epistemic uncertainty across the ensemble. The ensemble
  mean served as the point prediction, while the ensemble standard
  deviation quantified epistemic uncertainty, enabling construction
  of 95\% confidence bounds for risk assessment under unmapped
  operational conditions.

\end{enumerate}

To establish the statistical significance of performance differences
among the PR-Net, soft-constraint PINN, and purely data-driven NN architectures, a
hierarchical non-parametric testing strategy was adopted. Absolute
prediction errors under extrapolation conditions frequently violate
normality assumptions due to heavy-tailed distributions at extreme
pressures. Joint assessment via the Shapiro--Wilk and D'Agostino
$K^2$ tests was used to assess distributional assumptions for each
model error distribution before selecting the omnibus and post-hoc
procedures. The corresponding test outcomes are reported in
Section~\ref{sec:extrapolation} and Supplementary
Section~\ref{suppl:statistical_testing}.
Non-parametric procedures were therefore adopted throughout. Global
performance differences were first assessed using the Friedman
omnibus test \cite{friedman1937}. Upon rejecting the null hypothesis,
pairwise post-hoc comparisons were conducted using one-sided Wilcoxon
signed-rank tests \cite{wilcoxon1945} with Holm--Bonferroni
correction to control the family-wise error rate across the three
pairwise contrasts \cite{holm1979}. Effect sizes were quantified via
rank-biserial correlation (RBC, $r$), interpreted as small ($|r| < 0.3$),
medium ($0.3 \leq |r| < 0.5$), or large ($|r| \geq 0.5$) following
established conventions \cite{kerby2014}. Differences in mean absolute
error (MAE) were
further validated using distribution-free bootstrap confidence intervals
(10,000 resamples, percentile method)
\cite{efron1994}.


\section{Results and Discussion}
\label{sec:results}

This section should be read in light of two linked objectives of the
present study: first, to establish a unified comparative benchmark for
PEMWE hydrogen crossover prediction across purely data-driven NN,
soft-constraint PINN, and hard-constraint learning; and second, to use
that benchmark to test whether hard-constraint integration delivers a
structural advantage in stability, extrapolation, and scientific
interpretability. The discussion therefore treats the comparative
``first application'' claim as the enabling study design, while placing
primary emphasis on what that design reveals about the relative advantages of
hard-constraint learning.

\subsection{Phenomenological Validation of the Deterministic Backbone}
\label{sec:backbone_validation}

\begin{figure}
\centering
\includegraphics[width=\textwidth]{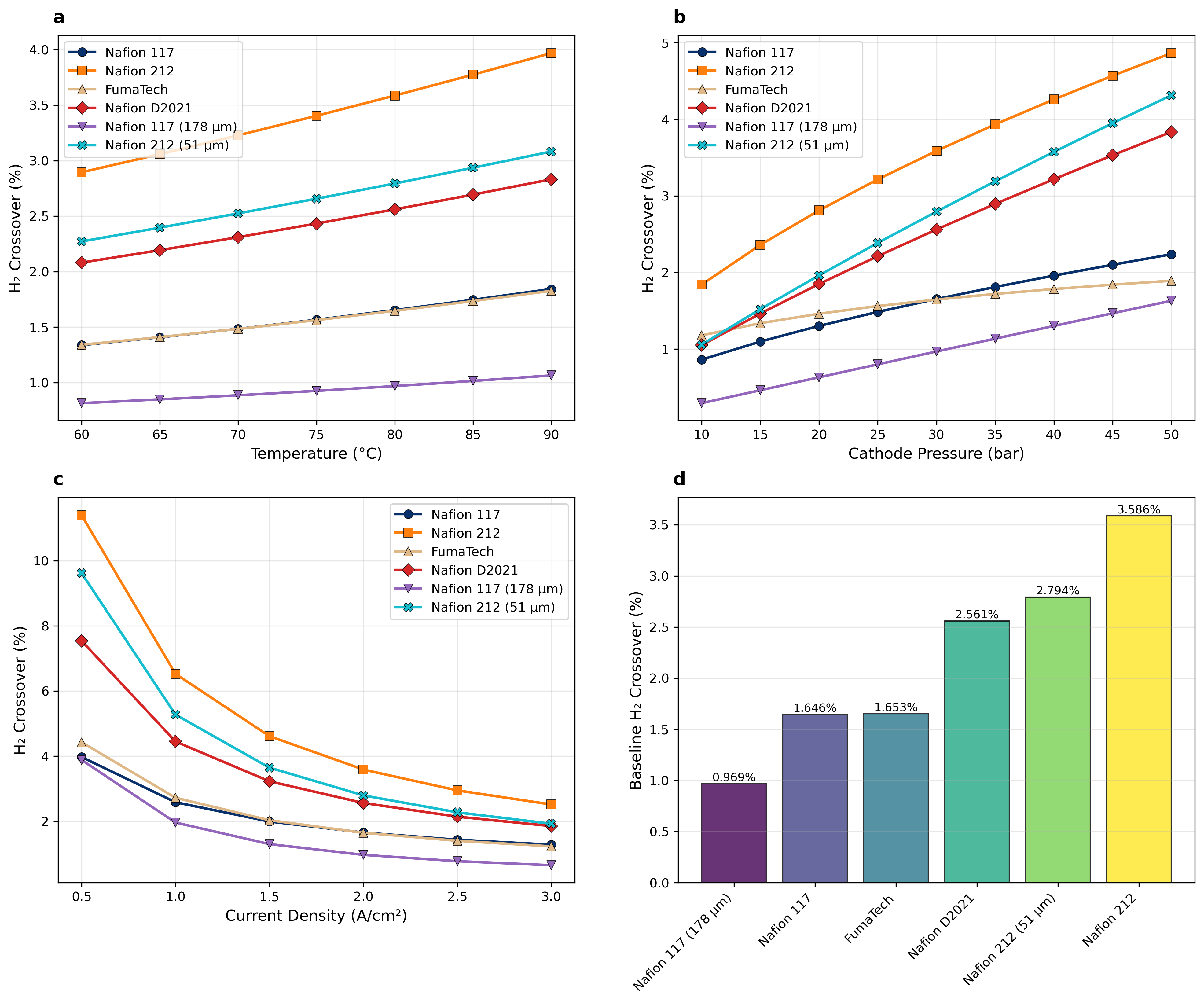}
\caption{Validation of the deterministic physics backbone across six
membrane types ($n = 184$).
(a) Temperature dependence of H$_2$ crossover at 2.0~A~cm$^{-2}$
and 10~bar.
(b) Pressure dependence at 80$^{\circ}$C and 1.0~A~cm$^{-2}$.
(c) Current-density dependence at 80$^{\circ}$C and 30~bar.
(d) Baseline membrane crossover at reference conditions
(80$^{\circ}$C, 10~bar, 2.0~A~cm$^{-2}$).
Marker shapes and colours jointly differentiate the six membrane types
and are kept consistent across panels~(a)--(c) — deep-blue circle
(Nafion~117, 209~$\mu$m), orange square (Nafion~212), light-tan upward
triangle (FumaTech), red diamond (Nafion~D2021), purple downward
triangle (Nafion~117 at 178~$\mu$m), and cyan cross (Nafion~212 at
51~$\mu$m). In panel~(a), the Nafion~117 (209~$\mu$m) curve lies
on the lower-crossover envelope and is partially occluded by the
foreground FumaTech curve; the deep-blue circular markers identify the
underlying Nafion~117 trajectory behind the light-tan FumaTech
triangles.}
\label{fig:backbone_validation}
\end{figure}

Prior to evaluating the machine-learning performance, it is imperative
to establish the phenomenological validity of the deterministic physics
backbone ($\Phi_{\mathrm{H_2}}^{\mathrm{phys}}$) that anchors the
PR-Net architecture. Unlike purely data-driven models that must infer
physical laws implicitly from limited experimental data, our framework embeds
mass transport mechanisms---governed by Henry's law, Fick's diffusion,
and Faraday's law---directly into the computational pipeline (as
detailed in Section~\ref{subsec:physics_model}). To demonstrate that
this backbone captures the macroscopic chemical
engineering trends without any neural network intervention, we
evaluated the baseline hydrogen crossover responses across six distinct
membrane types under controlled operational variations
(Fig.~\ref{fig:backbone_validation}).

As illustrated in Fig.~\ref{fig:backbone_validation}a, the temperature
dependence of H$_2$ crossover at a constant current density of
2.0~A~cm$^{-2}$ and 10~bar reveals significant membrane-specific
behavioural divergences. While thicker membranes (e.g., Nafion~117,
209~$\mu$m) exhibit minimal variation ($<$5\% change), thinner
membranes demonstrate a pronounced thermal sensitivity, with Nafion~212 (51~$\mu$m) showing a 14.8\% increase in crossover concentration
from 60 to 85$^{\circ}$C. The Nafion~117 (209~$\mu$m) trace in
panel~(a) lies along the lower envelope of the family (${\sim}0.7$--$1.0$\%~H$_2$)
and is partially occluded by the foreground FumaTech curve in that range; both
correspond to the lowest-crossover regime at the panel's reference
conditions and therefore appear nearly co-linear. This behaviour is consistent
with the Arrhenius-type temperature dependence of the hydrogen
self-diffusion coefficient ($D_{\mathrm{H_2},w}$) embedded in the
backbone (Eq.~\ref{eq:self_diffusion}).

Furthermore, the pressure response evaluated at 80$^{\circ}$C and
1.0~A~cm$^{-2}$ (Fig.~\ref{fig:backbone_validation}b) validates the
backbone's adherence to pressure--solubility consistency. The
crossover concentration scales proportionally with cathode pressure,
reflecting the elevated dissolved hydrogen concentration at the
catalyst--membrane interface as dictated by Henry's law. A
sub-linear pressure dependence emerges above 30~bar; this attenuation
is captured by the incorporated mechanical compression
model (Eqs.~\ref{eq:porosity_comp}--\ref{eq:permeability_comp}),
which accounts for the increased transport resistance resulting from
LGDL compression under high differential pressures.

The deterministic backbone accurately delineates the
transition between dominant mass transport regimes as a function of
current density (Fig.~\ref{fig:backbone_validation}c). In the low
current density regime ($i < 1.0$~A~cm$^{-2}$), the system is
strictly diffusion-dominated, resulting in steep increases in relative
H$_2$ concentration as the absolute oxygen production rate diminishes.
Conversely, operating in the high current density regime
($i > 2.0$~A~cm$^{-2}$) shifts the system towards a
Faradaic-recombination-limited regime, where the rapid generation of
anodic oxygen effectively dilutes the crossed-over hydrogen. At
reference conditions (80$^{\circ}$C, 10~bar, 2.0~A~cm$^{-2}$), the
backbone correctly ranks the baseline crossover performance of the
evaluated materials (Fig.~\ref{fig:backbone_validation}d),
identifying Nafion~117 (178~$\mu$m) as exhibiting the lowest baseline
crossover concentration (0.969\% H$_2$).

Two Nafion~117 variants are tracked separately throughout this
work and are not pooled: the nominal-thickness specimen
($\approx 209\,\mu$m, $n = 66$; encoded as
\texttt{Nafion\_117}) and a thinner 178\,$\mu$m subset
($n = 20$; encoded as \texttt{Nafion\_117\_178um}). Where the
text refers simply to ``Nafion~117'' the nominal 209\,$\mu$m
variant is intended; explicit thickness annotations are given
wherever the two variants appear together.

By establishing that the physics backbone alone captures a thermodynamically consistent operational envelope, we
fundamentally constrain the hypothesis space for the subsequent neural
network component. The residual network
($\mathrm{NN}_{\mathrm{residual}}$) is thereby relieved from the
burden of learning fundamental electrochemical laws; instead, it is
strictly tasked with isolating and correcting unmodelled, non-linear
interfacial phenomena such as dynamic membrane swelling and localised
temperature gradients. This structural decoupling is the key mechanism
that resolves the optimisation conflicts inherent in soft-constraint PINNs,
supporting the robust interpolation
and high-pressure extrapolation discussed in the following
sections.

\subsection{Resolution of Optimisation Conflicts in the Interpolation
Domain}
\label{sec:optimization_resolution}

In electrochemical systems modelling, conventional soft-constraint
PINNs incorporate physical laws merely as penalty terms within a
composite loss function, as in \cite{bib19} and \cite{bib20}. Within the unified
benchmark established in this study, that formulation can be assessed
against both ends of the modelling spectrum: purely data-driven NN on
one side and hard-constraint learning on the other. This comparative
positioning is important because it shows that the central issue is
not simply whether physics is included, but how it is integrated.
Gradient descent in soft-constraint PINNs must still balance
competing objectives between fitting limited, noisy experimental data
and satisfying idealised macroscopic transport equations
\cite{bib20} and \cite{bib24}. In our evaluation, this optimisation tension is
associated with less stable convergence and greater predictive
variability, which may limit the suitability of soft-constraint
approaches in safety-critical chemical-process settings.

To quantify this optimisation tension at the gradient level, the
per-epoch cosine similarity between the data-loss gradient
$\nabla_\theta \mathcal{L}_{\mathrm{data}}$ and the physics-loss
gradient $\nabla_\theta \mathcal{L}_{\mathrm{phys}}$ was recorded
over $700$ training epochs for the soft-constraint PINN baseline.
A negative cosine indicates that the two loss terms pull
the network in opposing parameter directions, so that any gradient
step necessarily worsens one objective while improving the other.
The cosine became negative in $144/700$ epochs ($20.6\%$) overall
and in $34.9\%$ of epochs restricted to the second half of training,
with a maximum consecutive negative-cosine streak of $6$ epochs
(Supplementary Fig.~\ref{fig:supp_exp2_gradient}). The
diagnostic provides direct quantitative grounding for the otherwise
qualitative statement that data-fidelity and physics objectives
remain in genuine conflict throughout much of training in this
data-limited PEMWE regime, consistent with the broader PINN
failure-mode literature on stiff and tightly coupled physical
systems reported in \cite{bib20}, \cite{bib24}, and \cite{Wang2021}. The hard-constraint
structural integration of PR-Net, which removes the physics
objective from the loss surface altogether rather than balancing it
against the data objective, is by construction immune to this
gradient-direction pathology.

\begin{figure}
\centering
\includegraphics[width=\textwidth]{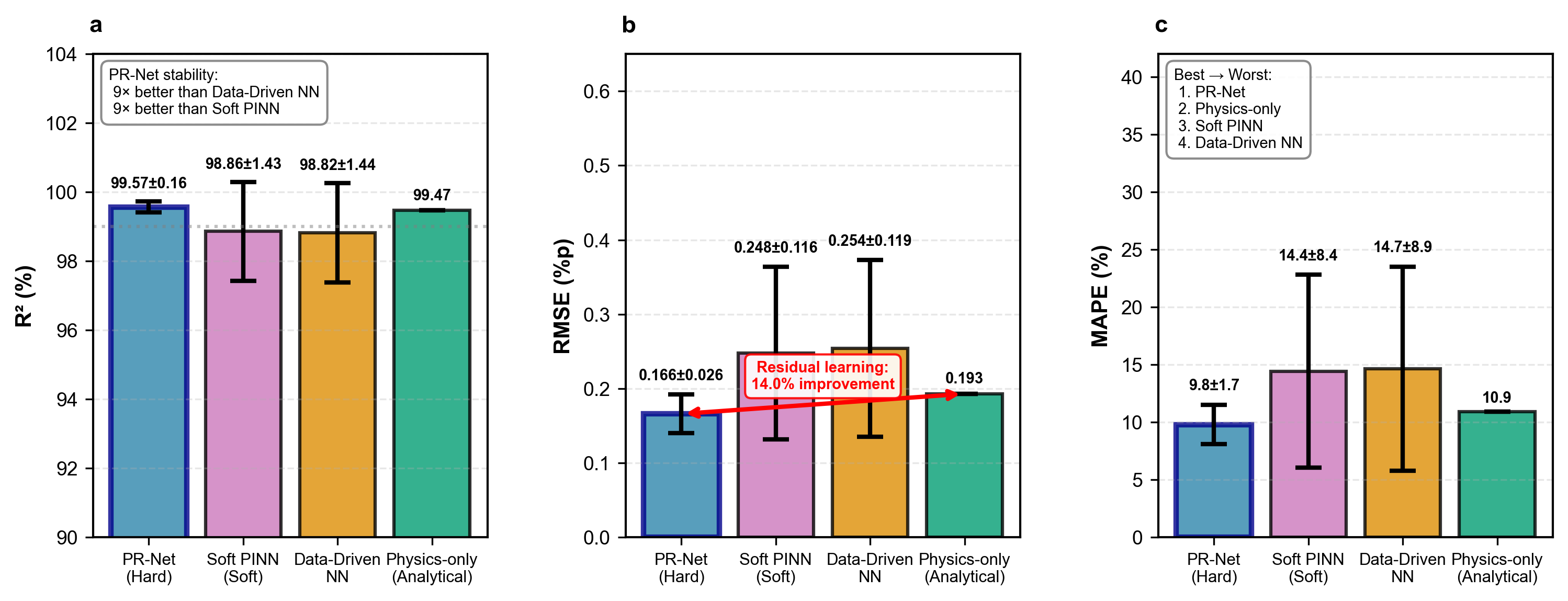}
\caption{Comparison of physics-integration strategies based on 100
model instances ($n = 184$; 5-fold cross-validation $\times$ 20 repetitions).
(a) Coefficient of determination ($R^2$).
(b) Root-mean-square error (RMSE).
(c) Mean absolute percentage error (MAPE).}
\label{fig:cv_comparison}
\end{figure}

By embedding the phenomenologically validated transport equations as
a deterministic backbone (Section~\ref{sec:backbone_validation}),
the proposed PR-Net structurally mitigates this optimisation
conflict. The neural network is constrained to learn only the
systematic residuals---the non-linear discrepancies between idealised
physics and experimental reality. Fig.~\ref{fig:cv_comparison} and
Table~\ref{tab:interpolation} comprehensively compare the predictive
stability of three modelling paradigms evaluated over 100 independent
training instances across the full experimental domain ($n = 184$,
1--200~bar). Fig.~\ref{fig:cv_comparison}a shows the tightly
concentrated $R^2$ distribution of PR-Net. The PR-Net framework
demonstrates high training stability, yielding $R^2 = 99.57 \pm 0.16$\%,
representing a 9-fold
reduction in variability compared to the purely data-driven NN
($R^2 = 98.82 \pm 1.44$\%) and a 9-fold reduction relative to
the soft-constraint PINN ($R^2 = 98.86 \pm 1.43$\%). Furthermore, by accurately
capturing unmodelled interfacial phenomena, the PR-Net residual
learning framework refines the physics-only baseline prediction,
reducing root-mean-square error (RMSE) by 14\% from 0.193\%p to
0.166\%p, consistent with
Fig.~\ref{fig:cv_comparison}b--c.

\begin{table}[ht]
\centering
\caption{Interpolation performance on the full experimental domain
($n = 184$). Values represent mean $\pm$ s.d. from stratified 5-fold
cross-validation repeated 20 times (100 model instances per
architecture). Reported metrics include root-mean-square error
(RMSE), mean absolute error (MAE), and mean absolute percentage error
(MAPE). $\mathrm{CV}_{R^2}$ denotes the coefficient of variation of
$R^2$.}
\label{tab:interpolation}
\begin{tabular}{lcccccc}
\hline
\textbf{Method} & \textbf{Architecture}
  & \textbf{$R^2$ (\%)}
  & \textbf{RMSE (\%p)}
  & \textbf{MAE (\%p)}
  & \textbf{MAPE (\%)}
  & \textbf{$\mathrm{CV}_{R^2}$ (\%)} \\
\hline
PR-Net (ours) & Hard-constraint
  & $\mathbf{99.57 \pm 0.16}$
  & $\mathbf{0.166 \pm 0.026}$
  & $\mathbf{0.124 \pm 0.019}$
  & $\mathbf{9.80 \pm 1.71}$
  & $\mathbf{0.16}$ \\
Soft PINN & Soft-constraint
  & $98.86 \pm 1.43$
  & $0.248 \pm 0.116$
  & $0.165 \pm 0.074$
  & $14.43 \pm 8.39$
  & $1.43$ \\
Data-driven NN & Data-driven
  & $98.82 \pm 1.44$
  & $0.254 \pm 0.119$
  & $0.165 \pm 0.082$
  & $14.66 \pm 8.86$
  & $1.44$ \\
Physics-only & Analytical
  & $99.47$
  & $0.193$
  & $0.140$
  & $10.93$
  & --- \\
\hline
\end{tabular}
\end{table}

The aggregated out-of-fold (OOF) predictions derived from repeated
cross-validation further substantiate the high precision of the
hard-constraint architecture. The ensemble parity plot
(Fig.~\ref{fig:prnet_detail}a) exhibits close agreement with
experimental measurements across the full range of crossover
concentrations (aggregate $R^2 = 0.9960$, RMSE $= 0.1673$\%p).
This predictive fidelity is consistently maintained
across all six structurally and chemically diverse membrane
types---spanning thicknesses from 51~$\mu$m (Nafion~212)
to 260~$\mu$m (FumaTech E-730)---supporting the interpretation that the residual
representation generalises across the evaluated membrane types without obvious overfitting to
membrane-specific artefacts. Analysis of the
prediction residuals (Fig.~\ref{fig:prnet_detail}b) reveals an
approximately normal distribution centred near zero ($\mu = 0.0131$\%),
indicating that the physics backbone helps mitigate
systematic bias. Fig.~\ref{fig:prnet_detail}c (MAPE, RMSE, and
$R^2$) supports this high-fidelity interpolation behaviour.

\subsubsection{Epistemic Uncertainty in the Interpolation Domain}

\begin{figure}
\centering
\includegraphics[width=\textwidth]{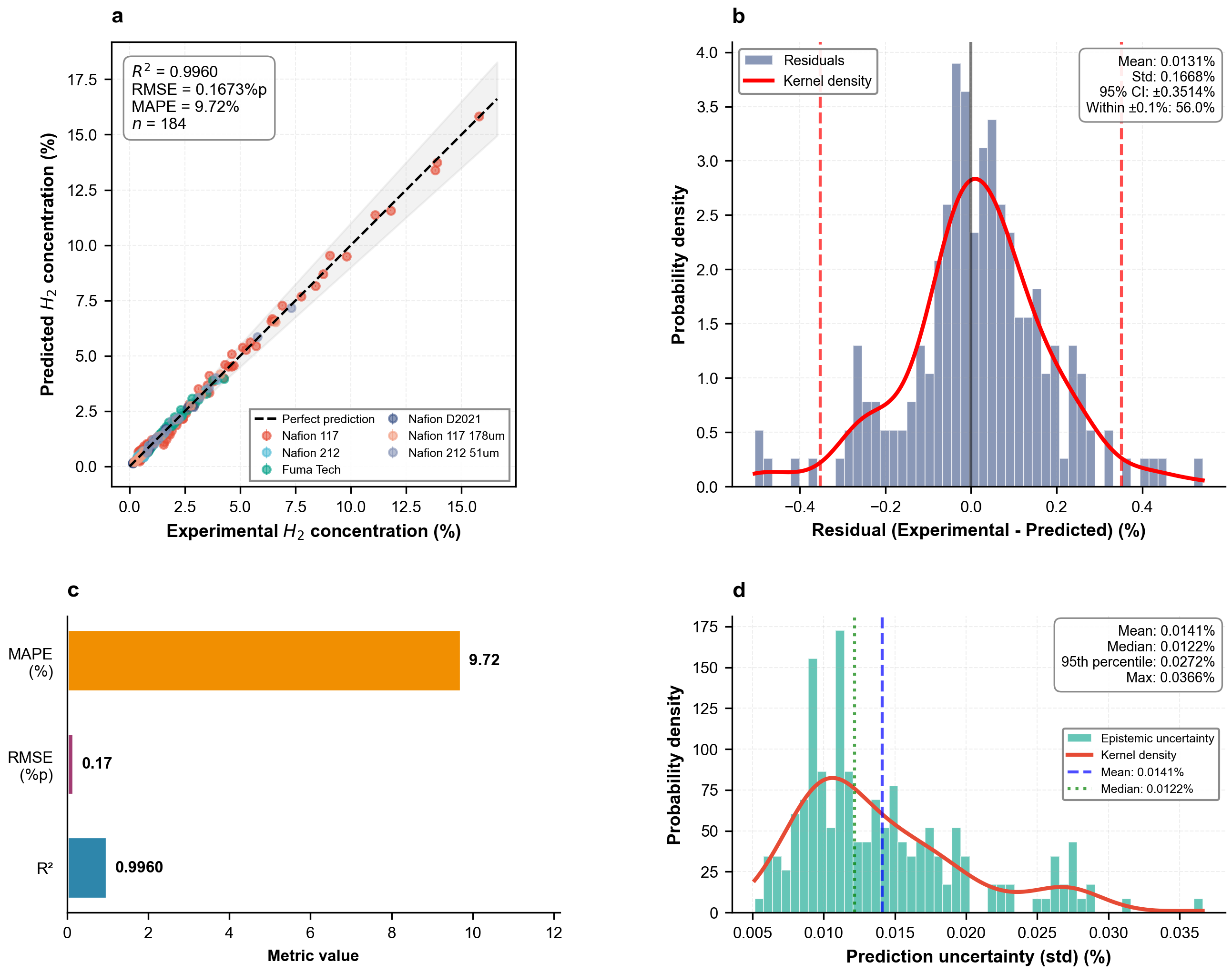}
\caption{Interpolation performance and training stability of PR-Net
on the full dataset ($n = 184$, out-of-fold predictions).
(a) Ensemble parity plot of out-of-fold predictions versus
experimental hydrogen crossover concentrations ($R^2 = 0.9960$,
RMSE $= 0.1673$\%p).
(b) Residual distribution with near-zero mean error
($\mu = 0.0131$\%).
(c) Aggregate out-of-fold performance metrics.
(d) Distribution of epistemic uncertainty across the 100-model CV
ensemble ($\sigma = 0.0141$\%; 95th percentile = 0.0272\%).}
\label{fig:prnet_detail}
\end{figure}

Beyond point accuracy, adaptive safety control in industrial
electrolysers requires rigorous bounds on predictive confidence,
particularly when operating near the explosive threshold of 4\%
H$_2$ in O$_2$. The epistemic uncertainty was quantified from the
spread of predictions across the 100-model cross-validation ensemble
(5-fold $\times$ 20 repetitions, trained on all 184 experimental
points). As shown in Fig.~\ref{fig:prnet_detail}d, the mean
prediction standard deviation is $\sigma = 0.0141$\%, with 95\% of
predictions exhibiting uncertainty below 0.0272\%. The well-separated
scale between this uncertainty bound and the 4\% safety threshold
(factor $>$100) provides a substantial confidence margin for
real-time process control within the training domain. This tight
confidence bound supports the view that the hard physics-residual constraint
effectively regularises the optimisation landscape within the training domain.
Uncertainty calibration under high-pressure extrapolation
conditions ($\leq 80$~bar training; 120--200~bar test) and
operational sensitivity analysis, including sensor placement
implications, are addressed in Section~\ref{sec:uncertainty}.

\subsubsection{Training Cost Across the Three Architectures}
\label{sec:training_cost_result}

The three paradigms also differ in training cost. Benchmarked across
both training regimes with 10 random seeds on CPU and normalised by
dataset size, the physics backbone raises the per-sample training cost
to roughly $4$--$6\times$ that of the bare data-driven NN ($221$ versus
$35$~ms/sample in interpolation). However, PR-Net's hard constraint
lowers the number of epochs to convergence ($233$ versus $409$ for the
soft-constraint PINN and $433$ for the data-driven NN), so its total
training time tracks the soft-constraint PINN and remains modest in
absolute terms ($40.8$ versus $40.4$~s in interpolation; $6.9$ versus
$8.6$~s in the $\leq 80$~bar extrapolation regime). The high-pressure
extrapolation pipeline (Section~\ref{sec:extrapolation}) additionally
carries a one-time, shared differential-evolution backbone calibration
of $24.9$~s. The improved predictive stability of PR-Net is thus
obtained without a training-time penalty relative to the
soft-constraint baseline, at a total cost well under a minute on CPU.
The benchmark protocol, full per-regime timings, time per sample, best
epoch, and seed variability are given in Supplementary
Section~\ref{suppl:computational_framework} (``Training-cost
comparison across architectures''; Table~\ref{tab:s_training_cost},
Fig.~\ref{fig:s_training_cost}).

\subsection{Pressure-Axis Extrapolation at Extreme Pressures}
\label{sec:extrapolation}

\subsubsection{Pressure-axis extrapolation performance}
\label{sec:extrapolation_performance}

In industrial PEM water electrolysis, adaptive safety
monitoring systems must provide reliable predictions even
when operational conditions deviate suddenly from
historical data, such as during extreme pressure surges
or emergency transients. To evaluate this
critical capability, the trained models (exposed
exclusively to cathode pressures $\leq 80$~bar, $n = 42$)
were subjected to a stringent stress test at 120, 160,
and 200~bar.
Extrapolating from
$\leq$80~bar to $\geq$120~bar constitutes entry into
a thermodynamic regime not represented in the training data, not merely
an interpolation beyond measured points. At cathode
pressures exceeding 120~bar, several physical phenomena
become dominant that are virtually absent from the
low-pressure experimental data range: hydrogen deviates
measurably from ideal gas behaviour (fugacity coefficient
$\phi \approx 1.07$ at 200~bar, 25$^{\circ}$C via the Peng--Robinson
equation of state (PR-EOS)~\cite{PengRobinson1976}, quantified below
in Section~\ref{sec:fugacity_correction}); membrane swelling dynamics enter
a strongly non-linear regime; and LGDL compressibility
effects approach saturation. These conditions represent
a 2.5-fold extrapolation beyond the training domain
in both pressure magnitude and governing physical
mechanism. All primary results in this section are
reported under Protocol-IEP (zero-leakage
calibration); Protocol-FCP (full-data calibration) results are included in
Table~\ref{tab:extrapolation_r2} for comparison.

\begin{figure}
\centering
\includegraphics[width=\textwidth]{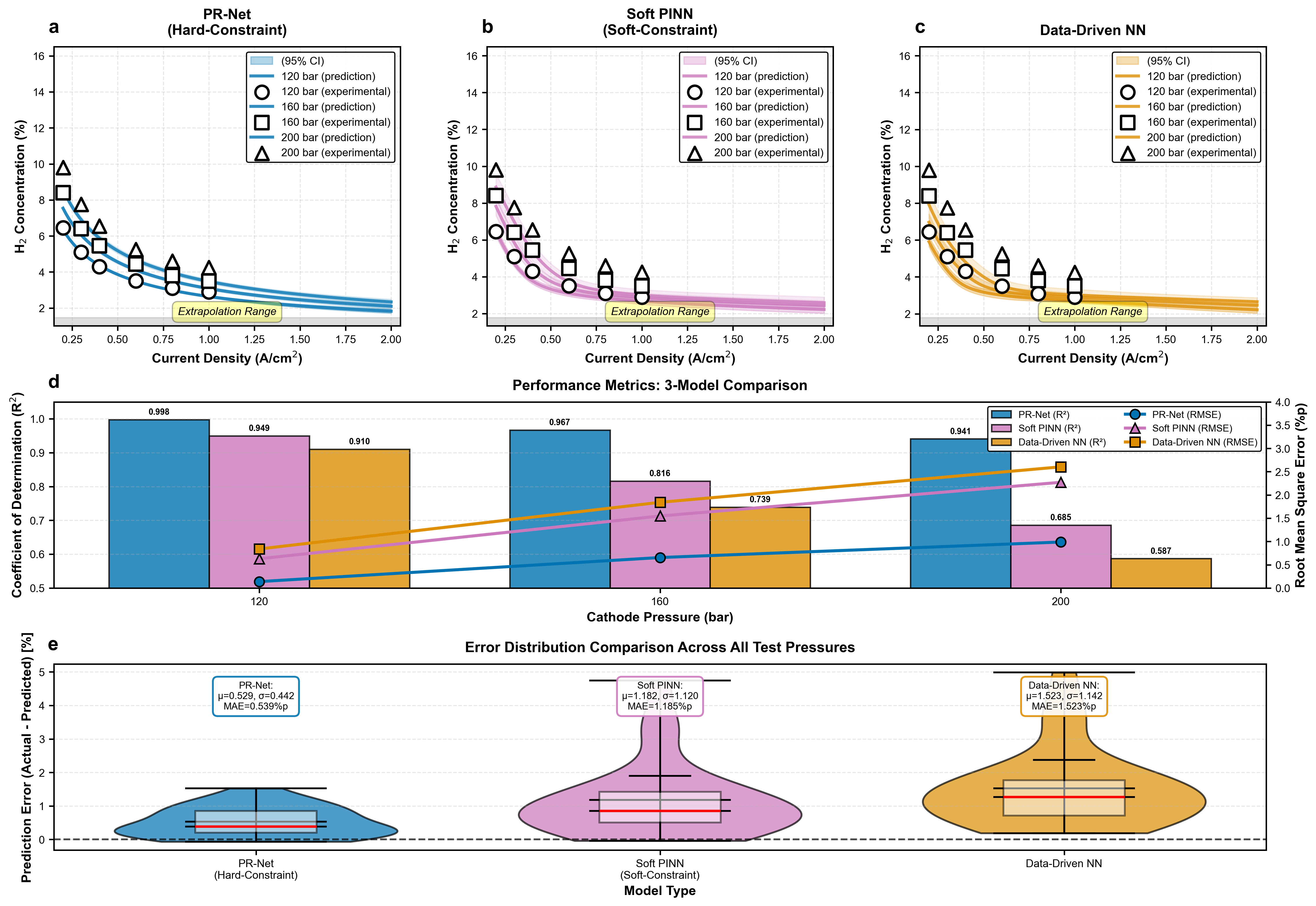}
\caption{Pressure-axis extrapolation beyond the training domain.
Training: Nafion~117 at $\leq 80$~bar
($n = 42$; mixed 25/80/85$^{\circ}$C across the three contributing
sources). Test: Nafion~117 at 120--200~bar, 25$^{\circ}$C ($n = 24$;
Wu et al.~only).
(a) PR-Net predictions with 95\% confidence intervals up to 200~bar
($R^2 = 94.02 \pm 0.92$\%).
(b) Soft-constraint PINN predictions up to 200~bar
($R^2 = 68.06 \pm 5.52$\%).
(c) Purely data-driven NN predictions up to 200~bar
($R^2 = 58.00 \pm 8.60$\%).
(d) RMSE at 200~bar, showing 62\% and 57\% reductions for PR-Net
relative to the purely data-driven NN and the soft-constraint PINN,
respectively.
(e) Signed-error distributions across all extrapolation pressures,
showing a narrower and more symmetric distribution for PR-Net
(mean signed error $\mu = 0.529$\%, $\sigma = 0.442$\%).}
\label{fig:extrapolation}
\end{figure}
Fig.~\ref{fig:extrapolation} illustrates the comparative extrapolation
behaviour of the three models under these extreme conditions.
As shown in Fig.~\ref{fig:extrapolation}a, the proposed hard-constraint model
remains closely aligned with the expected thermodynamic trend, maintaining $R^2 = 94.02 \pm 0.92\%$ at
200~bar---a regime where non-ideal gas behaviour and
strongly non-linear membrane swelling dynamics become prominent.
Apparatus-bias bounding analyses (Strategies A--C) for this comparison are presented in Section~\ref{sec:apparatus_bias_caveat}.
By comparison, the soft-constraint PINN (Fig.~\ref{fig:extrapolation}b) exhibits
a ``forgetting''-like pattern; despite
incorporating physics in its loss function, it
appears to prioritise the training manifold over physical trends
in the extrapolation regime, reaching
$R^2 = 68.06 \pm 5.52\%$. The purely data-driven NN (Fig.~\ref{fig:extrapolation}c)
shows substantial performance degradation ($R^2 = 58.00
\pm 8.60\%$), with a plateau-like response that appears inconsistent with the pressure--solubility
relationship encoded in Henry's law.
The $\Delta$ rows in Table~\ref{tab:extrapolation_r2}
further confirm that the PR-Net advantage is robust
to backbone calibration scope: the performance gap
between PR-Net and the soft-constraint PINN at 200~bar is
25.96 percentage points under Protocol-IEP and 25.62 percentage points under
Protocol-FCP, demonstrating that the architectural
advantage does not depend on the extent of backbone
pre-calibration. Under Protocol-FCP---which simulates
the industrial scenario of deploying a globally
calibrated physics model---PR-Net achieves an
upper-bound $R^2 = 97.34 \pm 0.69\%$ at 200~bar.
Pressure-stratified $R^2$ values across all three architectures are
summarised in Table~\ref{tab:extrapolation_r2}.

\begin{table}[ht]
\caption{Extrapolation performance at high-pressure conditions
beyond the training range (120, 160, and 200~bar). Models were trained
on Nafion~117 at cathode pressures $\leq 80$~bar ($n = 42$; mixed
25/80/85$^{\circ}$C across the three contributing sources); the test
partition is Nafion~117 at 25$^{\circ}$C, 120--200~bar
($n = 24$; Wu et al.~only).
Values represent mean $\pm$ s.d. across 100 ensemble members.
Protocol-IEP uses backbone calibration from training data only;
Protocol-FCP uses backbone pre-calibration on the full dataset.
$\Delta = \text{Protocol-FCP} - \text{Protocol-IEP}$.}
\label{tab:extrapolation_r2}
\resizebox{\textwidth}{!}{%
\begin{tabular}{llccccc}
\toprule
Method & Protocol
       & 120~bar $R^2$ (\%)
       & 160~bar $R^2$ (\%)
       & 200~bar $R^2$ (\%)
       & Overall $R^2$ (\%)
  & MAE (\%p) \\
\midrule
\textbf{PR-Net}
  & \textbf{IEP$^\dagger$}
  & $\mathbf{99.74\pm0.06}$
  & $\mathbf{96.68\pm0.52}$
  & $\mathbf{94.02\pm0.92}$
  & $\mathbf{96.51\pm0.54}$
  & $\mathbf{0.539\pm0.429}$ \\
  & FCP$^\ddagger$
  & $99.61\pm0.14$
  & $98.80\pm0.35$
  & $97.34\pm0.69$
  & $98.49\pm0.36$
  & $0.375\pm0.037$ \\
  & $\Delta$ (FCP$-$IEP)
  & $-0.13$
  & $+2.12$
  & $+3.32$
  & $+1.98$
  & $-0.164$ \\
\midrule
\textbf{Soft PINN}
  & \textbf{IEP$^\dagger$}
  & $\mathbf{94.78\pm1.12}$
  & $\mathbf{81.28\pm3.30}$
  & $\mathbf{68.06\pm5.52}$
  & $\mathbf{80.35\pm3.43}$
  & $\mathbf{1.185\pm1.116}$ \\
  & FCP$^\ddagger$
  & $95.02\pm0.69$
  & $84.00\pm1.98$
  & $71.72\pm2.88$
  & $82.71\pm1.89$
  & $0.991\pm0.058$ \\
  & $\Delta$ (FCP$-$IEP)
  & $+0.24$
  & $+2.72$
  & $+3.66$
  & $+2.36$
  & $-0.194$ \\
\midrule
\textbf{Data-driven NN}
  & —
  & $90.75\pm2.31$
  & $73.46\pm5.38$
  & $58.00\pm8.60$
  & $73.10\pm5.53$
  & $1.523\pm1.142$ \\
\bottomrule
\end{tabular}}
\end{table}

The performance of the PR-Net at 200~bar demonstrates an advantage
regarding thermodynamic non-ideality. The deterministic backbone
(Section~\ref{subsec:physics_model}) employs Henry's law with an
ideal gas assumption to calculate the effective hydrogen pressure at
the membrane interface. At pressures approaching 200~bar,
intermolecular interactions induce non-negligible deviations from
ideality; a rigorous treatment would require fugacity corrections of
the form $f_{\mathrm{H_2}} = \phi(T,P) \cdot P$, where $\phi$ is the
fugacity coefficient from a real-gas equation of state such as the
Peng--Robinson model \cite{PengRobinson1976}. Because
$\mathrm{NN}_{\mathrm{residual}}$ is structurally free to apply
targeted corrections where the idealised backbone deviates
most, it can empirically account for part of these non-ideal
contributions within the tested range, achieving high fidelity without
requiring explicit fugacity modelling.

The quantitative impact is summarised in
Fig.~\ref{fig:extrapolation}d. At 200~bar, the PR-Net reduces RMSE
by 62\% relative to the purely data-driven NN (2.612\%p
vs.\ 0.988\%p) and 57\% relative to the soft-constraint PINN (2.281\%p vs.\ 0.988\%p).
The error distribution analysis
(Fig.~\ref{fig:extrapolation}e) indicates that
the PR-Net achieves a compact error distribution
($\mu = 0.529$\%, $\sigma = 0.442$\%), whereas the
soft-constraint PINN exhibits a broad, heavy-tailed distribution
($\mu = 1.182$\%, $\sigma = 1.120$\%) unsuitable for
safety-critical decision-making near the 4\% H$_2$
explosive threshold.

\paragraph{Statistical validation.}

To formally establish the significance of observed performance
differences, a hierarchical non-parametric testing strategy was
applied to the $n = 24$ extrapolation test samples. As the soft-constraint PINN
and the purely data-driven NN error distributions departed significantly from normality
(Shapiro--Wilk and D'Agostino $K^2$, $p < 0.001$), non-parametric
procedures were adopted throughout, whereas the PR-Net error distribution
remained consistent with normality (Shapiro--Wilk: $p = 0.077$;
D'Agostino $K^2$: $p = 0.327$).
The Friedman omnibus test confirmed globally significant
performance differences among the three architectures ($\chi^2 = 30.33$, $p < 0.001$).
Post-hoc pairwise Wilcoxon signed-rank tests with
Holm--Bonferroni correction yielded large effect sizes
for all comparisons (Table~\ref{tab:statistical_validation}): the PR-Net significantly
outperformed the soft-constraint PINN ($r = -0.847$,
$p_{\mathrm{adj}} = 0.000038$) and the purely data-driven NN
($r = -0.993$, $p_{\mathrm{adj}} < 0.0001$).
The soft-constraint PINN likewise significantly outperformed
the purely data-driven NN ($r = -0.927$, $p_{\mathrm{adj}} = 0.000007$),
confirming that even partial physics
integration confers measurable benefit over purely data-driven
approaches.

\begin{table}[ht]
\centering
\caption{Pairwise statistical comparison of extrapolation
performance ($n = 24$; 120--200~bar; Nafion~117, 25$^{\circ}$C).
Post-hoc one-sided Wilcoxon signed-rank tests use Holm--Bonferroni
correction. Bootstrap 95\% confidence intervals (10,000 resamples,
percentile method) are reported for the mean MAE difference
(Model~A $-$ Model~B).}
\label{tab:statistical_validation}
\begin{tabular}{lcccc}
\toprule
Comparison (A vs B) & $r$ (RBC) & $p_{\mathrm{adj}}$
  & MAE diff.\ (\%p) & 95\% CI (\%p) \\
\midrule
PR-Net vs Soft PINN
  & $-0.847$ & $0.000038$
  & $-0.646$ & $[-1.031,\ -0.344]$ \\
PR-Net vs Data-driven NN
  & $-0.993$ & $< 0.0001$
  & $-0.984$ & $[-1.349,\ -0.685]$ \\
Soft PINN vs Data-driven NN
  & $-0.927$ & $0.000007$
  & $-0.338$ & $[-0.458,\ -0.223]$ \\
\bottomrule
\end{tabular}
\end{table}

The pressure-stratified MAE values and architecture-specific
degradation slopes are presented in Table~\ref{tab:mae_degradation},
providing a quantitative proxy for the rate at which performance
diverges with extrapolation distance.

\begin{table}[ht]
\centering
\caption{Pressure-stratified mean absolute error (MAE; mean $\pm$
s.d., \%p) and MAE degradation slope (\%p/bar) for the three
architectures under Protocol-IEP across 120--200~bar. Slopes are
estimated by linear regression of mean MAE against cathode pressure.}
\label{tab:mae_degradation}
\begin{tabular}{lccccc}
\toprule
Model & 120~bar & 160~bar & 200~bar
      & Slope (\%/bar) & Ratio vs PR-Net \\
\midrule
PR-Net
  & $0.106\pm0.088$ & $0.576\pm0.312$
  & $0.935\pm0.319$ & $0.01036$ & $1.00\times$ \\
Soft PINN
  & $0.450\pm0.439$ & $1.237\pm0.930$
  & $1.868\pm1.293$ & $0.01772$ & $1.71\times$ \\
Data-driven NN
  & $0.691\pm0.477$ & $1.579\pm0.948$
  & $2.300\pm1.220$ & $0.02012$ & $1.94\times$ \\
\bottomrule
\end{tabular}
\end{table}

The PR-Net MAE degradation slope
(0.0103\%/bar, estimated by linear regression across
120--200~bar) remains substantially lower than that
of the soft-constraint PINN (0.0177\%/bar, 1.71$\times$ ratio)
and the purely data-driven NN (0.0201\%/bar, 1.94$\times$ ratio),
confirming accelerating performance divergence with
extrapolation distance under leak-free conditions.
All pairwise comparisons reached
statistical significance at every extrapolation pressure,
including 120~bar ($p = 0.020$, uncorrected), confirming
that the architectural advantage is detectable even at
modest extrapolation distances once backbone calibration
is restricted to training-domain data.
The comparable degradation slopes of the soft-constraint PINN and
the purely data-driven NN (ratio: 1.14$\times$) suggest that soft-constraint
physics integration does not qualitatively alter the
extrapolation failure mechanism: both architectures
ultimately rely on interpolation within the training
manifold and diverge comparably once cathode pressure
substantially exceeds the training range.

\begin{figure}
\centering
\includegraphics[width=0.70\textwidth]{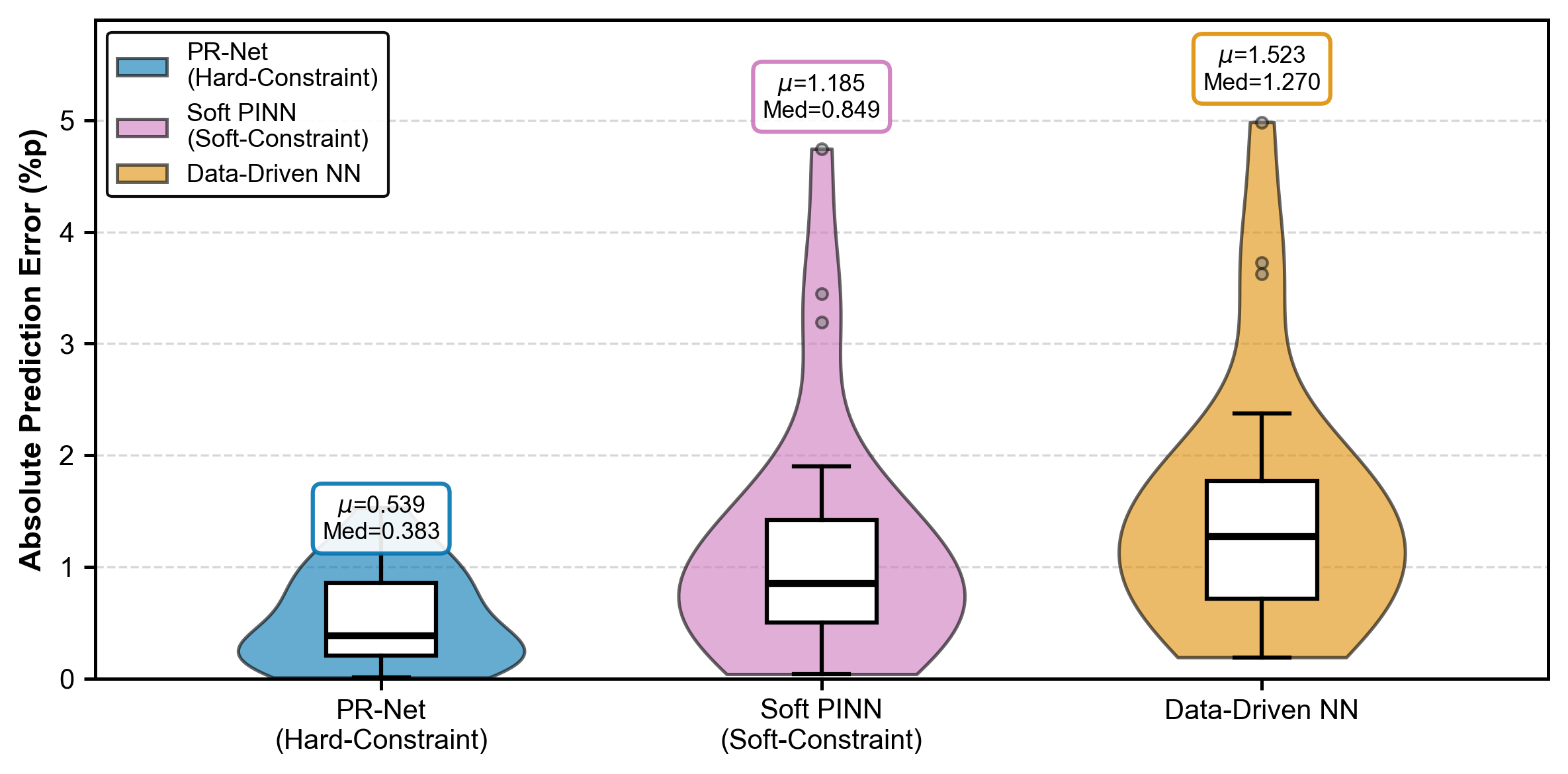}
\caption{Absolute prediction error distributions across the full
extrapolation range (120--200~bar; $n = 24$; Nafion~117,
25$^{\circ}$C). Violin width denotes the kernel density estimate;
box plots indicate the median, interquartile range, and 1.5$\times$
IQR whiskers; open circles denote outliers. PR-Net shows the most
compact distribution ($\mu_{\mathrm{MAE}} = 0.539$\%p, median
$= 0.383$\%p), compared with the soft-constraint PINN
($\mu_{\mathrm{MAE}} = 1.185$\%p, median $= 0.849$\%p) and the
purely data-driven NN ($\mu_{\mathrm{MAE}} = 1.523$\%p, median
$= 1.270$\%p). The Friedman test indicates overall differences among
architectures ($\chi^2 = 30.33$, $p < 0.001$), and all pairwise
Wilcoxon comparisons yield large effect sizes ($|r| \geq 0.847$;
Table~\ref{tab:statistical_validation}).}
\label{fig:violin_errors}
\end{figure}

Fig.~\ref{fig:violin_errors} presents the absolute error
distributions across all extrapolation pressures as violin plots.
The compact distribution of the PR-Net
($\mu = 0.539$\%p, median $= 0.383$\%p) contrasts markedly
with the broad, right-skewed distributions of the soft-constraint PINN ($\mu = 1.185$\%p, median $= 0.849$\%p) and the purely data-driven NN
($\mu = 1.523$\%p, median $= 1.270$\%p), reflecting the
stabilising effect of the hard-constraint physics backbone on
prediction behaviour beyond the training domain. The substantially
larger standard deviations of the soft-constraint PINN and the purely data-driven NN reflect
pressure-amplified error instability consistent with the degradation
slopes reported in Table~\ref{tab:mae_degradation}.

\begin{figure}
\centering
\includegraphics[width=\textwidth]{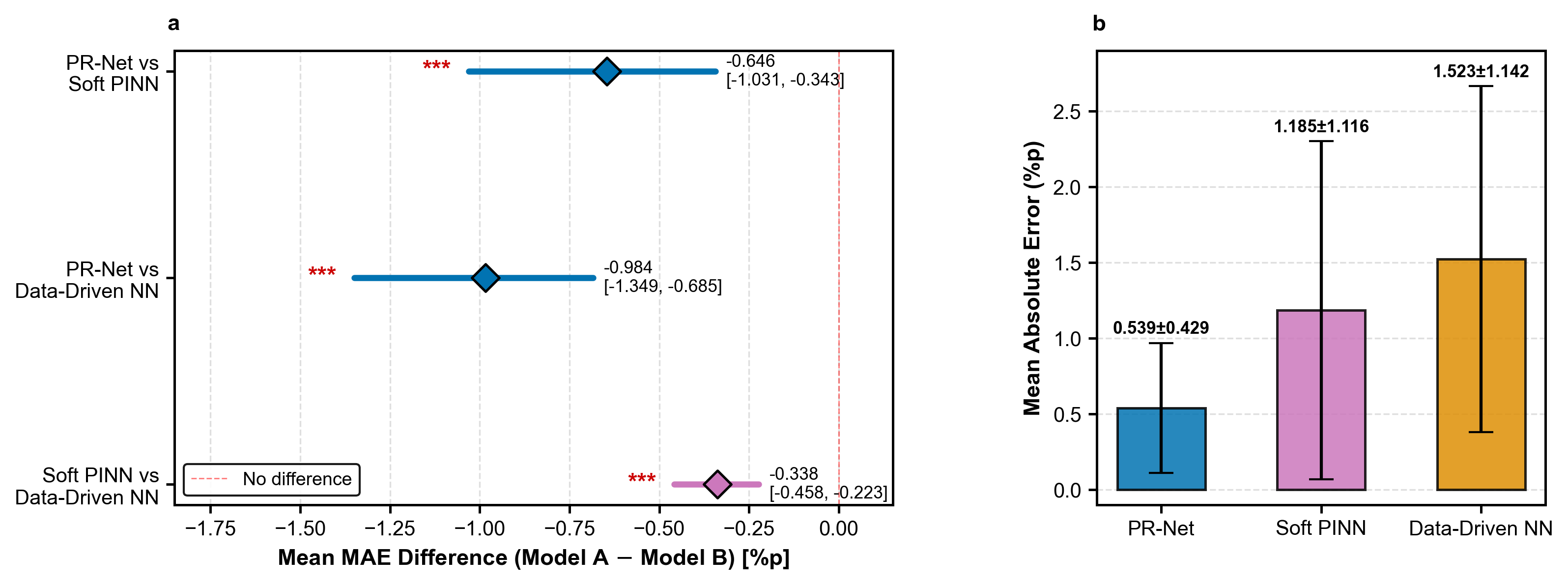}
\caption{Bootstrap confidence intervals for pairwise MAE
differences and architecture-specific MAE over the full extrapolation
domain (120--200~bar; $n = 24$; Nafion~117, 25$^{\circ}$C).
(a) Forest plot of 95\% bootstrap confidence intervals for mean MAE
differences (Model~A $-$ Model~B; percentile method, 10,000
resamples).
(b) Mean MAE (mean $\pm$ s.d.) across the extrapolation range for
PR-Net, the soft-constraint PINN, and the purely data-driven NN.}
\label{fig:bootstrap_ci}
\end{figure}

Fig.~\ref{fig:bootstrap_ci}a presents the bootstrap-derived 95\%
confidence intervals for pairwise MAE differences, and
Fig.~\ref{fig:bootstrap_ci}b summarises architecture-specific mean
MAE dispersion across the extrapolation domain. All three confidence
intervals lie below zero, confirming statistical and practical
significance. The non-overlapping intervals further demonstrate that
the PR-Net, the soft-constraint PINN, and the purely data-driven NN occupy distinct performance
tiers in the extrapolation domain, with no ambiguity about their
relative ordering.

These pressure-axis extrapolation results are further supported by source-holdout robustness checks across four feasible leave-one-source-out (LOSO) sources (St\"{a}hler et al.~(2020), Garbe et al.~(2019), Grigoriev et al.~(2011), and Schalenbach et al.~(2013); Supplementary Table~\ref{tab:s_loso_summary}). In the extrapolation protocol itself, low-pressure training data are pooled from Wu et al.~(2025) (20, 40, 80~bar) together with Grigoriev et al.~(2011) (1~bar) and Schalenbach et al.~(2013) (6~bar), whereas the high-pressure test domain (120, 160, 200~bar) is exclusively sourced from Wu et al.~(2025). Accordingly, the observed pressure-domain generalisation should be interpreted as robust within this pressure-unseen setting, while full high-pressure inter-laboratory independence remains a target for future data collection. Because per-source sample sizes and H$_2$-concentration variance are narrow, LOSO model ranking is interpreted primarily using absolute-error metrics (RMSE and MAE), with $R^2$ reported as a secondary descriptive statistic.

\subsubsection{Apparatus-bias caveat and bounding controls}
\label{sec:apparatus_bias_caveat}
Because the 120--200~bar test set is sourced exclusively from
Wu et al.~\cite{bib39} while training pools low-pressure Nafion~117
points from Wu et al.~($n = 24$ at 20, 40, 80~bar, 25\,$^{\circ}$C),
Grigoriev et al.~\cite{bib32} ($n = 9$ at 1~bar, 85\,$^{\circ}$C),
and Schalenbach et al.~\cite{bib6} ($n = 9$ at 6~bar, 80\,$^{\circ}$C),
the extrapolation gain conflates two contributions that
cannot be fully disentangled with the present corpus: (i) genuine
pressure-axis generalisation into a thermodynamic regime absent
from training, and (ii) apparatus-specific characteristics of the
Wu et al.~high-pressure cell (cell geometry, electrolyte
conditioning, and gas chromatography calibration). To bound the
apparatus-bias contribution, we conducted three complementary
alternative split validations on independent axes:
\emph{Strategy~A} (all-source current-density axis split, $i \leq
1.0$~A~cm$^{-2}$ train versus $i > 2.0$~A~cm$^{-2}$ test;
$n_{\mathrm{train}} = 123$, $n_{\mathrm{test}} = 30$);
\emph{Strategy~B} (the same current-density split with Wu et al.\
removed from both partitions; $n_{\mathrm{train}} = 75$,
$n_{\mathrm{test}} = 30$); and \emph{Strategy~C} (a within-Wu joint
two-dimensional $(P, i)$ split, $P \leq 30$~bar and $i \leq
0.4$~A~cm$^{-2}$ train versus $P > 30$~bar or $i > 0.4$~A~cm$^{-2}$ test;
$n_{\mathrm{train}} = 21$, $n_{\mathrm{test}} = 43$). Strategy~B
isolates the architectural contribution by removing Wu altogether;
Strategy~C isolates the pressure and current-density axes within a
single apparatus. Across all three controls, PR-Net retains the
highest extrapolation $R^2$ among the three machine-learning
architectures (Strategy~A: $0.681$ vs.\ soft-constraint PINN
$0.126$ and data-driven NN $-0.225$; Strategy~B: $0.893$ vs.\
$-0.264$ and $-1.352$; Strategy~C: $0.884$ vs.\ $0.379$ and
$0.365$). The Holm-corrected Wilcoxon contrasts against PR-Net are
non-significant under Strategy~A ($p_{\mathrm{adj}} = 0.061$ and
$0.152$ for the data-driven NN and the soft-constraint PINN) and
significant under Strategies~B and~C ($p_{\mathrm{adj}} < 10^{-6}$);
small training partitions ($n \leq 75$ for B and $n = 21$ for C)
caution against treating the latter as independent statistical
proof. The three alternative splits should therefore be read as
bounding evidence: removing Wu et al.~(Strategy~B) or restricting
the analysis to within-apparatus contrasts (Strategy~C) preserves
the PR-Net advantage over the data-driven and soft-constraint
baselines, indicating that the pressure-axis ranking is
not driven by Wu-specific apparatus signatures alone. The
high-pressure extrapolation claim is accordingly framed as
``pressure-axis generalisation under partial inter-laboratory
mixing,'' with broader cross-apparatus validation at $\geq 120$~bar
identified as a key direction for future data collection. Full
per-strategy parity plots, statistical tables, and per-subgroup
$R^2$ breakdowns are provided in Supplementary
Section~\ref{suppl:alt_split_validations},
Supplementary Fig.~\ref{fig:s_alt_strategies}, and Supplementary
Table~\ref{tab:s_alt_extrap}.

\subsubsection{Quantification of the fugacity correction and empirical test of residual consistency}
\label{sec:fugacity_correction}

A central interpretation of the PR-Net architecture is that the neural
residual $\mathrm{NN}_{\mathrm{residual}}$ can account empirically for
non-ideal contributions left uncompensated by the idealised deterministic
backbone, including fugacity-scale corrections at supercritical
hydrogen pressures. To test this interpretation quantitatively,
we (i) compute the expected magnitude of the Peng--Robinson
fugacity correction across the extrapolation pressure range and
(ii) directly compare PR-Net with an explicit PR-EOS backbone
against PR-Net with the ideal-gas backbone under the identical
Protocol-IEP training regime.

\begin{figure}
\centering
\includegraphics[width=0.95\textwidth]{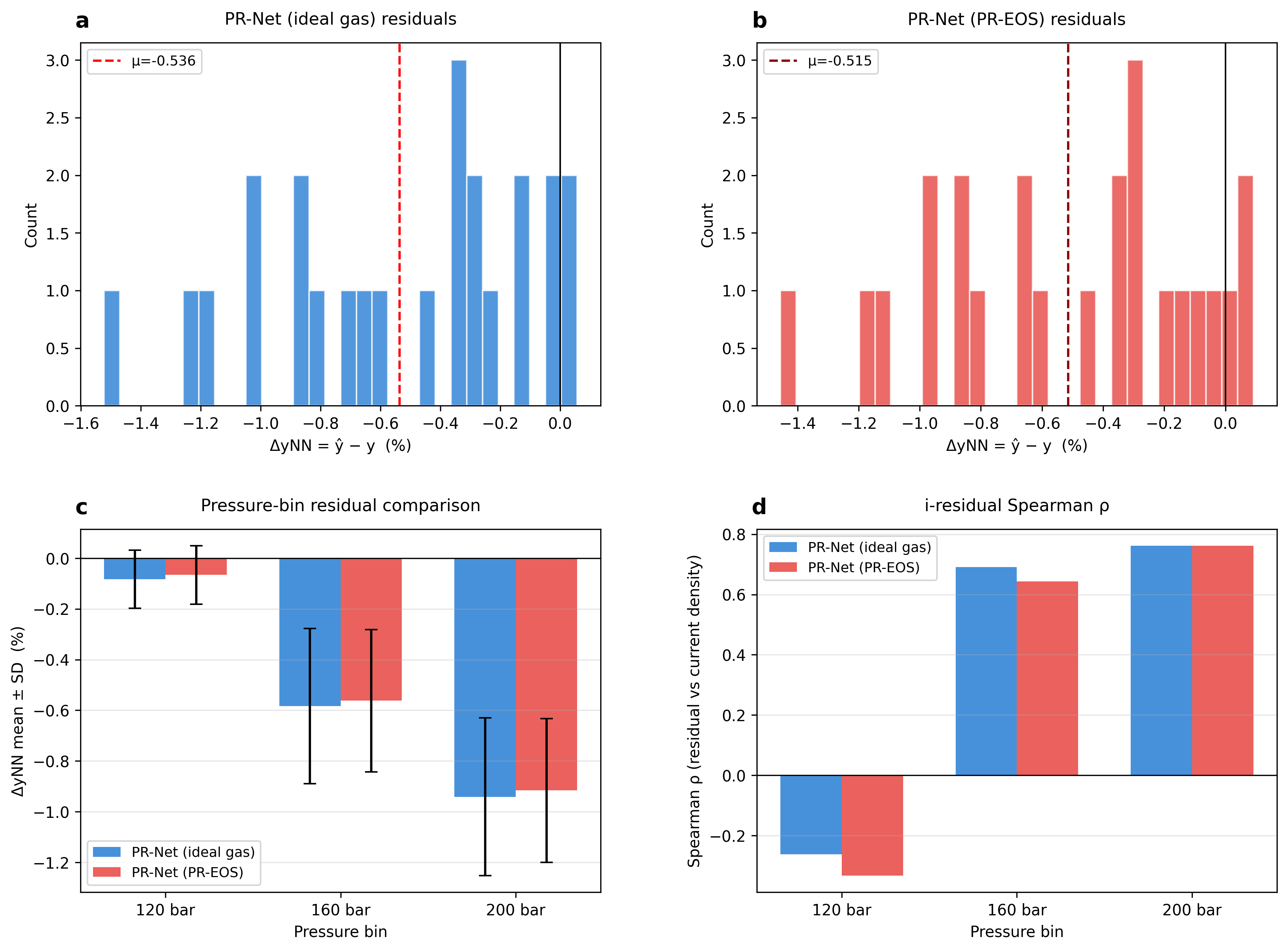}\\[2pt]
\includegraphics[width=0.60\textwidth]{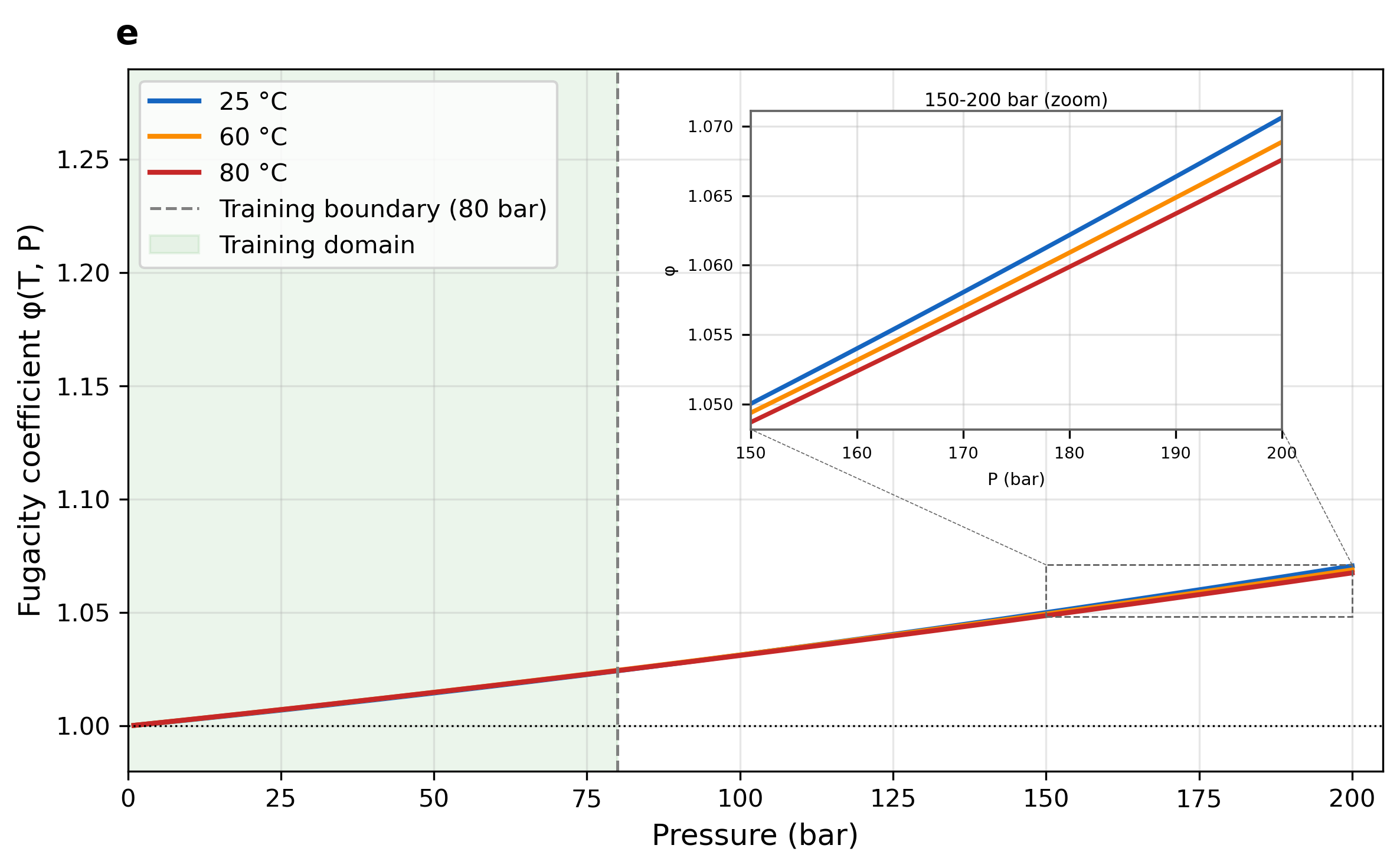}
\caption{Comparison of residuals from PR-Net with the ideal-gas
backbone and PR-Net with the PR-EOS backbone, alongside the
underlying Peng--Robinson fugacity correction.
(a) Per-sample residual distribution for the ideal-gas PR-Net
over the full extrapolation range (120--200~bar; $n = 24$).
(b) Per-sample residual distribution for the PR-EOS variant
over the same range.
(c) Pressure-binned mean residuals with $\pm$1\,s.d.\ for the
two backbones.
(d) Spearman rank correlation between residual and current
density at each pressure bin.
(e) Peng--Robinson fugacity coefficient $\phi(T, P)$ for
hydrogen at 25, 60, and 80$^{\circ}$C from 1 to 200~bar
(inset: 150--200~bar).}
\label{fig:exp1_preos_residuals}
\end{figure}

\paragraph{Fugacity-correction magnitude.}
Using the canonical PR-EOS for hydrogen
($T_c = 33.19$~K, $P_c = 13.13$~bar, $\omega = -0.219$;
\cite{PengRobinson1976}, with the cubic compressibility equation
and the closed-form $\ln \phi$ expression detailed in
Supplementary Section~\ref{suppl:fugacity_details}), the
fugacity coefficient remains $\phi > 1$ across the operating
range owing to the negative acentric factor. At the
training-domain boundary (80~bar, 25--80$^{\circ}$C) $\phi$
ranges from $1.024$ to $1.029$ ($2$--$3\%$ deviation from
ideal gas), rising to $1.066$--$1.072$
(${\sim}7\%$ correction) at the extrapolation extreme
(200~bar, 25--80$^{\circ}$C; Fig.~\ref{fig:exp1_preos_residuals}e).
The $7\%$ fugacity correction at 200~bar shifts the predicted
$\Phi_{\mathrm{H_2}}$ by at most ${\sim}0.25$\%p, roughly one
quarter of the observed mean signed residual of the ideal-gas
backbone ($\mu_{\mathrm{res}}^{\mathrm{ideal}} = -0.935$\%p at
200~bar; Table~\ref{tab:mae_degradation}); the remaining
$\sim 70$--$75\%$ is consistent with non-fugacity high-pressure
phenomena (membrane swelling, LGDL compressibility,
electroosmotic drag) that the residual network must therefore
also represent. A per-component decomposition of the residual is
provided in Supplementary Section~\ref{suppl:fugacity_details}.

\paragraph{Empirical backbone-variant test.}
To evaluate this decomposition empirically, we implemented PR-Net
with a PR-EOS backbone in which the Henry's-law interface
concentration is replaced by
$c^*_{\mathrm{H_2}} = S_{\mathrm{H_2}}(T)\,\phi(T,P)\, p^{\mathrm{mem}}_{\mathrm{H_2}}$
while keeping the residual network unchanged, and trained it
under the same Protocol-IEP regime (cathode pressures
$\leq 80$~bar; ensemble $M = 100$). Under PR-Net the gap between
the two backbones collapses
(Table~\ref{tab:exp1_preos_comparison};
Fig.~\ref{fig:exp1_preos_residuals}):
$R^2_{200\,\mathrm{bar}} = 0.940$ (ideal gas) versus $0.944$
(PR-EOS), and the mean signed residuals remain very close
($\mu_{\mathrm{res}}^{\mathrm{ideal}} = -0.935$\%p,
$\mu_{\mathrm{res}}^{\mathrm{PR\text{-}EOS}} = -0.916$\%p;
$\sigma_{\mathrm{res}}^{\mathrm{ideal}} = 0.319$\%p,
$\sigma_{\mathrm{res}}^{\mathrm{PR\text{-}EOS}} = 0.284$\%p).
The near-identical pressure-stratified residual distributions
(Fig.~\ref{fig:exp1_preos_residuals}c) are consistent with the interpretation
that the residual network accounts for a fugacity-scale
component at the system level: adding the correction explicitly to
the backbone yields only a marginal change in end-to-end prediction,
and the persistent
negative bias common to both variants indicates that the
dominant remaining systematic deviation is more consistent with
non-fugacity high-pressure phenomena than with a purely
backbone-resolvable error component.

\paragraph{Design-choice interpretation.}
The near-equivalence of the two variants implies that, within
the present operating range, the choice between explicit PR-EOS
embedding and an ideal-gas backbone with a residual network is
a design trade-off rather than an accuracy trade-off. The
ideal-gas backbone is retained in the default PR-Net configuration because
(i) it preserves analytical simplicity and avoids per-evaluation
cubic-EOS root finding, which is consequential for the real-time
edge-deployment regime (Section~\ref{sec:edge_deployment}) and
for the inverse-design sweeps over the
$(P, T)$ safety and $i_{\min}$ grids
(Section~\ref{sec:economic_causal_chain}); and
(ii) it allows the residual network to represent non-ideality sources
jointly---fugacity, swelling, compressibility, and
apparatus terms---yielding a compact interpretable residual
signal that is subsequently mined for hypothesis-generating
mechanistic discovery
(Section~\ref{sec:mechanistic_discovery}). The PR-EOS variant
is documented as a drop-in alternative for users who prefer
explicit real-gas modelling; the inverse-design cost overhead it
incurs is quantified in
Supplementary Section~\ref{suppl:fugacity_details}.

\begin{table}[ht]
\centering
\caption{Backbone-variant comparison of PR-Net under the
Protocol-IEP extrapolation regime (train: $\leq 80$~bar, $n = 42$;
test: 120, 160, 200~bar, $n = 24$; Nafion~117, 25$^{\circ}$C;
ensemble $M = 100$). The two variants compared are PR-Net with the
ideal-gas backbone (the default configuration) and PR-Net with the
PR-EOS backbone,
i.e., Henry's-law interface concentration replaced by
$S_{\mathrm{H_2}}(T)\, \phi(T,P)\, p^{\mathrm{mem}}_{\mathrm{H_2}}$.
Residual statistics ($\mu_{\mathrm{res}}$, $\sigma_{\mathrm{res}}$)
refer to signed prediction error $\hat{y} - y$ at 200~bar
(\%p H$_2$).}
\label{tab:exp1_preos_comparison}
\resizebox{\textwidth}{!}{%
\begin{tabular}{llccccc}
\toprule
Model & Backbone
  & 120~bar $R^2$
  & 160~bar $R^2$
  & 200~bar $R^2$
  & $\mu_{\mathrm{res}}$ (200~bar, \%p)
  & $\sigma_{\mathrm{res}}$ (200~bar, \%p) \\
\midrule
PR-Net (ideal-gas backbone)
  & Ideal gas
  & $0.997$ & $0.967$ & $0.940$
  & $-0.935$ & $0.319$ \\
PR-Net (PR-EOS backbone)
  & PR-EOS
  & $0.998$ & $0.970$ & $0.944$
  & $-0.916$ & $0.284$ \\
\bottomrule
\end{tabular}}
\end{table}

\subsection{Autonomous Discovery of Unmodelled Transport Mechanisms
via Residual Analysis}
\label{sec:mechanistic_discovery}

Beyond resolving optimisation conflicts and enabling extreme
extrapolation, an important advantage of the PR-Net architecture lies
in its ability to support hypothesis-generating mechanistic
interpretation. By structurally decoupling the deterministic physics
baseline ($\Phi_{\mathrm{H_2}}^{\mathrm{phys}}$) from the neural
network residual ($\Delta y_{\mathrm{NN}}$), the framework makes
model discrepancy explicit and therefore provides a transparent window
into the epistemic gaps of conventional electrochemical models. In this
setting, the residual network is not interpreted as direct mechanistic
proof; rather, it isolates structured deviations from idealised physics
that can be examined as candidate signatures of unmodelled transport
phenomena.

\begin{figure}
\centering
\includegraphics[width=\textwidth]{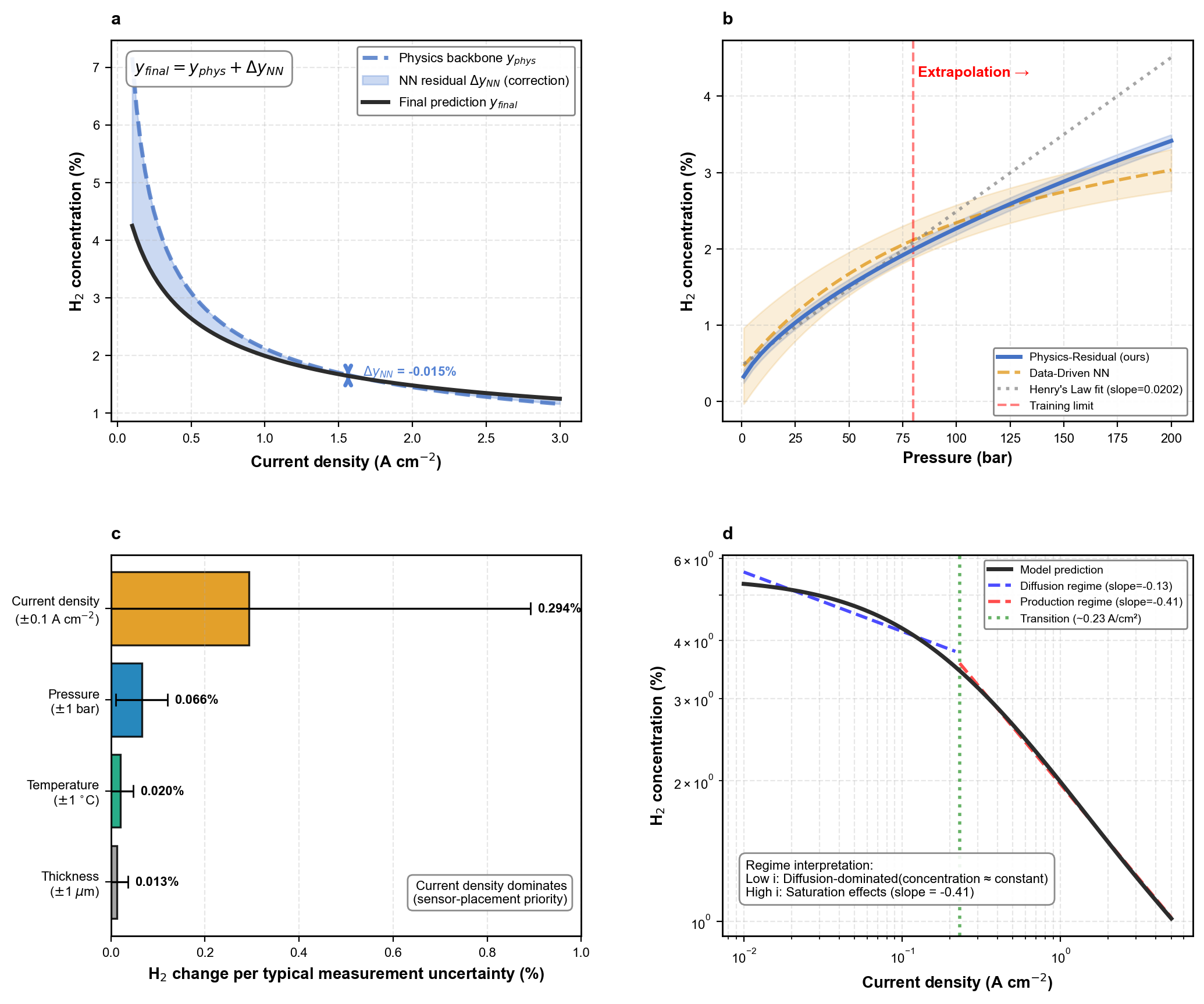}
\caption{Model interpretation and physics consistency of PR-Net.
(a) Prediction decomposition at 25$^{\circ}$C and 80~bar, showing
the final prediction ($\hat{y}_{\mathrm{final}}$), the deterministic
physics backbone ($\Phi_{\mathrm{H_2}}^{\mathrm{phys}}$), and the
residual correction ($\Delta y_{\mathrm{NN}}$).
(b) Pressure dependence of the physics backbone, showing near-linear
Henry's law scaling up to 200~bar (slope $\approx 0.0202$).
(c) Operational sensitivity of the prediction to each input under
typical measurement uncertainties (full analysis in
Section~\ref{sec:uncertainty}); bars show the H$_2$ change per
typical measurement uncertainty, with current density dominant.
Error bars denote standard deviation across operating conditions.
(d) Current-density dependence of the prediction, showing a transition
near 0.23~A~cm$^{-2}$ from a Fickian diffusion-dominated regime
(slope $\approx -0.13$) to a Faradaic production-dominated regime
(slope $\approx -0.41$).}
\label{fig:mechanistic_discovery}
\end{figure}

Fig.~\ref{fig:mechanistic_discovery}a deconstructs the additive
inference structure under representative Nafion~117 conditions
(25$^{\circ}$C, 80~bar): the final prediction is the sum of the
deterministic backbone $\Phi_{\mathrm{H_2}}^{\mathrm{phys}}$ and the
neural residual $\Delta y_{\mathrm{NN}}$, rendering the model's
reliance on physics versus data explicit and inspectable at every
operating point. Across the measured interpolation domain
($\leq 80$~bar) the residual acts as a small correction to a
backbone that already captures the dominant signal, so that within
the nominal operating envelope the deterministic model alone is
near-bias-free. Quantitatively, the temperature-binned mean residual
$\Delta y_{\mathrm{NN}} = y_{\mathrm{final}} - y_{\mathrm{phys}}$
contracts from $-0.86$\%p H$_2$ at 25$^{\circ}$C ($n=24$) and
$-0.79$\%p at 60$^{\circ}$C ($n=52$) to $-0.21$\%p at
80$^{\circ}$C ($n=75$) and $+0.12$\%p at 85$^{\circ}$C
($n=9$); the pooled 80--85$^{\circ}$C mean is $-0.17$\%p
($n=84$). This reduction should be interpreted as a near-zero
systematic residual (mean backbone bias), not as vanishing
pointwise residuals, because the 80$^{\circ}$C subset spans a broad
pressure--current range. The near-zero mean is most pronounced in
the industrially dominant 80--85$^{\circ}$C band, where the
idealised constitutive assumptions of the Henry--Fick--Faraday
backbone---in particular equilibrium membrane hydration and a
well-developed dissolved-gas concentration profile---are best
satisfied. The residual head accordingly contributes most
where the analytical physics is extended or stressed, especially
the high-pressure extrapolation regime, in which it
is consistent with accounting for part of the gas-phase non-ideality
decomposed quantitatively in
Section~\ref{sec:fugacity_correction}.

To identify which operating variables most strongly govern
hydrogen crossover, we quantified the model sensitivity to each
input under typical operational measurement uncertainties
(Fig.~\ref{fig:mechanistic_discovery}c; the full analysis is
reported in Section~\ref{sec:uncertainty}). Current density exerts
by far the dominant influence---several-fold larger than pressure
and more than an order of magnitude larger than temperature and
membrane thickness. The comparatively minor temperature
sensitivity is consistent both with the reduced systematic residual
in the industrial band and with the known weak temperature
dependence of hydrogen solubility in polymer membranes~\cite{bib44}. The
dominance of current density further carries direct implications
for sensor placement in distributed safety monitoring, detailed in
Section~\ref{sec:uncertainty}. The deployment-level consequences of
this accuracy at nominal conditions---safe-operating-envelope and
efficiency gains---are quantified in
Section~\ref{sec:economic_causal_chain}.

The thermodynamic consistency of the physics backbone in the
pressure domain underpins this residual decomposition: the strict
adherence to Henry's law linearity up to 200~bar
(Fig.~\ref{fig:mechanistic_discovery}b) indicates that pressure-
dependent contributions are largely accounted for by the analytical
backbone, isolating the current-density-driven and high-pressure
non-ideal deviations captured by $\mathrm{NN}_{\mathrm{residual}}$; see
Section~\ref{sec:extrapolation} for the full quantitative
extrapolation analysis.

Furthermore, without any explicit mechanistic programming or
supervision, the PR-Net identifies the non-linear
transition between macroscopic transport regimes
(Fig.~\ref{fig:mechanistic_discovery}d). The model detects a critical inflection point near 0.23~A~cm$^{-2}$. Below
this threshold, the concentration profile exhibits a shallow slope
($\approx -0.13$), accurately reflecting a Fickian
diffusion-dominated regime where the absolute crossover flux is
relatively constant but the relative concentration decreases as
oxygen production begins. Above this threshold, the system undergoes
a regime shift characterised by a steeper slope ($\approx -0.41$),
indicating a Faradaic production-dominated transport regime where
the dilution effect of rapidly evolving anodic oxygen becomes the
governing factor. This identification of the inflection
at 0.23~A~cm$^{-2}$ is consistent with the macroscopic transition
reported in the experimental literature \cite{bib4} and \cite{bib31}.

By capturing both the high-pressure gas-phase non-ideality
(Section~\ref{sec:fugacity_correction}) and the
current-density transport-regime transition, the PR-Net extends
beyond conventional predictive modelling towards a residual-based
reverse-engineering approach, in which
structured residuals generated by the machine-learning component can
directly inform and refine future theoretical equations for
multiphase electrochemical transport.

\subsection{Uncertainty Quantification and Operational Sensitivity
Analysis}
\label{sec:uncertainty}

Reliable uncertainty estimates are essential for risk assessment
in safety-critical applications. The 100-model PR-Net Deep Ensemble (Protocol-IEP),
trained exclusively on data $\leq 80$~bar ($n = 42$), demonstrated
well-calibrated confidence intervals: 94.6\% of experimental
observations in the extrapolation domain fell within the predicted
95\% confidence bounds, corresponding to an empirical coverage
probability (ECP) of $\mathrm{ECP} = 0.946$ and an absolute calibration gap
of 0.004 from nominal 95\% coverage. This indicates well-calibrated uncertainty quantification
in this extrapolation test. This calibration result is particularly
notable because it was obtained under extrapolation conditions
(120--200~bar) where the ensemble had no direct training signal,
suggesting that the physics backbone provides a useful predictive
anchor that helps reduce overconfident failure.

Sensitivity analysis (Fig.~\ref{fig:mechanistic_discovery}c) was
conducted by perturbing each input variable by typical operational
measurement uncertainties
($\pm 1^{\circ}$C, $\pm 1$~bar, $\pm 0.1$~A~cm$^{-2}$,
$\pm 1$~$\mu$m) and quantifying the resulting change in predicted
H$_2$ concentration. Current density exerts the dominant influence
on hydrogen crossover ($2.94 \pm 5.98$\% per A~cm$^{-2}$), with
the high standard deviation reflecting strong condition-dependent
non-linearity near the transport regime transition identified in
Section~\ref{sec:mechanistic_discovery}. Pressure shows moderate
sensitivity ($0.066 \pm 0.055$\%/bar), while temperature
($0.020 \pm 0.027$\%/$^{\circ}$C) and membrane thickness
($0.013 \pm 0.023$\%/$\mu$m) exhibit comparatively minor
influences, consistent with the known weak temperature dependence
of hydrogen solubility in polymer membranes \cite{bib44}.

These quantitative sensitivity rankings carry direct implications
for industrial sensor placement and measurement frequency. The
pronounced sensitivity to current density, coupled with its high
variability, underscores the importance of precise current
control for maintaining hydrogen purity in industrial PEM
electrolysers. Accordingly, current density measurement nodes
should be prioritised in distributed monitoring architectures,
followed by pressure sensors at high-pressure segments, while
thermal instrumentation may be deployed at lower spatial density
with limited loss of safety prediction fidelity.

\subsection{Real-Time Edge Deployment and Process Control
Integration}
\label{sec:edge_deployment}

The ultimate objective of phenomenological modelling in chemical
engineering is its successful translation into actionable process
control. While conventional computational fluid dynamics (CFD) and
rigorous mechanistic models provide deep mechanistic insights, their
computational overhead strictly precludes real-time application,
with typical runtimes on the order of minutes to hours per
simulation. The PR-Net architecture, comprising only 34,305
trainable parameters, was benchmarked across three representative
hardware platforms to validate its industrial applicability
(Table~\ref{tab:hardware_benchmarking}).

\begin{table}[ht]
\centering
\caption{Inference performance across hardware platforms. Each
platform executed 1,000 forward passes with random inputs; the first
100 were discarded as warm-up. The International Organization for
Standardization (ISO) 26142
$t_{90} \leq 30$~s threshold for stationary hydrogen detection is
included for reference \cite{iso26142}.}
\label{tab:hardware_benchmarking}
\begin{adjustbox}{max width=\textwidth}
\begin{tabular}{lccccc}
\hline
\textbf{Platform} & \textbf{CPU/GPU}
  & \textbf{RAM}
  & \textbf{Model Load (ms)}
  & \textbf{Single Inference (ms)}
  & \textbf{Batch-100 (ms)} \\
\hline
Desktop PC
  & Intel i7-13700 + RTX-4060
  & 48~GB & 125 & $0.31 \pm 0.04$ & 28.7 \\
Jetson AGX Orin
  & ARM A78AE + GPU
  & 64~GB & 340 & $1.36 \pm 0.03$ & 122.1 \\
Raspberry Pi 5
  & ARM Cortex-A76
  & 8~GB  & 1850 & $1.08 \pm 0.34$ & 99.8 \\
\hline
\multicolumn{5}{l}{ISO 26142 safety threshold:}
  & $t_{90} \leq 30{,}000$~ms \\
\hline
\end{tabular}
\end{adjustbox}
\end{table}

All three platforms achieved millisecond-level inference,
representing a 5--6 orders of magnitude improvement over full CFD
simulation runtimes. Even on the most resource-constrained edge CPU
(Raspberry Pi 5), the PR-Net achieves $1.08 \pm 0.34$~ms per
forward pass, comfortably satisfying the $t_{90} \leq 30$~s
response-time requirement specified by ISO 26142 for stationary
hydrogen detection systems \cite{iso26142}, with context from
\cite{hubert2013}. A conservative
three-sigma upper bound for this edge setting ($\mu + 3\sigma \approx
2.10$~ms) remains approximately $1.4\times10^4$ times faster than the
ISO threshold, supporting a substantial real-time safety margin under
latency variability. This
millisecond-level responsiveness supports rapid tracking of
crossover dynamics during load fluctuations, pressure transients,
and emergency shutdowns.

The compact architecture additionally supports a distributed
monitoring paradigm essential for large-scale electrolyser
installations. Multiple PR-Net inference nodes can be deployed
across individual stack segments, enabling spatially resolved
crossover monitoring without centralised computational
infrastructure. In the event of anomalous epistemic uncertainty
from the residual component, the architecture provides a built-in
fail-safe: the system can revert to predictions from the
deterministic physics backbone ($\Phi_{\mathrm{H_2}}^{\mathrm{phys}}$)
alone, ensuring physically consistent safety estimates even under
network failure conditions.

\paragraph{Online residual updating with a frozen physics backbone.}
A natural extension of the architecture is online updating: as new
operating data arrive, the residual network is fine-tuned on those
measurements while the calibrated, non-trainable backbone
$(\alpha_k, \beta_k)$ of Section~\ref{sec:prnet} stays fixed, so the
update is a transfer-learning step that never disturbs the physics layer. We assess this on the Martin et al.~(2022)
compression-varied dataset (Nafion~212, $n = 104$, PTL compression
10--85~$\mu$m)~\cite{Martin2022}, an operating regime absent from
training: the deployed PR-Net, applied zero-shot without any
fine-tuning, already reaches $R^2 = 82.9\%$, and fine-tuning the residual
on as few as five new points (few-shot size $k$) raises this to
${\sim}91\%$ and then to $98.6\%$ at $k = 20$ on held-out data;
using all 104 measurements gives $99.6\%$ as an apparent in-sample
full-data calibration endpoint rather than a held-out test. Full dataset
characteristics and the complete
per-$(\lambda, k)$ results are given in Supplementary
Section~\ref{suppl:online_update} (Supplementary
Tables~\ref{tab:s_martin2022} and~\ref{tab:s_online_update}). The
physics-regularisation weight $\lambda$, which penalises the residual magnitude (Fig.~\ref{fig:exp4_lambda_tradeoff}),
can be annealed as data accumulate---a moderate value regularises and
stabilises the scarce-data updates, whereas a small value improves
accuracy once data are ample, so the per-$k$ optimum in this experiment slides from
$\lambda \approx 0.5$ at $k = 5$ to a broad plateau at $\lambda \lesssim
0.1$ for $k \gtrsim 20$. Because only the residual network is updated,
the procedure is lightweight---fitting the compact 34,305-parameter
residual to at most ${\sim}100$ new points---and preserves compatibility
with the
calibrated physics layer used for the inverse-design queries of
Sections~\ref{sec:optimization_resolution} and
\ref{sec:economic_causal_chain}. Extending this mechanism to streaming,
temporally-resolved data and long-duration degradation effects is left
for future work.

\subsection{Techno-Economic Assessment and Industrial Deployment
Pathway}
\label{sec:economic_assessment}

The edge deployment capability established above is used here to
construct illustrative economic and safety scenarios at industrial
scale, rather than to claim realised plant-scale benefits.
Table~\ref{tab:economic_benefits} summarises the conditional
projected impact of integrating the PR-Net into closed-loop process
control systems. All projections are derived from prediction-output
translations and published cost benchmarks rather than direct field
measurements; their realisation requires integration with
closed-loop control systems validated under sustained dynamic
conditions. The underlying cost and policy benchmarks are drawn from
the global hydrogen outlook of the International Energy Agency (IEA)
\cite{IEA2019}, the alkaline and PEM electrolyser cost projections
of Krishnan et al.~\cite{Krishnan2023}, and the NREL manufacturing-cost
analysis for PEM water electrolysers by Mayyas et al.~\cite{NREL2019}.

\subsubsection{Quantitative scenario linkage between prediction
accuracy and economic outcomes}
\label{sec:economic_causal_chain}

To link the techno-economic figures of
Table~\ref{tab:economic_benefits} explicitly to the prediction-level
metrics without overstating operational causality, this subsection
presents a four-step accounting pathway that maps PR-Net's
prediction accuracy to plant-level economic scenarios under stated
assumptions. Each step is traceable to a
specific quantitative result from the extrapolation experiments
(Section~\ref{sec:extrapolation}) and four auxiliary
techno-economic analyses---EA-1 (maximum safe pressure), EA-2 (safe
operating window), EA-3 (early-warning accuracy), and EA-4
(residual-dispersion reduction and IEA-basis operating-expenditure (OPEX) recovery)---defined
in Supplementary Table~\ref{tab:ea_definitions} and reported in the
supplementary material; Fig.~\ref{fig:exp13_economic_chain} consolidates the
scenario-defining results that map prediction accuracy to operating
envelope estimates (panels a--b), recovered-efficiency estimates
(panel c), and plant-scale OPEX scenarios (panel d).

\paragraph{Step 1 — Prediction-level accuracy reduces residual
dispersion.}
In the high-pressure extrapolation regime (81--200~bar,
Nafion~117, $n = 24$), PR-Net attains a band signed-error standard
deviation $\sigma_{\mathrm{res}}^{\mathrm{PR\text{-}Net}} = 0.319$\%p~H$_2$,
a ${\sim}74\%$ reduction relative to the data-driven NN out-of-fold
baseline ($1.214$\%p) and a ${\sim}18\%$ reduction relative to the
globally pre-calibrated physics model (the Henry--Fick--Faraday
backbone with no residual head; $0.390$\%p); both $\sigma$ values use
the globally pre-calibrated backbone. Because this physics
model is the conventional
industrial incumbent, the operating-envelope and OPEX gains below
are referenced to PR-Net's improvement over it; the per-band
breakdown is provided in Supplementary
Section~\ref{suppl:economic_details}.

\paragraph{Step 2 — Reduced dispersion can widen the safe operating
envelope.}
Propagating the tighter residual SD through the
95\%-confidence upper bound on the 4\%~H$_2$ explosive threshold
gives scenario-level operating-envelope gains: at $T = 60^{\circ}$C
the maximum safe pressure rises from 94.5 to 103.7~bar
($+9.2$~bar; Fig.~\ref{fig:exp13_economic_chain}b, EA-1), and
at $P = 200$~bar, $T = 80^{\circ}$C the minimum current density
for safe partial-load operation drops from
$i_{\min} = 4.28$ to $3.78$~A~cm$^{-2}$ ($+69.6\%$
load-following window; EA-2). Both estimates are computed from
the Step~1 reduction in $\sigma$ using standard Gaussian
upper-bound propagation; the full per-$(P, T)$ grid is reported
in Supplementary Section~\ref{suppl:economic_details}.

\paragraph{Step 3 — Operating-envelope estimates are translated into
recovered-efficiency scenarios on an IEA cost basis.}
On the IEA~2019 hydrogen cost basis
(H$_2$ price $\$5.0$~kg$^{-1}$, capacity factor 0.7, plant
efficiency $\eta = 0.7$ \cite{IEA2019}), PR-Net's residual-SD
reduction over the physics model is converted to a band-resolved
efficiency-recovery scenario for Nafion~117 of $+9.0\%$ at 1--20~bar and
$+2.1\%$ at 81--200~bar (Fig.~\ref{fig:exp13_economic_chain}c,
EA-4). Summed over the bands in which PR-Net improves on the
physics model, this gives illustrative OPEX scenarios of
${\sim}\$71$k~yr$^{-1}$ at 1~MW, ${\sim}\$0.71$M~yr$^{-1}$ at
10~MW, and ${\sim}\$7.1$M~yr$^{-1}$ at 100~MW, bracketing the
$\$200$k--$\$1.5$M~yr$^{-1}$ range quoted in
Table~\ref{tab:economic_benefits} at representative
($\approx 10$~MW) plant scale.

\paragraph{Step 4 — Extrapolation robustness reduces observed
false-negative safety alerts in the hold-out test.}
The binary risk-classification error at the 3.5\%~H$_2$
early-warning threshold falls from $23.8$--$28.6\%$ for the
data-driven NN and soft-constraint PINN baselines to $0.0\%$
for PR-Net in the extrapolation hold-out (EA-3). When combined
with the industrial benchmark of $\$50$k--$\$200$k per unplanned
PEMWE shutdown, this result provides a scenario basis for the
``incident-prevention'' row of Table~\ref{tab:economic_benefits},
but it should not be interpreted as a measured reduction in field
incident rates.

Together, EA-1--4 furnish a four-step scenario pathway from
prediction-level accuracy (MAE, residual $\sigma$) to operating
envelope estimates ($P_{\max}$, $i_{\min}$) to plant-level OPEX
scenarios (IEA~2019 basis). The Table~\ref{tab:economic_benefits}
figures should accordingly be read as assumption-dependent
scenarios traceable to model outputs, not as ground-truth
operational measurements.

\begin{figure}
\centering
\includegraphics[width=0.95\textwidth]{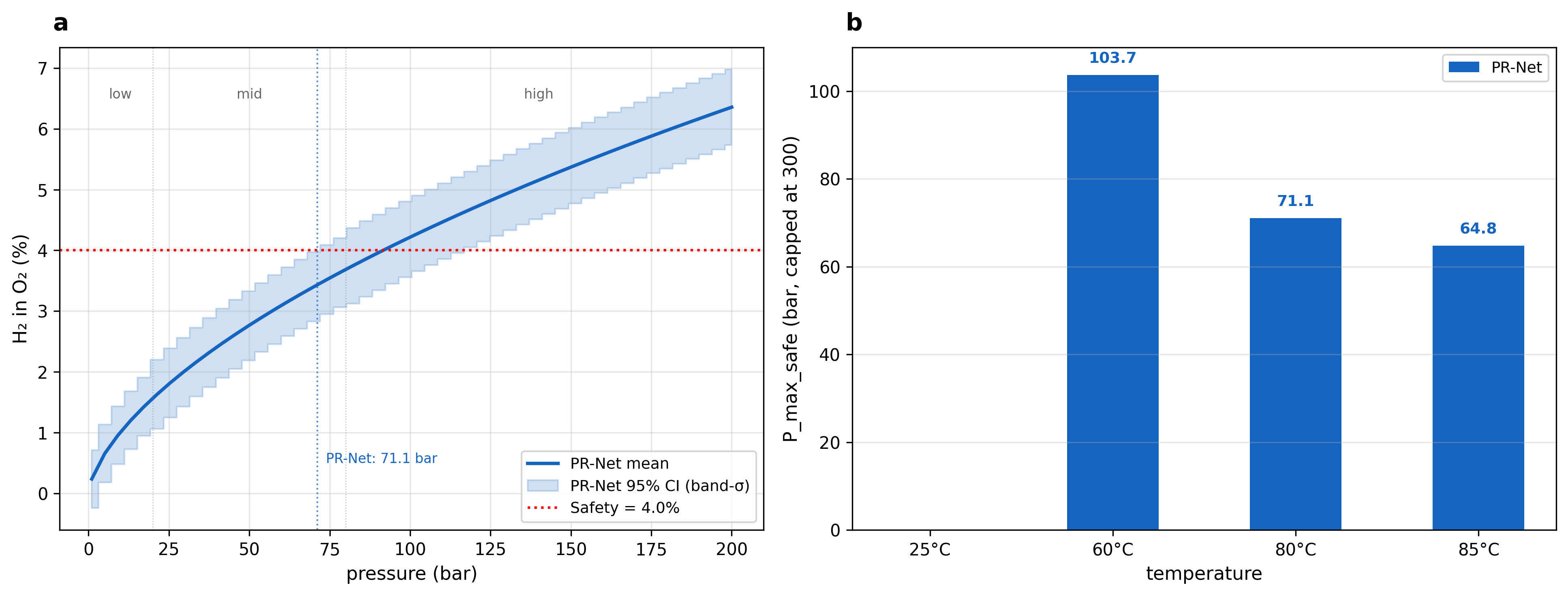}\\[2pt]
\includegraphics[width=0.95\textwidth]{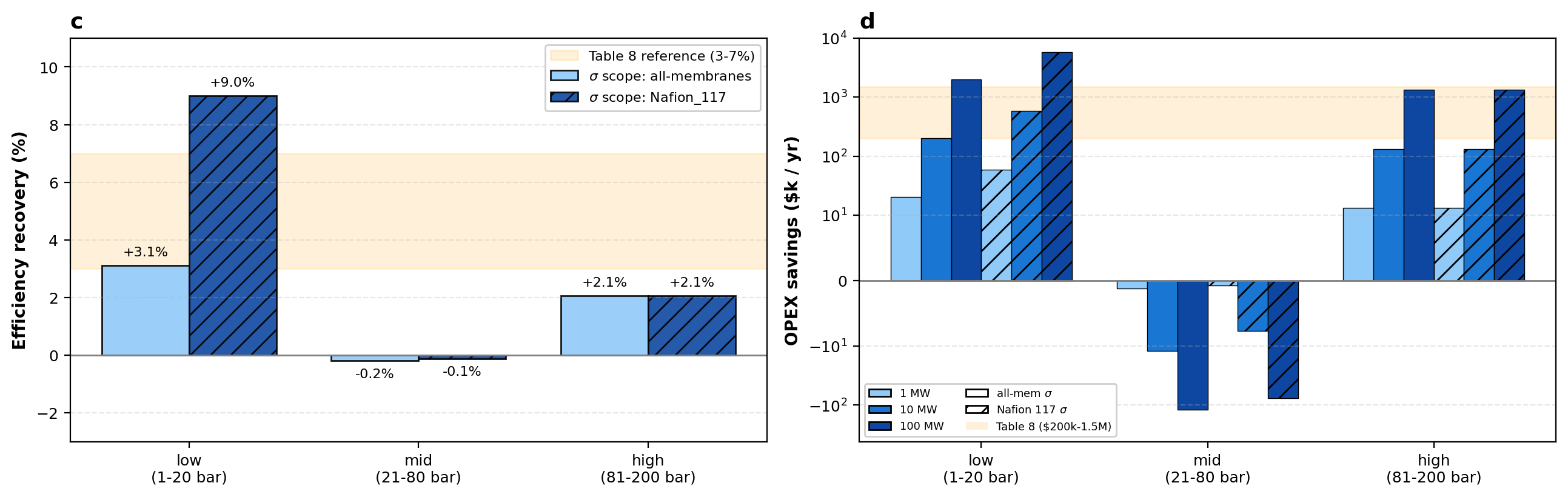}
\caption{Scenario pathway from PR-Net prediction accuracy to
techno-economic estimates (EA-1 and EA-4).
(a) Pressure-resolved hydrogen crossover prediction at
$T = 80^{\circ}$C on Nafion~117 with the PR-Net 95\%-confidence band
($i = 1.5$~A~cm$^{-2}$); the 4\%~H$_2$ explosive threshold and the
inferred maximum safe pressure $P_{\max}$ are highlighted.
(b) Maximum safe pressure $P_{\max}$ as a function of operating
temperature on Nafion~117 (EA-1); PR-Net (residual-SD-based bound)
extends the safe envelope by $+9.2$~bar at $T = 60^{\circ}$C relative
to the calibrated Henry--Fick--Faraday backbone alone.
(c) Assumption-dependent recovered-electrical-efficiency scenario by
pressure band (EA-4), obtained from the band-resolved reduction in
residual SD against the physics model; the scenario recovery is
$+9.0$\% (1--20~bar) and $+2.1$\% (81--200~bar) on a
Nafion~117 basis.
(d) Illustrative, not field-measured, OPEX scenarios mapped from panel (c) onto the IEA~2019
hydrogen cost model
($\$5.0$~kg$^{-1}$~H$_2$, CF~$=0.7$, $\eta = 0.7$) for 1, 10, and
100~MW plants. The four panels together define the engineering
assumptions linking Steps 1--2 (envelope) and Step 3 (OPEX) of the
scenario pathway in Section~\ref{sec:economic_causal_chain}.}
\label{fig:exp13_economic_chain}
\end{figure}

\begin{table}[htbp]
\centering
\caption{Illustrative economic and safety scenarios for PR-Net
implementation in industrial PEM water electrolysis. Values are
derived from prediction-output analyses EA-1--4
(Fig.~\ref{fig:exp13_economic_chain}) and published cost benchmarks;
they are assumption-dependent scenarios anchored to IEA~2019 hydrogen
cost benchmarks rather than direct field measurements.}
\label{tab:economic_benefits}
\resizebox{\textwidth}{!}{%
\begin{tabular}{@{}llp{8cm}@{}}
\toprule
\textbf{Benefit Category}
  & \textbf{Scenario Value}
  & \textbf{Key Engineering Drivers} \\
\midrule
\multicolumn{3}{@{}l}{\textit{\textbf{Operational \& Economic
Impact}}} \\
Illustrative annual OPEX scenario
  & \$200,000--1,500,000/yr
  & Assumed improvement in maintenance scheduling and reduced
    unplanned-downtime exposure via real-time H$_2$ crossover monitoring. \\
Illustrative system efficiency gain
  & 3--7\% improvement
  & Adaptive current density and pressure control during renewable
    energy transients; electricity constitutes 60--80\% of total
    hydrogen production OPEX \cite{IEA2019}. \\
Potential stack CAPEX deferral
  & \$100,000--500,000/cycle
  & Scenario based on 15--25\% stack lifetime extension via predictive
    maintenance and earlier detection of elevated crossover
    \cite{Krishnan2023}. \\
\midrule
\multicolumn{3}{@{}l}{\textit{\textbf{Process Safety
Management}}} \\
Incident-prevention scenario
  & Mitigation of undetected leaks
  & Real-time monitoring could support control actions that keep H$_2$
    concentration below the 4\% explosive threshold; hydrogen safety
    sensors are widely recognised as critical elements of safe
    hydrogen system deployment \cite{bib45}. \\
\bottomrule
\end{tabular}%
}
\end{table}

From an OPEX perspective, electricity consumption dominates,
constituting 60--80\% of total hydrogen production costs
\cite{IEA2019}. Traditional safety protocols mandate conservative
operating boundaries to prevent H$_2$ concentration from
approaching the 4\% explosive limit. By providing real-time
crossover predictions, the PR-Net could support adaptive-control
studies in which current density and differential pressure are
adjusted closer to efficiency-favourable setpoints during periods
of high renewable energy availability, subject to closed-loop
validation. Under the assumptions in
Table~\ref{tab:economic_benefits}, this dynamic optimisation
corresponds to an illustrative 3--7\% overall
system-efficiency scenario. The present model is trained on static
experimental observations and does not incorporate time-series
ageing trajectories; quantitative efficiency projections therefore
await future work combining PR-Net's real-time accuracy with
dedicated time-resolved degradation datasets.

Regarding capital expenditure (CAPEX), prolonged exposure to elevated crossover rates
accelerates membrane thinning and localised thermal degradation.
Continuous crossover monitoring via distributed PR-Net edge
sensors could facilitate predictive-maintenance strategies by
flagging cells with elevated crossover before severe degradation.
Stack replacement costs for PEM systems have been reported in the
range of
\$100,000--500,000 per cycle \cite{Krishnan2023}; extending stack
lifetime by 15--25\% through targeted maintenance could defer a
significant fraction of these costs, though the specific magnitude
depends on installation scale, maintenance baseline, and control
architecture.

Real-time hydrogen detection and safety sensing
are widely recognised as critical elements of safe hydrogen system
deployment \cite{bib45}. The deterministic structure of the
PR-Net physics backbone, combined with millisecond-level edge
execution and the extrapolation robustness demonstrated in
Section~\ref{sec:extrapolation}, motivates a staged deployment
pathway for evaluating safe industrial use in green hydrogen
infrastructure. A three-stage deployment roadmap is envisaged:
(1) static crossover prediction for real-time safety monitoring
(present work); (2) integration with closed-loop adaptive current
control during renewable energy transients; and
(3) incorporation of time-resolved membrane degradation models to
enable predictive lifetime management.


\section{Conclusions}
\label{sec:conclusions}

This study addresses safety and predictive limitations of
conventional data-driven NNs and soft-constraint PINNs
in high-pressure PEM water electrolysis. It also addresses the
previously unmet need for reliable data-driven prediction of hydrogen
crossover concentration as a direct safety target and, to the best of
our knowledge, constitutes the first systematic study in this domain
to place purely data-driven NN, soft-constraint PINN, and
hard-constraint physics-integrated modelling within a single
comparative framework.
The proposed hard-constraint PR-Net integrates Henry's, Fick's, and
Faraday's laws with deep learning in a unified architecture.
Evaluation across six ionomer membranes and a broad operating domain
(1--200~bar, 25--85$^{\circ}$C; 184 points from eight peer-reviewed sources)
supports five contributions to chemical engineering and process
safety:

\begin{enumerate}

  \item \textbf{First Systematic Benchmark for PEMWE Hydrogen-Crossover Prediction.}
  This study presents a unified comparative benchmark for PEMWE
  hydrogen crossover prediction across purely data-driven NN,
  soft-constraint PINN, and hard-constraint physics-integrated approaches under a common
  dataset, protocol, and prospective pressure-axis extrapolation setting,
  revealing performance tiers that were previously uncharacterised in the PEMWE literature.
  \item \textbf{Stable and Physics-Consistent Prediction Under Limited Experimental Data.}
  By embedding theoretical transport equations as a deterministic
  backbone, the PR-Net reduces the optimisation trade-off
  observed in conventional soft-constraint PINNs. This
  hard-constraint design constrains the neural network to learn systematic
  residuals, with improved training stability and
  interpolation accuracy ($R^2 = 99.57 \pm 0.16$\%) while reducing
  prediction variability by 9-fold compared to purely data-driven
  NN ($\mathrm{CV}_{R^2} = 0.16$\% versus 1.44\%).

  \item \textbf{Pressure-Axis Extrapolation Performance.}
  Under a rigorous zero-data-leakage protocol in which
  backbone parameters are calibrated exclusively from
  training-domain data ($\leq$80~bar), the hard-constraint
  architecture maintained $R^2 = 94.02 \pm 0.92\%$ at 200~bar---a
  2.5-fold extension beyond the training pressure range---compared with
  $68.06 \pm 5.52\%$ for the soft-constraint PINN and
  $58.00 \pm 8.60\%$ for the purely data-driven NN, with all differences
  statistically significant ($p < 0.001$). Extrapolation RMSE was
  reduced by 62\% relative to the purely data-driven NN and 57\%
  relative to the soft-constraint PINN at 200~bar.

  \item \textbf{Residual-Based Mechanistic Insight.}
  Through residual analysis, the framework extends
  standard machine-learning prediction towards a diagnostic
  tool for mechanistic interpretation. In the high-pressure
  extrapolation regime the residual behaviour is consistent with
  accounting for part of the gas-phase non-ideality omitted by the
  ideal-gas backbone (decomposed in
  Section~\ref{sec:fugacity_correction}), and---without explicit
  mechanistic programming---the model identifies a macroscopic
  transition between Fickian diffusion-dominated and Faradaic
  production-dominated transport regimes near 0.23~A~cm$^{-2}$,
  indicating that structured residuals can provide candidate
  mechanistic explanations for future electrochemical transport
  modelling.

  \item \textbf{Industrial Edge Deployment.}
  With only 34,305 trainable parameters, the model achieved
  a computation time of $1.08 \pm 0.34$~ms on low-power embedded
  hardware (Raspberry Pi 5), representing a 5--6-order-of-magnitude reduction relative to representative full CFD runtimes. This
  millisecond-level responsiveness is well within the
  $t_{90} \leq 30$~s response-time requirement specified by ISO
  26142~\cite{iso26142} for stationary hydrogen detection and is compatible with practical deployment for adaptive process
  control, predictive stack maintenance, and dynamic operational
  optimisation in industrial green hydrogen facilities.

\end{enumerate}

Several limitations warrant explicit acknowledgement.
First, the present model is trained on static,
steady-state observations and does not capture
time-resolved membrane ageing trajectories;
quantitative lifetime projections therefore require
future integration with dedicated degradation datasets.
The frozen-backbone design is well suited to this via online
residual updating (Section~\ref{sec:edge_deployment}); extending that
mechanism to streaming, temporally-resolved data and coupling it with
degradation trajectories is a natural route to continual field
adaptation.
Second, the default physics backbone assumes ideal-gas behaviour;
at cathode pressures routinely exceeding 150~bar,
explicit fugacity corrections via a real-gas equation
of state (e.g., Peng--Robinson~\cite{PengRobinson1976}) may
be incorporated into the backbone when thermodynamic
self-consistency is prioritised.
Third, the training dataset deliberately excludes
platinum-interlayer membranes to maintain dataset
homogeneity; extending the framework to
recombination-layer membrane designs~\cite{bib36}
constitutes a planned architectural expansion
in future work.
Fourth, while our evaluation provides evidence of pressure-axis extrapolation
to high-pressure thermodynamic regimes beyond the training range ($\ge$ 120~bar), the extrapolation
protocol combines low-pressure training data from multiple sources---Wu et al.
(20, 40, and 80~bar) plus additional low-pressure observations from Grigoriev et al.
(1~bar) and Schalenbach et al. (6~bar)---whereas the high-pressure test data
(120, 160, and 200~bar) come exclusively from Wu et al.~\cite{bib39}. Thus, despite
including partial low-pressure source diversity, the high-pressure extrapolation evaluation may
still retain source/apparatus-specific effects at high pressure. To partially address this
concern, we conducted feasible LOSO evaluations on four source-holdout settings
(St\"{a}hler et al.~(2020), Garbe et al.~(2019), Grigoriev et al.~(2011), and Schalenbach et al.~(2013);
Supplementary Table~\ref{tab:s_loso_summary}). The full source-wise feasibility rationale
for LOSO inclusion is provided in Supplementary Table~\ref{tab:s_loso_feasibility}. Across
these source-holdout evaluations, PR-Net consistently reduced RMSE and MAE relative to the soft-constraint PINN
and the purely data-driven NN. Nevertheless, explicitly verifying comprehensive inter-laboratory
generalisation via LOSO across all sources remains constrained by
dataset coverage and is an important future direction.

Ultimately, structurally merging deterministic
physical laws with residual machine learning---illustrated here for
hydrogen crossover prediction in PEM electrolysers---provides a
useful template for hybrid physics--machine learning
modelling across electrochemical energy systems where mechanistic
transport models exist but require data-driven refinement of
unmodelled interfacial phenomena. Under the present feasible LOSO and
pressure-axis extrapolation evidence, this approach remains a
practical path towards safer and more efficient green-hydrogen
deployment, while broader inter-laboratory validation is a logical
next step.


\clearpage

\appendix
\section{Supplementary Information}
\label{sec:supplementary_information}

\renewcommand{\thefigure}{S\arabic{figure}}
\renewcommand{\thetable}{S\arabic{table}}
\renewcommand{\theequation}{S\arabic{equation}}
\renewcommand{\thealgorithm}{S\arabic{algorithm}}
\setcounter{figure}{0}
\setcounter{table}{0}
\setcounter{equation}{0}
\setcounter{algorithm}{0}

\subsection{Comprehensive Exploratory Data Analysis}
\label{suppl:eda}

To ensure the PR-Net's robustness against inter-laboratory measurement
variances and systematic equipment biases, the training and validation
datasets were compiled from eight independent peer-reviewed studies
($n = 184$). This section provides a comprehensive exploratory data
analysis to characterise the distribution, inter-parameter
correlations, and study-specific variances inherent in this
meta-analytical dataset.

\begin{figure}
\centering
\includegraphics[width=\textwidth]{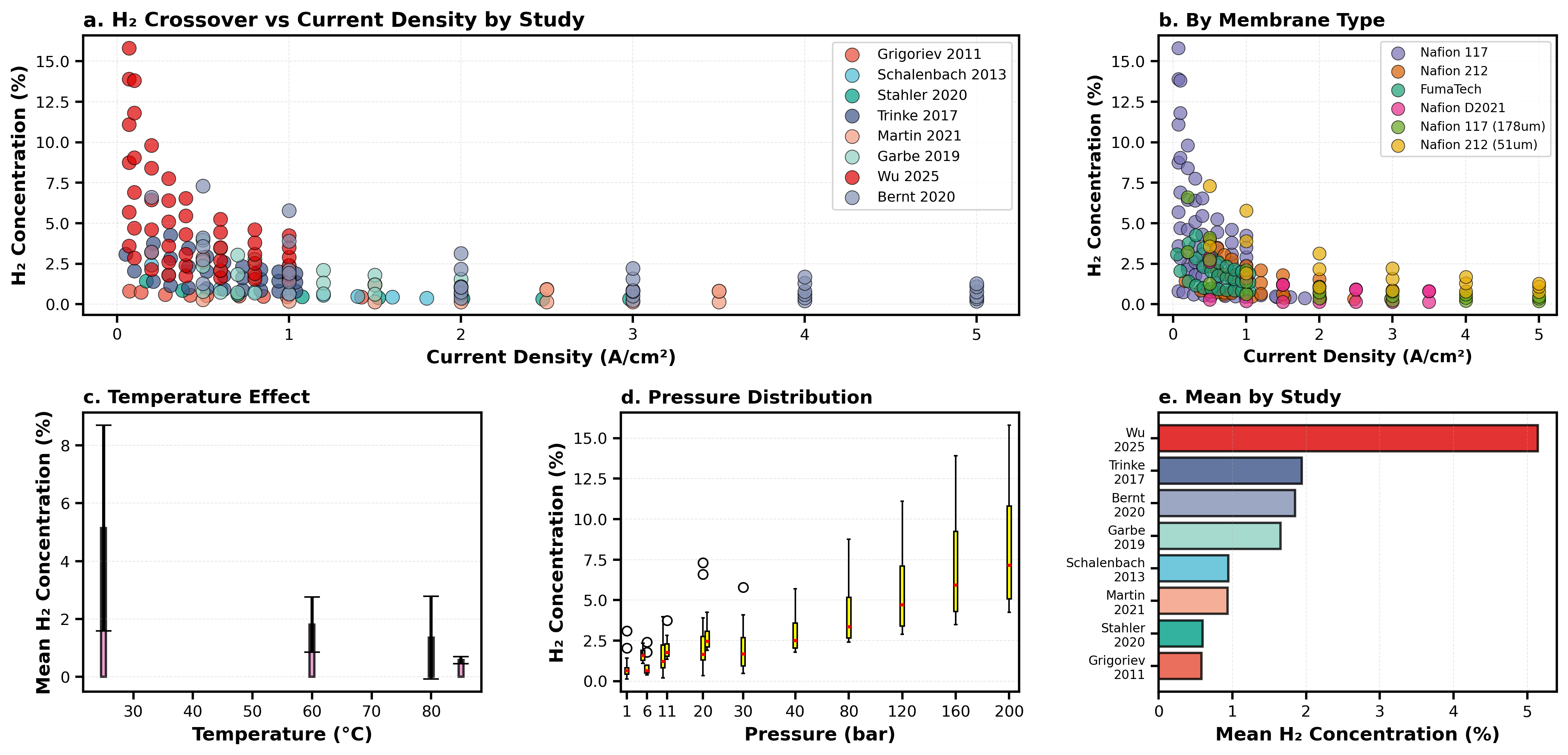}
\caption{Compiled experimental dataset characterisation:
operating condition coverage and study-specific crossover
distributions ($n = 184$).
(a) Relationship between H$_2$ crossover concentration and
current density across eight independent studies.
(b) Membrane-specific H$_2$ crossover behaviour for six
membrane types under varying current densities (0.05--5.00~A~cm$^{-2}$).
(c) Temperature dependence of mean H$_2$ concentration
across 25--85$^{\circ}$C. Error bars represent $\pm$1~s.d.
(d) Distribution of H$_2$ concentration as a function of
operating pressure (1--200~bar).
(e) Study-specific mean H$_2$ crossover concentrations.
Data compiled from Grigoriev et al.~(2011), Schalenbach et al.\
(2013), St\"{a}hler et al.~(2020), Trinke et al.~(2017),
Martin et al.~(2021), Garbe et al.~(2019), Wu et al.~(2025),
and Bernt et al.~(2020).}
\label{fig:s1_dataset1}
\end{figure}

\begin{figure}
\centering
\includegraphics[width=\textwidth]{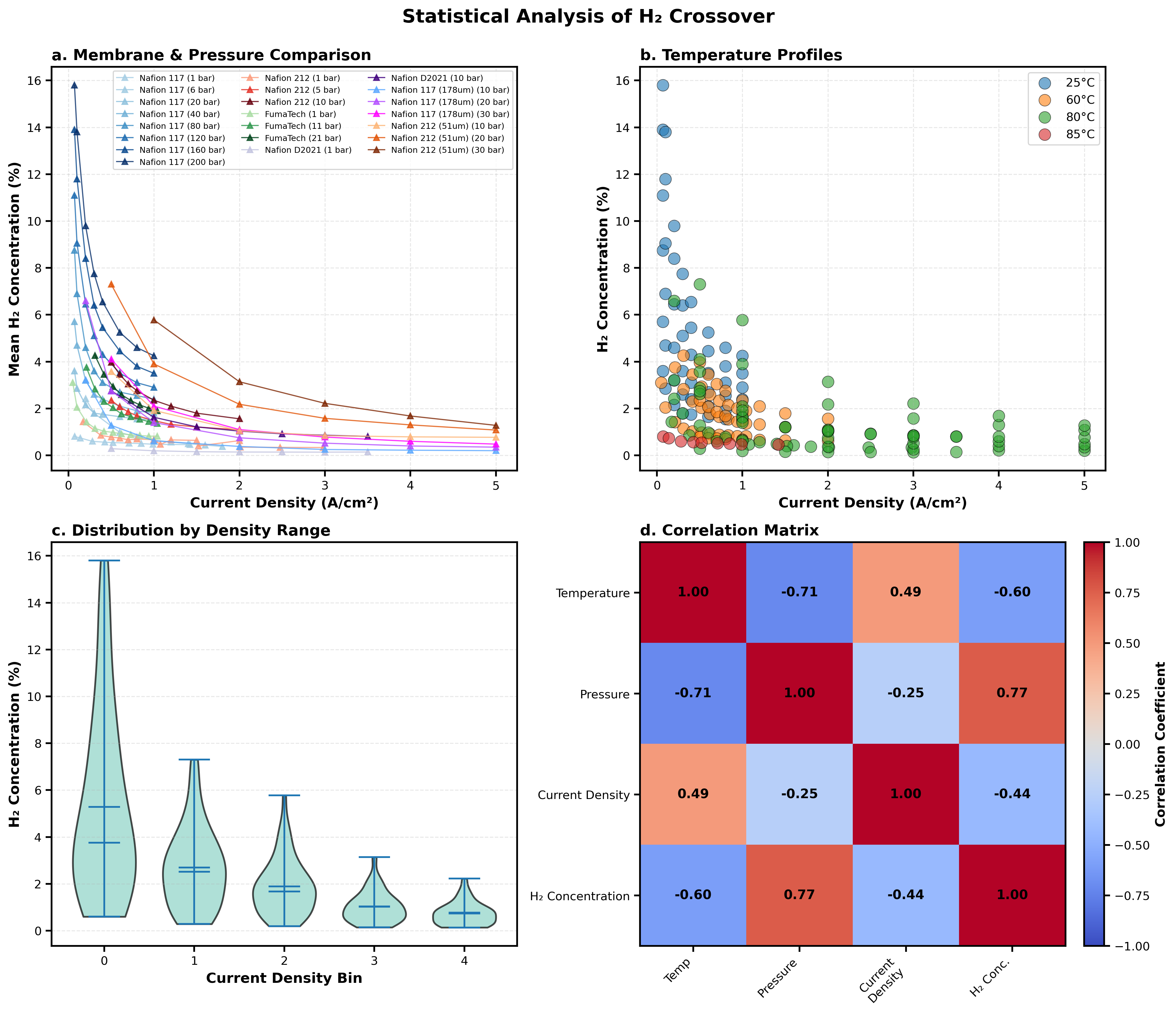}
\caption{Membrane-type crossover trends, temperature-dependent
concentration profiles, and inter-parameter correlation structure.
(a) Comparative analysis of mean H$_2$ crossover trends for
six membrane types and pressures as a function of current density (0--5~A~cm$^{-2}$).
(b) Temperature-dependent H$_2$ concentration profiles at
four operating temperatures (25, 60, 80, and 85$^{\circ}$C).
(c) Violin plots illustrating the concentration distribution
within discrete current density intervals.
(d) Pearson correlation matrix depicting pairwise
relationships between key operational parameters and H$_2$ crossover
concentration. Pressure demonstrates the strongest positive
correlation ($r = 0.77$); temperature and current density exhibit
moderate negative associations ($r = -0.71$ and $r = -0.44$,
respectively).}
\label{fig:s2_dataset2}
\end{figure}

\begin{figure}
\centering
\includegraphics[width=\textwidth]{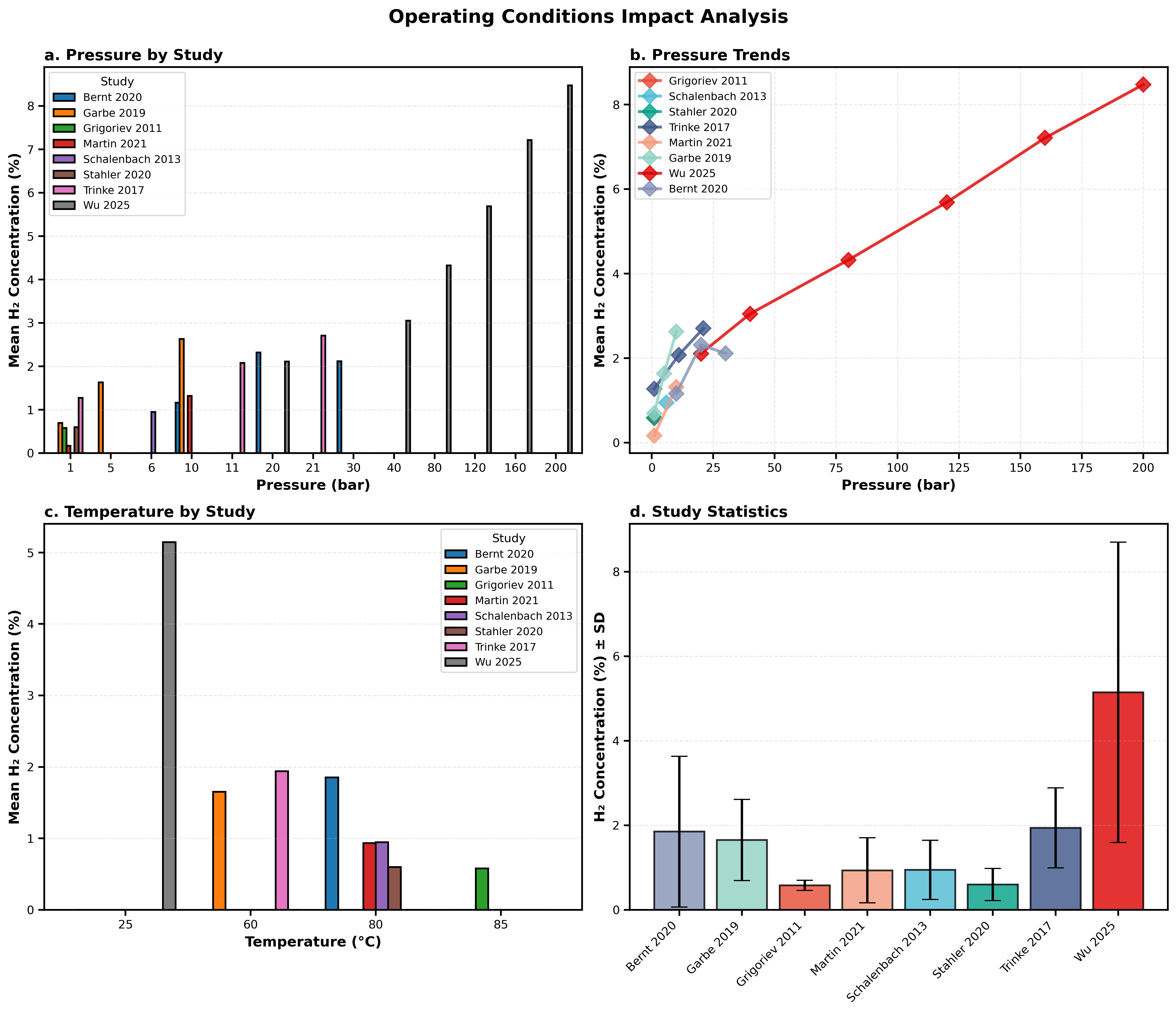}
\caption{Study-stratified pressure sensitivity and thermal
response across the compiled dataset.
(a, b) Study-specific mean H$_2$ concentration at discrete
cathode pressure conditions (1--200~bar), revealing marked inter-study
variability in pressure sensitivity.
(c) Temperature-dependent H$_2$ concentration profiles at
four operating temperatures.
(d) Study-stratified mean H$_2$ crossover concentrations
with $\pm$1~s.d.\ error bars. Per-study statistics: Bernt et al.\
2020 ($n = 37$, $1.85 \pm 1.78$\%), Garbe et al.~2019 ($n = 24$,
$1.65 \pm 0.96$\%), Grigoriev et al.~2011 ($n = 9$,
$0.58 \pm 0.12$\%), Martin et al.~2021 ($n = 21$,
$0.94 \pm 0.77$\%), Schalenbach et al.~2013 ($n = 9$,
$0.95 \pm 0.70$\%), St\"{a}hler et al.~2020 ($n = 8$,
$0.60 \pm 0.38$\%), Trinke et al.~2017 ($n = 28$,
$1.94 \pm 0.95$\%), Wu et al.~2025 ($n = 48$,
$5.14 \pm 3.55$\%).}
\label{fig:s3_dataset3}
\end{figure}

\begin{figure}
\centering
\includegraphics[width=\textwidth]{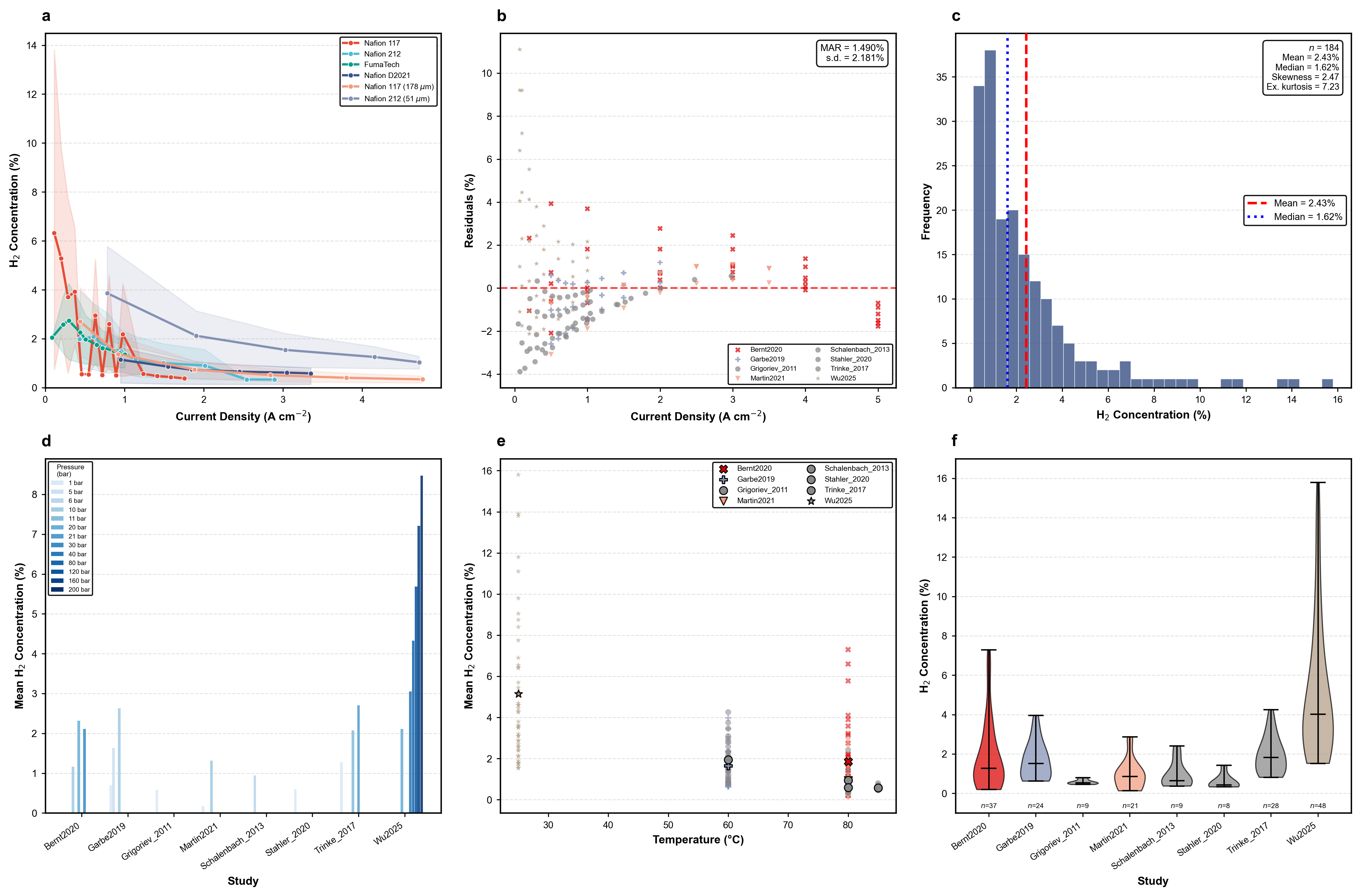}
\caption{Membrane-specific crossover performance envelopes
and inter-study variability characterisation ($n = 184$).
(a) Membrane-specific H$_2$ crossover performance envelopes
with min--max shaded regions.
(b) Residuals from a quadratic polynomial fit characterising
non-linear variability and inter-study scatter
(MAR $= 1.490$\%, s.d.\ $= 2.181$\%).
(c) Histogram of the overall H$_2$ concentration distribution
(mean $= 2.43$\%, median $= 1.62$\%, skewness $= 2.47$).
(d) Study-stratified mean H$_2$ concentration at each
discrete cathode pressure level.
(e) Individual H$_2$ measurements as a function of
temperature for each contributing study.
(f) Violin plots of H$_2$ concentration distributions per
study.}
\label{fig:s4_dataset4}
\end{figure}

As illustrated in Figs.~\ref{fig:s1_dataset1} and
\ref{fig:s2_dataset2}, the dataset captures a wide spectrum of
operational states, avoiding the artificial uniformity typically found
in single-laboratory datasets. Fig.~\ref{fig:s3_dataset3} highlights
the study-stratified pressure sensitivity, demonstrating the
heterogeneous nature of high-pressure responses across different
experimental setups. Fig.~\ref{fig:s4_dataset4} details the
non-linear variability and inter-study scatter, indicating that the
dataset provides a useful environment to stress-test
the physical consistency of the evaluated modelling architectures.

\clearpage

\subsection{Extended Model Evaluation Metrics}
\label{suppl:extended_metrics}

\begin{figure}
\centering
\includegraphics[width=\textwidth]{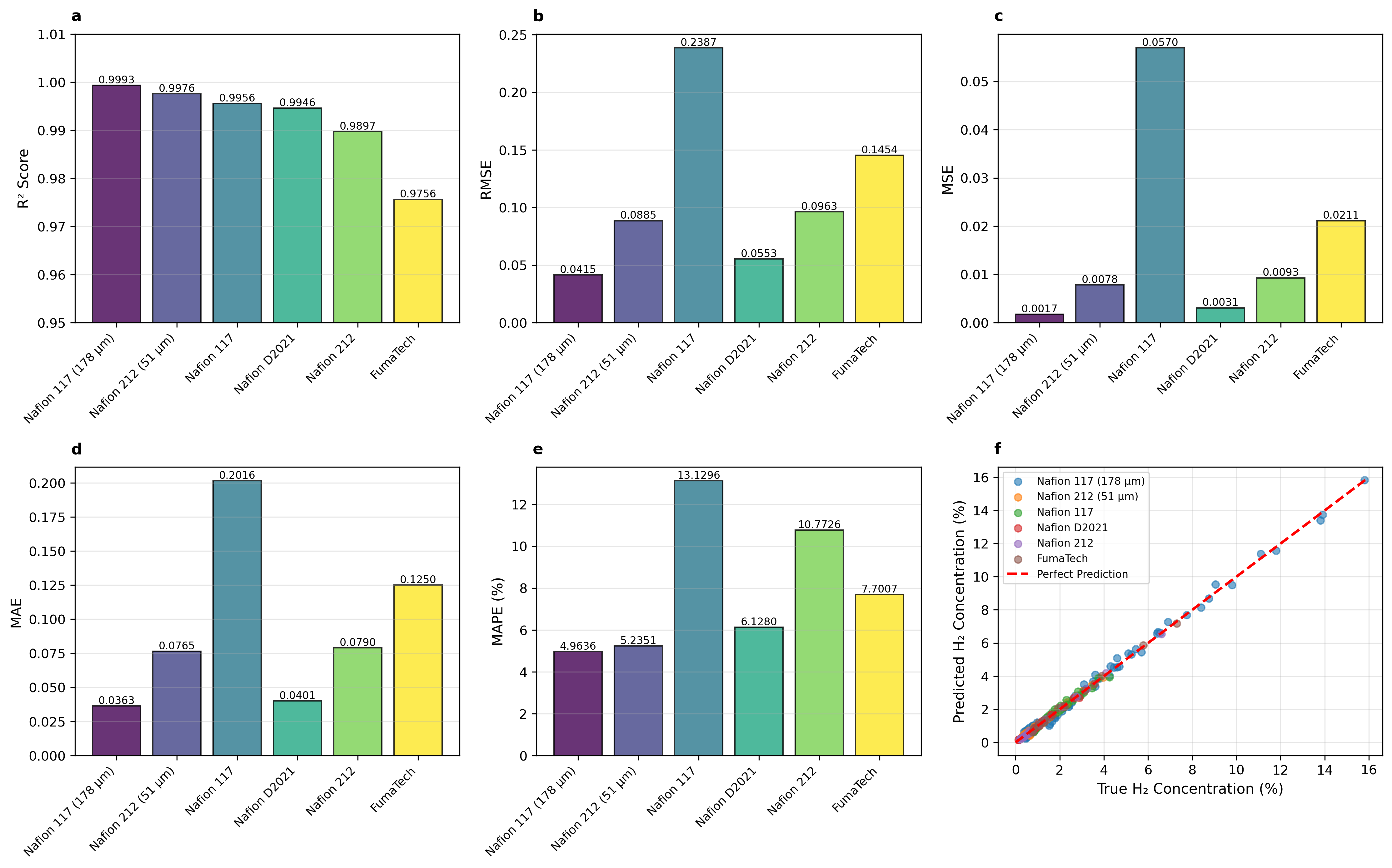}
\caption{PR-Net predictive accuracy across all six membrane
types evaluated via repeated cross-validation ($n = 184$).
(a--e) Membrane-stratified performance metrics ($R^2$, RMSE,
MSE, MAE, and MAPE) derived from out-of-fold predictions (20
repeats $\times$ 5 folds; $n_{\mathrm{models}} = 100$). $R^2 \geq
0.975$ for all membrane types; Nafion~117 (178~$\mu$m) and Nafion~212 (51~$\mu$m) achieve the highest $R^2$ (0.9993 and 0.9976,
respectively).
(f) Parity plot of PR-Net OOF predictions versus experimental
measurements across all six membrane types (overall $R^2 = 0.9960$),
supporting structural generalisation within the evaluated membrane set.}
\label{fig:s5_membrane_metrics}
\end{figure}

This section provides the disaggregated, membrane-specific predictive
performance metrics supporting the aggregated results presented in the
main text (Section~\ref{sec:optimization_resolution}).
Fig.~\ref{fig:s5_membrane_metrics} details the OOF
cross-validation predictions for the PR-Net across all six
structurally distinct ionomer membranes. The consistent clustering
along the parity line across a 40-fold range in crossover
concentration supports the interpretation that the PR-Net's hard-constraint architecture
generalises across the evaluated materials without obvious overfitting to specific material
artefacts or operating conditions. Detailed numerical benchmarks
across cross-validation and extrapolation settings are summarised in
Tables~\ref{tab:s_detailed_interp} and \ref{tab:s_detailed_extrap}.

\begin{table}[htbp]
\centering
\caption{Detailed interpolation performance metrics (mean $\pm$ s.d.)
derived from 100 independent cross-validation models per architecture.
Extends the summary provided in the main text
(Table~\ref{tab:interpolation}).}
\label{tab:s_detailed_interp}
\begin{tabular}{lccccc}
\toprule
\textbf{Method}
  & \textbf{$R^2$ (\%)}
  & \textbf{RMSE (\%p)}
  & \textbf{MAE (\%p)}
  & \textbf{MAPE (\%)}
  & \textbf{CV$_{R^2}$ (\%)} \\
\midrule
PR-Net
  & $99.57 \pm 0.16$
  & $0.166 \pm 0.026$
  & $0.124 \pm 0.019$
  & $9.80 \pm 1.71$
  & $0.16$ \\
Soft PINN
  & $98.86 \pm 1.43$
  & $0.248 \pm 0.116$
  & $0.165 \pm 0.074$
  & $14.43 \pm 8.39$
  & $1.43$ \\
Data-driven NN
  & $98.82 \pm 1.44$
  & $0.254 \pm 0.119$
  & $0.165 \pm 0.082$
  & $14.66 \pm 8.86$
  & $1.44$ \\
Physics-only
  & $99.47$
  & $0.193$
  & $0.140$
  & $10.93$
  & --- \\
\bottomrule
\end{tabular}
\end{table}

\begin{table}[htbp]
\centering
\caption{Aggregate predictive accuracy of the three
architectures over the full extrapolation domain
(120--200~bar) ($n = 24$; Nafion~117,
25$^{\circ}$C). Protocol-FCP values are provided in
Table~\ref{tab:extrapolation_r2} for comparison.}
\label{tab:s_detailed_extrap}
\begin{tabular}{lccc}
\toprule
\textbf{Model} & \textbf{MAE mean (\%p)}
  & \textbf{MAE s.d. (\%p)}
  & \textbf{Overall $R^2$ (\%)} \\
\midrule
PR-Net        & 0.539 & 0.429 & 96.51 \\
Soft PINN & 1.185 & 1.116 & 80.35 \\
Data-driven NN       & 1.523 & 1.142 & 73.10 \\
\bottomrule
\end{tabular}
\end{table}

\clearpage

\subsection{Gradient-Conflict Diagnostic for Soft-Constraint PINNs}
\label{suppl:gradient_conflict}

To quantify the optimisation pathology discussed in the Introduction
and in Section~\ref{sec:optimization_resolution}, the per-epoch
cosine similarity between the data-loss gradient
$\nabla_\theta \mathcal{L}_{\mathrm{data}}$ and the physics-loss
gradient $\nabla_\theta \mathcal{L}_{\mathrm{phys}}$ was recorded
over $700$ training epochs for the soft-constraint PINN baseline
used elsewhere in the manuscript. A negative cosine
indicates that the two losses pull the network in opposing parameter
directions, so that any gradient step necessarily increases one term
while decreasing the other.

\begin{figure}
\centering
\includegraphics[width=0.75\textwidth]{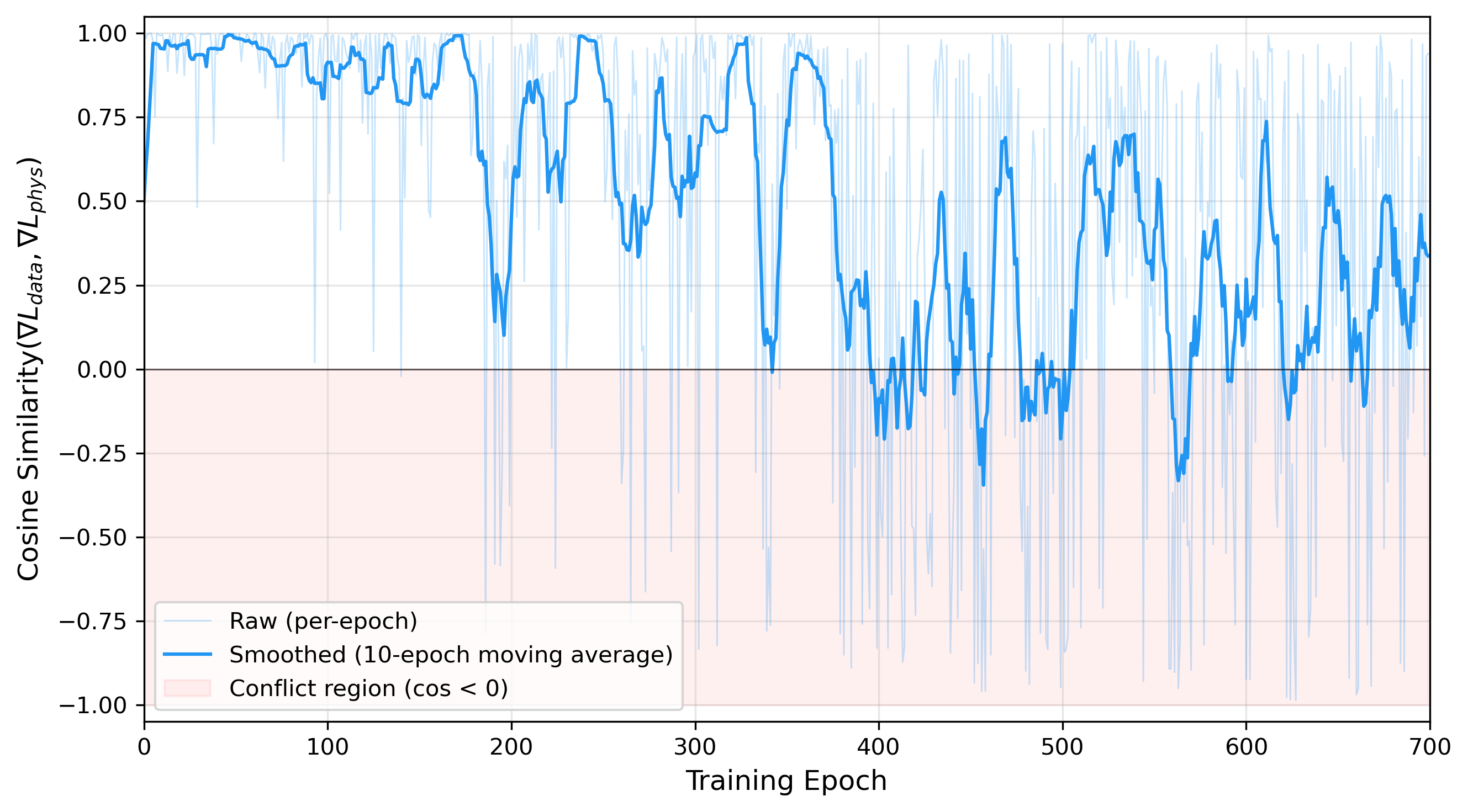}
\caption{Gradient-conflict diagnostic for the soft-constraint
PINN baseline.
Cosine similarity between the data-loss and physics-loss
gradients,
$\cos\angle(\nabla_\theta\mathcal{L}_{\mathrm{data}},
\nabla_\theta\mathcal{L}_{\mathrm{phys}})$,
recorded over $700$ training epochs for the soft-constraint PINN
baseline. The faint line shows the per-epoch raw values
and the bold line shows a $10$-epoch moving average for trend
clarity. Shaded zones below zero (red) denote epochs in which the
two losses oppose each other in parameter space; counted on the raw
trajectory, their fraction is $20.6\%$ overall ($144/700$) and
$34.9\%$ in the second half of training, with a maximum consecutive
negative-cosine streak of $6$ epochs.}
\label{fig:supp_exp2_gradient}
\end{figure}

Across the full training trajectory, the cosine becomes negative in
$144/700$ epochs ($20.6\%$); restricted to the second half of
training, the negative-cosine fraction rises to $34.9\%$, with a
maximum consecutive negative-cosine streak of $6$ epochs
(Fig.~\ref{fig:supp_exp2_gradient}). These data provide
the direct quantitative grounding for the qualitative statement,
documented in the PINN literature~\cite{bib20}, \cite{bib24}, and \cite{Wang2021},
that data-fidelity and physics objectives remain in genuine conflict
during soft-constraint training of tightly coupled physical systems.

\clearpage

\subsection{Supplementary Statistical Testing for Extrapolation Performance Comparison}
\label{suppl:statistical_framework}

To ensure methodological transparency and reproducibility,
Table~\ref{tab:stat_framework} summarises the complete
hierarchical testing strategy applied to the $n = 24$
extrapolation test samples across all three architectural
comparisons. Normality of absolute error distributions was
screened jointly via the Shapiro--Wilk and D'Agostino
$K^{2}$ tests (Table~\ref{tab:s_normality}) to determine
whether parametric assumptions were tenable before selecting
the inferential framework.

\begin{table}[htbp]
\centering
\caption{Statistical testing framework for three-way
architectural performance comparison ($n = 24$;
cathode pressures 120, 160, and 200\,bar; Nafion~117,
25$^{\circ}$C).}
\label{tab:stat_framework}
\renewcommand{\arraystretch}{1.35}
\begin{tabular}{>{\raggedright\arraybackslash}p{3.8cm} >{\raggedright\arraybackslash}p{8.5cm}}
\hline
\textbf{Analysis Level} & \textbf{Method} \\
\hline
Omnibus test &
  Friedman test~\cite{friedman1937} --- matched,
  non-parametric; appropriate for small-sample,
  non-normally distributed repeated-measures data \\
Post-hoc comparisons &
  One-sided Wilcoxon signed-rank
  tests~\cite{wilcoxon1945} with Holm--Bonferroni
  correction~\cite{holm1979} applied across three
  pairwise contrasts (PR-Net vs.\ soft-constraint PINN;
  PR-Net vs.\ purely data-driven NN; soft-constraint PINN vs.\ purely data-driven NN) \\
Effect size &
  Rank-biserial correlation $r$~\cite{kerby2014},
  interpreted as small ($|r| < 0.3$), medium
  ($0.3 \leq |r| < 0.5$), or large ($|r| \geq 0.5$) \\
Confidence intervals &
  Non-parametric bootstrap (percentile method,
  10,000 resamples)~\cite{efron1994} for mean
  absolute error differences \\
Interaction analysis &
  MAE degradation slope (\%p/bar) estimated via linear
  regression of pressure-stratified mean MAE against
  cathode pressure (120, 160, 200\,bar); slope ratios
  quantify accelerating performance divergence with
  extrapolation distance \\
\hline
\end{tabular}
\end{table}

\subsubsection{Rigorous Statistical Testing for Extrapolation}
\label{suppl:statistical_testing}

To formally establish the statistical significance of the PR-Net's
advantage in the high-pressure extrapolation regime
(Section~\ref{sec:extrapolation}), a hierarchical non-parametric
testing framework was applied to $n = 24$ test samples (Nafion~117,
25$^{\circ}$C; cathode pressures 120, 160, and 200~bar).

\paragraph{Normality assessment.}
Normality was evaluated jointly using the Shapiro--Wilk and
D'Agostino $K^2$ tests (Table~\ref{tab:s_normality}).
These diagnostics were used to determine whether parametric
assumptions were tenable for each model before selecting the
comparison procedure.

\paragraph{Omnibus and post-hoc tests.}
The Friedman omnibus test confirmed globally significant performance
differences under Protocol-IEP ($\chi^2 = 30.33$, $p < 0.001$).
Pairwise post-hoc Wilcoxon signed-rank tests with Holm--Bonferroni
correction (Table~\ref{tab:s_posthoc}) demonstrated that the PR-Net
significantly outperformed both baseline models with large effect
sizes ($|r| \geq 0.847$). Post-hoc power analysis confirmed achieved power $= 1.000$
for detecting the observed effect sizes ($|r| \geq 0.847$)
at $\alpha = 0.05$ ($n = 24$, one-sided Wilcoxon signed-rank).

\begin{table}[htbp]
\centering
\caption{Normality assessment of absolute prediction error
distributions for each architecture ($n = 24$). Distributions
classified as non-normal if either test yields $p < 0.05$.}
\label{tab:s_normality}
\begin{tabular}{lccccc}
\toprule
\textbf{Model}
  & \textbf{SW stat} & \textbf{SW \textit{p}}
  & \textbf{$K^2$ stat} & \textbf{$K^2$ \textit{p}}
  & \textbf{Normal?} \\
\midrule
PR-Net        & 0.9254 & 0.077    & 2.2378  & 0.327    & Yes \\
Soft PINN & 0.7901 & $<$0.001 & 17.7230 & $<$0.001 & No  \\
Data-driven NN       & 0.8437 & 0.002    & 13.2846 & 0.001    & No  \\
\bottomrule
\end{tabular}
\end{table}

\begin{table}[htbp]
\centering
\caption{Post-hoc pairwise Wilcoxon signed-rank test results with
Holm--Bonferroni correction ($n = 24$). Effect size $r$ (RBC):
large if $|r| \geq 0.5$.}
\label{tab:s_posthoc}
\begin{tabular}{lcccccc}
\toprule
\textbf{Pair}
  & \textbf{\textit{W}} & \textbf{raw \textit{p}}
  & \textbf{adj \textit{p}} & \textbf{Reject}
  & \textbf{\textit{r} (RBC)} & \textbf{Effect} \\
\midrule
PR-Net $<$ soft-constraint PINN
  & 23.0 & 0.000038  & 0.000038  & Yes & $-0.847$ & Large \\
PR-Net $<$ purely data-driven NN
  & 1.0  & $<$0.0001 & $<$0.0001 & Yes & $-0.993$ & Large \\
soft-constraint PINN $<$ purely data-driven NN
  & 11.0 & 0.000003  & 0.000007  & Yes & $-0.927$ & Large \\
\bottomrule
\end{tabular}
\end{table}

\subsubsection{Alternative Split Validations for the Apparatus-Bias Caveat}
\label{suppl:alt_split_validations}

The three alternative split validations introduced in
Section~\ref{sec:extrapolation} bound the contribution of
apparatus-specific characteristics to the pressure-axis
extrapolation gain. This appendix provides the full parity plots,
per-subgroup $R^2$ breakdowns, statistical tests, and architectural
discussion that supplement the in-text summary.

Across all three controls, PR-Net delivers the highest extrapolation
$R^2$ among the three machine-learning architectures
(Table~\ref{tab:s_alt_extrap}). The contrast is most pronounced under
Strategy~B, where eliminating Wu et al.~from both partitions drives
the data-driven NN and the soft-constraint PINN to negative $R^2$
while PR-Net sustains $R^2 = 0.893$ on the held-out
$i > 2.0$~A~cm$^{-2}$ partition; under Strategy~C the within-apparatus
contrast shows PR-Net at $R^2 = 0.884$ versus
$R^2 \approx 0.37$ for the two non-physics-integrated baselines. The
hard-constraint architecture therefore preserves its advantage when
apparatus-mixing confounders are removed (Strategy~B) and when the
train--test split is confined to a single apparatus (Strategy~C).

PR-Net's residual head functions as a complementary corrector that
converges towards a near-zero correction when the calibrated backbone
already captures the operative physics, and provides a bounded,
auditable adjustment in regimes where systematic
apparatus-dependent deviations exist (e.g., the 120--200~bar
pressure-axis extrapolation, or the 25/60$^{\circ}$C low-temperature
regime of Section~\ref{sec:mechanistic_discovery}). The
relatively lower Strategy~A current-density-axis extrapolation
$R^2 = 0.681$ for PR-Net reflects the sensitivity of the
current-density axis to membrane-specific kinetic parameters that
are captured directly through the optimised backbone coefficients
$(\alpha_k, \beta_k)$; the residual head's calibrated correction
over the training apparatus distribution does not transfer cleanly
to an unseen $i$ extrapolation regime unless apparatus mixing is
present at training time. This is consistent with the broader
finding that PR-Net's principal accuracy gains are concentrated
where (i) the deterministic backbone alone cannot fit the data
within its noise floor and (ii) the training set spans the
apparatus distribution from which the deployment data are drawn.

\begin{figure}
\centering
\includegraphics[width=\textwidth]{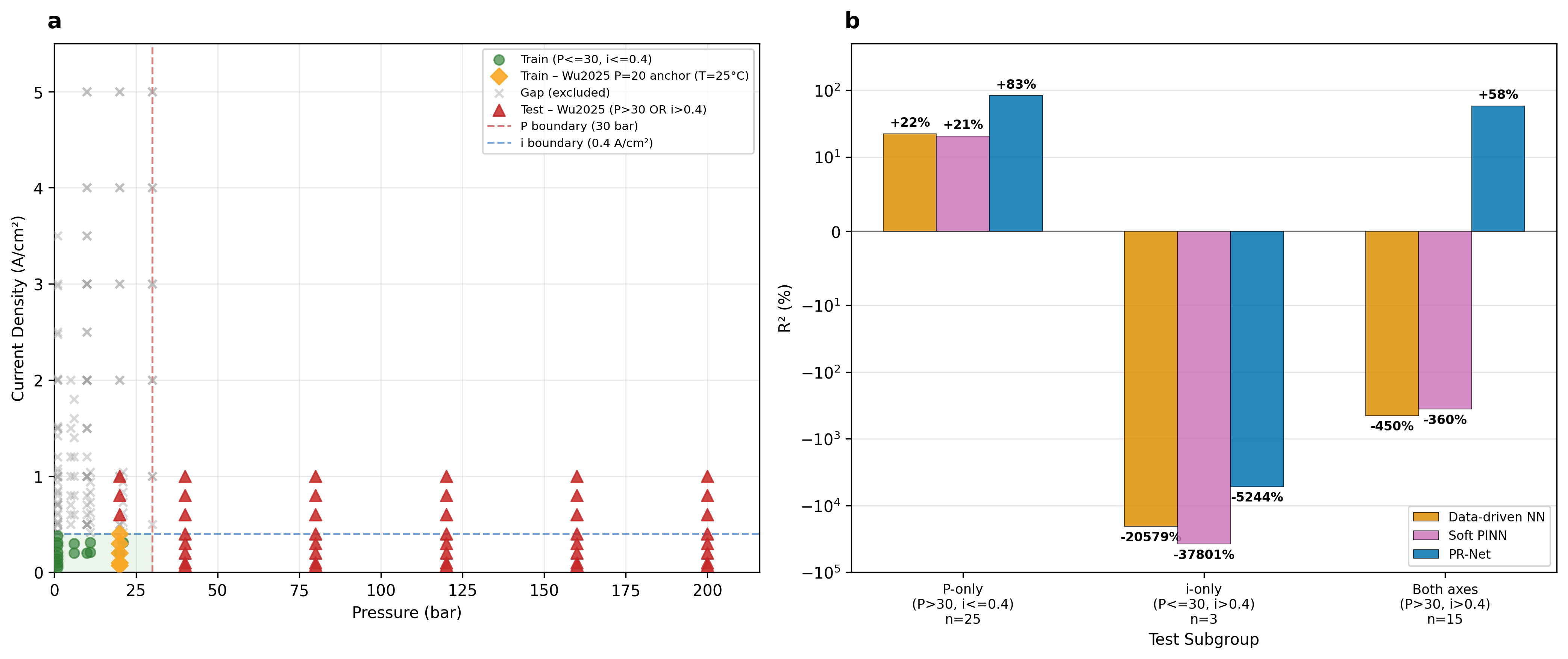}
\caption{Alternative extrapolation splits used to bound the
Wu-apparatus contribution to the high-pressure performance gap.
(a) Two-dimensional coverage map of the $(P,\ i)$ design space under
Strategy~C: training is restricted to $P \leq 30$~bar and
$i \leq 0.4$~A~cm$^{-2}$ (with the Wu et al.~$P = 20$~bar anchor
included), and testing covers Wu et al.~points with $P > 30$~bar or
$i > 0.4$~A~cm$^{-2}$, yielding a within-apparatus extrapolation along
both the pressure and current-density axes.
(b) Model-wise $R^2$ stratified by the test subgroup (pressure-only
extrapolation, current-density-only extrapolation, and both-axis
extrapolation). PR-Net sustains high $R^2$ in every subgroup, whereas
the data-driven NN and the soft-constraint PINN collapse, reproducing
the qualitative ranking of the pressure-axis test on a
within-apparatus contrast.}
\label{fig:s_alt_strategies}
\end{figure}

\begin{table}[ht]
\centering
\caption{Extrapolation performance under three alternative train/test
splits designed to bound the Wu-apparatus contribution. Strategy~A
uses all sources and splits on current density
($i \leq 1.0$ train, $i > 2.0$ test). Strategy~B repeats Strategy~A
after completely removing Wu et al.~from training and testing.
Strategy~C confines the analysis to Wu et al.~only and splits
two-dimensionally ($P \leq 30$ and $i \leq 0.4$ train; $P > 30$ or
$i > 0.4$ test). $R^2$ values are reported on the held-out test
partition; interpolation $R^2$ is on the training partition for the
same models. Boldface marks the best test $R^2$ in each row.}
\label{tab:s_alt_extrap}
\resizebox{\textwidth}{!}{%
\begin{tabular}{lccccccc}
\toprule
Strategy & $n_{\mathrm{train}}$ & $n_{\mathrm{test}}$
  & Interp.\ $R^2_{\mathrm{PR\text{-}Net}}$
  & \multicolumn{3}{c}{Extrapolation $R^2$ (test partition)}
  & Holm $p_{\mathrm{adj}}$ \\
\cmidrule(lr){5-7}
 & & & & Data-NN & Soft PINN & PR-Net
 & (PR-Net vs.\ Data-NN/Soft PINN) \\
\midrule
A (all sources, $i$-axis)
  & 123 & 30 & 0.996
  & $-0.225$ & $0.126$ & $\mathbf{0.681}$
  & $0.061 / 0.152$ \\
B (Wu excluded, $i$-axis)
  & 75 & 30 & 0.995
  & $-1.352$ & $-0.264$ & $\mathbf{0.893}$
  & $< 10^{-8}\ /\ <10^{-7}$ \\
C (Wu only, $P{+}i$ 2D)
  & 21 & 43 & 0.995
  & $0.365$ & $0.379$ & $\mathbf{0.884}$
  & $< 10^{-6}\ /\ <10^{-6}$ \\
\bottomrule
\end{tabular}}
\end{table}

\subsubsection{Per-Pressure Wilcoxon Signed-Rank Results}

To provide granular insight into the pressure-dependent
evolution of architectural performance differences,
pairwise one-sided Wilcoxon signed-rank tests were
additionally conducted at each individual extrapolation
pressure ($n = 8$ per pressure level, uncorrected
$\alpha = 0.05$; exploratory per-pressure analysis). Results are summarised in
Table~\ref{tab:per_pressure_wilcoxon}. Under Protocol-IEP, all pairwise comparisons reached
statistical significance at every extrapolation
pressure (120, 160, and 200\,bar), with large effect
sizes ($|r| \geq 0.833$) throughout. The
PR-Net vs.\ soft-constraint PINN comparison at 120\,bar
($1.5\times$ training pressure), which did not reach
significance under full-data backbone calibration
($p = 0.055$), achieves significance under
Protocol-IEP ($p = 0.020$), reflecting the removal
of the artificial performance boost conferred on the
soft-constraint PINN by high-pressure data inclusion in
backbone calibration.

\begin{table}[ht]
\centering
\caption{Per-pressure one-sided Wilcoxon signed-rank
test results ($n = 8$ per pressure
level; $\alpha = 0.05$, uncorrected; exploratory). Effect size $r$
denotes rank-biserial correlation; $|r| \geq 0.5$
indicates a large effect. All pairwise comparisons
are statistically significant across all three
extrapolation pressures.}
\label{tab:per_pressure_wilcoxon}
\begin{tabular}{llcccc}
\toprule
Pressure & Pair & $W$ & $p$ & $r$ (RBC)
         & Significant? \\
\midrule
\multirow{3}{*}{120~bar}
  & PR-Net $<$ soft-constraint PINN  & 3.0 & 0.020 & $-0.833$ & Yes \\
  & PR-Net $<$ purely data-driven NN   & 1.0 & 0.008 & $-0.944$ & Yes \\
  & soft-constraint PINN $<$ purely data-driven NN & 2.0 & 0.012 & $-0.889$ & Yes \\
\midrule
\multirow{3}{*}{160~bar}
  & PR-Net $<$ soft-constraint PINN  & 3.0 & 0.020 & $-0.833$ & Yes \\
  & PR-Net $<$ purely data-driven NN   & 0.0 & 0.004 & $-1.000$ & Yes \\
  & soft-constraint PINN $<$ purely data-driven NN & 2.0 & 0.012 & $-0.889$ & Yes \\
\midrule
\multirow{3}{*}{200~bar}
  & PR-Net $<$ soft-constraint PINN  & 3.0 & 0.020 & $-0.833$ & Yes \\
  & PR-Net $<$ purely data-driven NN   & 0.0 & 0.004 & $-1.000$ & Yes \\
  & soft-constraint PINN $<$ purely data-driven NN & 1.0 & 0.008 & $-0.944$ & Yes \\
\bottomrule
\end{tabular}
\end{table}

\subsubsection{Feasible Leave-One-Source-Out (LOSO) Robustness Summary}

To complement the shared-apparatus extrapolation analysis in Section~\ref{sec:extrapolation}, we conducted feasible inter-laboratory LOSO tests on four hold-out sources: St\"{a}hler et al.~(2020), Garbe et al.~(2019), Grigoriev et al.~(2011), and Schalenbach et al.~(2013). Because each held-out source contains a small number of samples and a narrow response variance, absolute-error metrics (RMSE and MAE) are treated as the primary basis for cross-model comparison, while $R^2$ is reported as a secondary descriptive indicator.

\begin{table}[htbp]
\centering
\caption{Feasible LOSO summary across four held-out sources (Deep Ensemble, $M = 100$). Per-source \textbf{overall} metrics are reported. In narrow-variance source subsets, RMSE/MAE are the primary comparison metrics and $R^2$ is secondary.}
\label{tab:s_loso_summary}
\scriptsize
\begin{adjustbox}{max width=\textwidth}
\begin{tabular}{llcccc}
\toprule
Held-out source & Model & $R^2$ (\%) & RMSE(\%p) & MAE(\%p) & MAPE (\%) \\
\midrule
\multirow{4}{*}{St\"{a}hler et al.~(2020)}
& Physics-only (IEP) & 3.90 & 0.3481 & 0.2748 & 41.3843 \\
& PR-Net & $\mathbf{42.96 \pm 20.38}$ & $\mathbf{0.2644 \pm 0.0448}$ & $\mathbf{0.1900 \pm 0.0472}$ & $\mathbf{26.5894 \pm 8.7081}$ \\
& Soft PINN & $-281.11 \pm 471.72$ & $0.5857 \pm 0.3707$ & $0.4641 \pm 0.2939$ & $73.4124 \pm 44.6489$ \\
& Data-driven NN & $-268.25 \pm 425.81$ & $0.5845 \pm 0.3502$ & $0.4431 \pm 0.2602$ & $67.9923 \pm 36.2922$ \\
\midrule
\multirow{4}{*}{Garbe et al.~(2019)}
& Physics-only (IEP) & \textbf{37.81} & 0.7420 & 0.7074 & 50.3566 \\
& PR-Net & $31.96 \pm 53.45$ & $\mathbf{0.7222 \pm 0.2840}$ & $\mathbf{0.6823 \pm 0.2976}$ & $\mathbf{48.8574 \pm 24.2246}$ \\
& Soft PINN & $-725.60 \pm 988.76$ & $2.3127 \pm 1.3997$ & $2.1463 \pm 1.4517$ & $169.5651 \pm 121.5693$ \\
& Data-driven NN & $-743.25 \pm 1010.67$ & $2.3241 \pm 1.4362$ & $2.1634 \pm 1.4845$ & $171.6795 \pm 123.7491$ \\
\midrule
\multirow{4}{*}{Grigoriev et al.~(2011)}
& Physics-only (IEP) & \textbf{-901.53} & \textbf{0.3569} & 0.3075 & 53.1604 \\
& PR-Net & $-964.61 \pm 89.41$ & $0.3677 \pm 0.0153$ & $\mathbf{0.3032 \pm 0.0118}$ & $\mathbf{51.6597 \pm 2.2808}$ \\
& Soft PINN & $-9951.51 \pm 6493.86$ & $1.0805 \pm 0.3333$ & $0.7959 \pm 0.1700$ & $121.1287 \pm 22.9608$ \\
& Data-driven NN & $-12988.33 \pm 8031.65$ & $1.2323 \pm 0.3825$ & $0.8866 \pm 0.2031$ & $133.7846 \pm 27.9829$ \\
\midrule
\multirow{4}{*}{Schalenbach et al.~(2013)}
& Physics-only (IEP) & \textbf{97.31} & \textbf{0.1084} & \textbf{0.0628} & \textbf{5.1228} \\
& PR-Net & $93.50 \pm 2.42$ & $0.1647 \pm 0.0359$ & $0.1496 \pm 0.0399$ & $20.9700 \pm 6.8913$ \\
& Soft PINN & $-0.24 \pm 14.92$ & $0.6602 \pm 0.0495$ & $0.4694 \pm 0.0614$ & $42.8356 \pm 10.6427$ \\
& Data-driven NN & $0.91 \pm 17.56$ & $0.6556 \pm 0.0589$ & $0.4131 \pm 0.0669$ & $32.5425 \pm 9.9670$ \\
\bottomrule
\end{tabular}
\end{adjustbox}
\end{table}

\begin{table}[htbp]
\centering
\caption{LOSO feasibility matrix across all eight contributing sources. ``Complete'' indicates sufficient overlap in membrane, temperature, and pressure domains to define a strict leave-one-source-out test; ``Partial'' indicates at least one source-specific domain element; ``Infeasible'' indicates critical domain uniqueness that prevents a meaningful strict LOSO comparison under the current dataset composition.}
\label{tab:s_loso_feasibility}
\scriptsize
\begin{adjustbox}{max width=\textwidth}
\begin{tabular}{lccccccl}
\toprule
Source & $n$ & Membrane & Temperature & Pressure & Class & Included & Note \\
\midrule
Garbe et al.~(2019)        & 24 & Yes & Yes & Yes & Complete   & Yes & Full overlap \\
Schalenbach et al.~(2013) &  9 & Yes & Yes & Yes & Complete   & Yes & Full overlap \\
St\"{a}hler et al.~(2020)      &  8 & Yes & Yes & Yes & Complete   & Yes & Full overlap \\
Grigoriev et al.~(2011)    &  9 & Yes & Partial & Yes & Partial & Yes & 85$^{\circ}$C treated as near-overlap with 80$^{\circ}$C \\
Martin et al.~(2021)       & 21 & Partial & Yes & Yes & Partial    & No  & Source-unique Nafion\_D2021 membrane \\
Bernt et al.~(2020)        & 37 & Partial & Yes & Yes & Partial    & No  & Source-unique 178/51~$\mu$m membrane settings \\
Trinke et al.~(2017)       & 28 & Partial & Yes & Yes & Partial    & No  & Source-unique FumaTech membrane family \\
Wu et al.~(2025)           & 48 & Yes & Partial & Partial & Infeasible & No & Source-unique 25$^{\circ}$C and 20--200~bar span \\
\bottomrule
\end{tabular}
\end{adjustbox}
\end{table}

\subsection{Fugacity Correction: PR-EOS Equations and Design-Choice Details}
\label{suppl:fugacity_details}

This appendix records the Peng--Robinson equation-of-state (PR-EOS)
machinery underlying the fugacity-coefficient values quoted in
Section~\ref{sec:fugacity_correction}, decomposes the high-pressure
residual into fugacity and non-fugacity contributions, and quantifies
the inverse-design overhead incurred by an explicit PR-EOS backbone.

\subsubsection{Peng--Robinson Cubic Equation and Closed-Form
$\ln \phi$ Expression}

With the canonical hydrogen parameters
($T_c = 33.19$~K, $P_c = 13.13$~bar,
$\omega = -0.219$~\cite{PengRobinson1976}), the fugacity coefficient
$\phi(T, P)$ is obtained as the unique real root of the cubic in
compressibility factor:
\begin{equation}
Z^3 - (1-B) Z^2 + (A - 2B - 3B^2) Z - (AB - B^2 - B^3) = 0,
\label{eq:pr_cubic}
\end{equation}
with $A = a(T) P / (R T)^2$ and $B = b P / (R T)$, where
$a(T)$ embeds the temperature-dependent attraction term and $b$ is
the co-volume parameter. The fugacity coefficient is recovered from
\begin{equation}
\ln \phi = (Z - 1) - \ln(Z - B)
- \frac{A}{2\sqrt{2}\, B}
\ln \! \left[\frac{Z + (1+\sqrt{2})B}{Z + (1-\sqrt{2})B}\right].
\label{eq:lnphi}
\end{equation}
For supercritical hydrogen ($T \gg T_c$) the negative acentric
factor produces $\phi > 1$ across the operating range:
$\phi \in [1.024, 1.029]$ at 80~bar (training-domain boundary,
25--80$^{\circ}$C) and $\phi \in [1.066, 1.072]$ at 200~bar
(extrapolation extreme, 25--80$^{\circ}$C). The
fugacity-induced bias in the predicted membrane-interface
hydrogen concentration therefore grows monotonically with pressure
but remains modest in absolute terms.

\subsubsection{Decomposition of the 200~bar Residual into Fugacity
and Non-Fugacity Contributions}

The pressure-stratified residuals of the ideal-gas backbone-only
prediction (Henry $\times$ ideal gas) provide a direct comparison
scale. At 200~bar, the mean signed residual
$\mu_{\mathrm{res}}^{\mathrm{ideal}} = -0.935$\%p
(Table~\ref{tab:mae_degradation}; absolute level relative to the
observed $\Phi_{\mathrm{H_2}} \in [2.5, 4.0]$\%p) is roughly
$25$--$30\%$ of the measured H$_2$ crossover. By contrast, the
$7\%$ fugacity correction at 200~bar shifts the predicted
$\Phi_{\mathrm{H_2}}$ by at most ${\sim}0.25$\%p
($\approx 7\%$ of ${\sim}3.5$\%p), i.e., roughly one quarter of
the observed residual magnitude. The fugacity coefficient is
therefore one of several contributions represented by
$\mathrm{NN}_{\mathrm{residual}}$, with the remaining
$\sim 70$--$75\%$ of the residual consistent with other
high-pressure phenomena (membrane swelling under elevated water
activity, LGDL compressibility, electroosmotic drag corrections,
and apparatus-specific transport effects). The empirical
backbone-variant comparison reported in
Table~\ref{tab:exp1_preos_comparison} and
Fig.~\ref{fig:exp1_preos_residuals} corroborates this
decomposition at the system level: replacing the ideal-gas backbone
with an explicit PR-EOS backbone closes the per-sample residual
distribution by an amount commensurate with the predicted
$0.25$\%p fugacity shift while leaving the residual standard
deviation essentially unchanged ($0.319 \to 0.284$\%p, a marginal
${\sim}11\%$ reduction).

\subsubsection{Inverse-Design Cost Overhead of an Explicit PR-EOS
Backbone}

The ideal-gas backbone admits closed-form analytical sensitivities
for the inverse-design queries of
Sections~\ref{sec:optimization_resolution}--\ref{sec:economic_causal_chain}
(operating-envelope optimisation, safety-margin quantification, and
the IEA-basis economic analysis), whereas a PR-EOS backbone would
require iterative cubic-EOS root finding for every inverse-design
evaluation. In our prototype implementations this overhead is
approximately one order of magnitude per inference call, which
becomes consequential when the safety-margin and $i_{\min}$ grids
of Section~\ref{sec:economic_causal_chain} are repeatedly swept
over $(P, T)$ ranges. The PR-EOS embedding is therefore documented
as a drop-in alternative for downstream studies that prioritise
high-pressure analytical fidelity over inversion speed, while the
ideal-gas backbone remains the default configuration on the basis
of (i) comparable end-to-end accuracy, (ii) analytical tractability,
and (iii) diagnostic richness of the residual signal that supports
the autonomous-discovery analysis in
Section~\ref{sec:mechanistic_discovery}.

\subsection{Quantitative Detail of the Techno-Economic Scenario Pathway (EA-1 to EA-4)}
\label{suppl:economic_details}

\begin{table}[ht]
\centering
\caption{Definitions of the four auxiliary techno-economic analyses
(EA-1 to EA-4) that constitute the prediction-to-economics scenario
pathway of Section~\ref{sec:economic_causal_chain}. Each analysis maps
a prediction-level input to an operating-envelope estimate or
plant-level scenario; the corresponding quantitative results are reported in the
subsections below and in Section~\ref{sec:economic_causal_chain}.}
\label{tab:ea_definitions}
\begin{tabular}{l p{0.20\linewidth} p{0.26\linewidth} p{0.30\linewidth}}
\toprule
ID & Analysis & Input & Output \\
\midrule
EA-1 & Safety-margin analysis & PR-Net prediction $+$ residual-SD
confidence interval & Maximum allowable operating pressure
($P_{\max}$ gain) \\
\addlinespace
EA-2 & Safe operating window & PR-Net prediction $+$ residual-SD
confidence interval & Minimum safe partial-load current density
($i_{\min}$) and operating-window width \\
\addlinespace
EA-3 & Early-warning accuracy & PR-Net vs.\ baseline prediction
error & False-alarm / miss rate \\
\addlinespace
EA-4 & Conservative-margin $\sigma$-recovery $+$ IEA \$/yr
conversion & Residual SD $+$ IEA~(2019) H$_2$ price &
$\sigma$-reduction efficiency-recovery scenario (\%) $+$ OPEX scenario
(\$/yr; 1/10/100~MW plant) \\
\bottomrule
\end{tabular}
\end{table}

This appendix records the per-band, per-$(P, T)$ quantitative
detail underlying the four-step scenario pathway summarised in
Section~\ref{sec:economic_causal_chain}.
In every analysis (defined in Table~\ref{tab:ea_definitions}),
$\sigma$ is the per-band empirical residual standard deviation,
computed for PR-Net and the physics baseline on the same globally
pre-calibrated backbone.
EA-1 and EA-2 use Nafion~117 only---the single membrane with
above-80~bar data---whereas EA-3 covers all six membranes in
interpolation ($n = 184$) and Nafion~117 in extrapolation
($n = 24$), and EA-4 is reported for both an all-membranes and a
Nafion~117 scope (coinciding in the high-pressure band).

\subsubsection{EA-4: Band-Resolved Residual-SD Reduction
(Step~1 detail)}

In the high-pressure extrapolation regime (81--200~bar,
Nafion~117, $n = 24$), PR-Net attains a signed-error standard
deviation $\sigma_{\mathrm{res}} = 0.319$\%p~H$_2$, a ${\sim}74\%$
reduction relative to the data-driven NN out-of-fold ensemble
baseline ($1.214$\%p, rising to $1.353$\%p at 200~bar alone) and
a ${\sim}18\%$ reduction relative to the physics model
($0.390$\%p). The latter reduction is the one
used in the operating-envelope and OPEX scenario analyses below,
which reference the physics model as the conventional
industrial incumbent.

\subsubsection{EA-1 and EA-2: Per-$(P, T)$ Safety-Margin Grids
(Step~2 detail)}

Propagating the Step~1 reduction in $\sigma$ through the
95\%-confidence upper bound on the 4\%~H$_2$ explosive threshold
produces the scenario per-$(T)$ maximum-safe-pressure grid (EA-1) and the
per-$(P, T)$ minimum current-density grid (EA-2) visualised in
Fig.~\ref{fig:exp13_economic_chain}b. At $T = 60^{\circ}$C, the
maximum pressure that remains safe rises from
94.5~bar under the calibrated Henry--Fick--Faraday backbone alone
to 103.7~bar under PR-Net (EA-1), a $+9.2$~bar margin gain. In the
industrial band the corresponding gains are smaller---from 68.3 to
71.1~bar at $80^{\circ}$C ($+2.8$~bar) and from 61.9 to 64.8~bar at
$85^{\circ}$C ($+2.9$~bar)---consistent with the near-zero residual
there, while at $25^{\circ}$C both models remain safe across the
full pressure sweep.
At fixed $P = 200$~bar, $T = 80^{\circ}$C, the minimum current
density required for safe partial-load operation drops from
$i_{\min} = 4.28$ to $3.78$~A~cm$^{-2}$, a $+69.6$\% widening of
the load-following window (EA-2). Both estimates are computed from
the Step~1 reduction in $\sigma$ under the standard Gaussian
upper-bound propagation assumption.

\subsubsection{EA-4 to OPEX: IEA-Basis Plant-Scale Translation
(Step~3 detail)}

Adopting the IEA~2019 hydrogen cost model
(H$_2$ price $\$5.0$~kg$^{-1}$, capacity factor 0.7, plant
efficiency $\eta = 0.7$ \cite{IEA2019}), the $\sigma$-reduction
recovered by PR-Net over the physics model is converted to a
recovered-electrical-efficiency scenario band-by-band
(Fig.~\ref{fig:exp13_economic_chain}c; EA-4): $+9.0\%$ at
1--20~bar, ${\sim}0\%$ at 21--80~bar, and $+2.1\%$ at 81--200~bar
on a Nafion~117 basis ($n = 24$ in the high band). Summed over the
bands in which PR-Net improves on the physics model, the IEA
cost model gives illustrative OPEX scenarios of $\sim\$71$k~yr$^{-1}$
at 1~MW, $\sim\$0.71$M~yr$^{-1}$ at 10~MW, and
$\sim\$7.1$M~yr$^{-1}$ at 100~MW.
The OPEX range quoted in Table~\ref{tab:economic_benefits}
($\$200$k--$\$1.5$M~yr$^{-1}$) is the IEA-benchmarked industrial
maintenance/downtime band, which the recovery envelope derived
here brackets at representative ($\approx 10$~MW) plant scale.

\subsubsection{EA-3: Early-Warning Classification Error
(Step~4 detail)}

On the extrapolation hold-out, the binary risk-classification
error at a 3.5\%~H$_2$ early-warning threshold falls from
$23.8$--$28.6\%$ for purely data-driven NN and soft-constraint
PINN baselines to $0.0\%$ for PR-Net (EA-3). Together with
the IEA-benchmarked OPEX scenarios of EA-4, these four
prediction-to-economic-outcome translations furnish the
quantitative scenario pathway summarised in
Section~\ref{sec:economic_causal_chain}: prediction-level
accuracy (MAE, residual $\sigma$) $\to$ operating-envelope estimates
($P_{\max}$, $i_{\min}$) $\to$ plant-level OPEX scenarios (IEA~2019 basis)
and event-cost-priced incident prevention.

\subsection{State-of-the-Art PINN Applications in
Electrochemical Systems}
\label{suppl:sota_pinn}

Table~\ref{tab:s_literature} uses the following abbreviations:
SVR, support vector regression; ANN, artificial neural network; STL,
seasonal-trend decomposition using Loess; DNN, deep neural network;
KNN, $k$-nearest neighbours; RF, random forest; GBDT,
gradient-boosted decision tree; CNN, convolutional neural network; AL,
active learning; SHAP, Shapley additive explanations; GA, genetic
algorithm; RSM, response surface methodology; MSE, mean squared error;
LSTM, long short-term memory; FFNN, feedforward neural network; FCNN,
fully convolutional neural network; ODEs, ordinary differential
equations; SOH, state of health; and NCM, nickel cobalt manganese.
The superscript $^{a}$ denotes a preprint that had not been peer
reviewed at the time of writing.

\begin{sidewaystable}[htbp]
\centering
\caption{Supplementary literature map organised into two blocks: recent PEMWE machine-learning review articles and representative PINN applications in PEM-related or broader electrochemical systems. This table is referenced in the Introduction
and Section 2.3 to support the claims that hydrogen crossover concentration has not been established as a primary PEMWE machine-learning target and that prior PEM-related PINN studies have predominantly adopted soft-constraint formulations.}
\label{tab:s_literature}
\begin{adjustbox}{max width=\textheight}
\begin{tabular}{>{\raggedright\arraybackslash}p{2.8cm}>{\raggedright\arraybackslash}p{1.3cm}>{\raggedright\arraybackslash}p{4.4cm}>{\raggedright\arraybackslash}p{2.7cm}>{\raggedright\arraybackslash}p{4.2cm}>{\raggedright\arraybackslash}p{4.7cm}>{\raggedright\arraybackslash}p{2.5cm}}
\toprule
\textbf{Application}
& \textbf{System}
& \textbf{Key Innovation}
& \textbf{Reported Results}
& \textbf{Model architecture / ML form}
& \textbf{Relevance to present work}
& \textbf{Reference} \\
\midrule

\multicolumn{7}{@{}l}{\textit{Review articles on PEMWE machine learning}} \\
\midrule

PEMWE ML Review
& PEMWE
& Review of ML applications for PEM electrolyser development, optimisation, and diagnostics
& Review article
& Literature review; supervised ML methods broadly surveyed
& Confirms broad ML activity, but no direct H$_2$ crossover concentration target identified
& Ding et al. (2023) \cite{bib17} \\

Knowledge-Integrated ML Review/Framework
& PEMWE
& Ladder of Knowledge-integrated Machine Learning; uncertainty analysis and knowledge decomposition for PEMWE
& Framework paper with case studies
& SVR, decision tree, ANN; STL-based data decomposition; PINN-like soft constraint; symbolic regression
& Frames PEMWE ML landscape without a hydrogen-crossover target focus
& Chen et al. (2024) \cite{bib18} \\

Component-Level Review
& PEMWE
& Systematic review of ML applications across PEMWE subsystems
& Review article
& ANN, DNN, SVR, KNN, RF, GBDT, XGBoost, CatBoost, CNN, AL, SHAP, GA, RSM, etc.
& Broad coverage of PEMWE ML targets; crossover concentration not identified as a primary ML target in the reviewed scope
& Albadwi et al. (2024) \cite{bib55} \\

\midrule
\multicolumn{7}{@{}l}{\textit{PINN applications relevant to PEM-related or broader electrochemical systems}} \\
\midrule

Temperature Prediction
& PEMWE
& Inverse and parametric PINNs for 0D/1D thermal modelling and parameter identification
& MSE $= 0.1596$ vs LSTM $= 1.2132$
& PINN with 0D FFNN, 1D FCNN, and DeepONet-inspired parametric PINN; soft-constraint loss
& PEM electrolysis thermal dynamics, not crossover
& Zerrougui et al. (2025) \cite{bib52} \\

RUL Prediction
& PEM fuel cell
& Soft-constraint PINN with membrane/catalyst degradation physics and current-density replication in prediction phase
& 9.2 percentage-point improvement; 26\% input data required
& DNN-based PINN (2 inputs, 8 hidden layers, 20 neurons each, tanh); multi-loss physics-informed training
& PEM-related PINN for prognostics, not PEMWE crossover
& Ko et al. (2025) \cite{bib51} \\

Membrane Degradation
& PEMWE
& First PINN-based framework for membrane thinning; coupled voltage and degradation ODEs with inverse inference of degradation rate
& Test RMSE: voltage 0.0047, membrane thickness 0.000061; outperforms ANN
& FFNN PINN with 2 hidden layers (10, 5), sigmoid; two outputs; soft-constraint coupled ODEs
& Closest PEMWE PINN application, but targets degradation rather than crossover
& Polo-Molina et al. (2025)$^{a}$ \cite{bib53} \\

Battery SOH
& Li-ion (NCM)
& Hybrid PINN combining empirical degradation and state-space equations; feature extraction from pre-charge segment
& MAPE $= 0.87\%$ on 387 batteries
& Dual-branch fully connected PINN (feature-to-SOH network + degradation-dynamics network)
& Cross-domain example of physics-informed prognostics rather than PEMWE crossover
& Wang et al. (2024) \cite{s15} \\

\bottomrule
\end{tabular}
\end{adjustbox}
\end{sidewaystable}

\clearpage

\subsection{Physics-Backbone Calibration: Differential-Evolution Solver Details}
\label{suppl:de_calibration_details}

This section documents the exact differential-evolution
(DE) hyperparameters and pseudocode used to calibrate
the per-membrane mass-transfer coefficients
$(a_{\alpha}, b_{\alpha}, a_{\beta}, b_{\beta})_k$ in
Eqs.~\eqref{eq:alpha_param}--\eqref{eq:beta_param}. The
two calibration protocols introduced in
Section~\ref{subsec:physics_model} (Protocol-IEP and
Protocol-FCP) share an identical solver configuration
and differ only in the calibration subset
$\mathcal{D}_k$.

The optimisation objective for each membrane $k$ is the
mean squared error between the analytical backbone
prediction and the measured H$_2$ crossover,
$\mathrm{MSE}(a_{\alpha}, b_{\alpha}, a_{\beta},
b_{\beta}) = \frac{1}{n_k}\sum_{j=1}^{n_k}
(\Phi_{\mathrm{H_2}, j} - \Phi_{\mathrm{H_2},
j}^{\mathrm{phys}})^2$. The fixed solver settings,
identical for IEP and FCP, are:
\begin{itemize}
  \setlength{\itemsep}{2pt}
  \item \textit{Box constraints:}
        $a_{\alpha} \in [10^{-5}, 10^{-1}]$,
        $b_{\alpha} \in [-2.0, 0.0]$,
        $a_{\beta} \in [-1.0, 2.0]$,
        $b_{\beta} \in [-1.0, 1.0]$. Bounds for
        $(a_{\alpha}, b_{\alpha})$ keep
        $\alpha = a_{\alpha} P^{b_{\alpha}}$ positive
        and monotonically decreasing with pressure;
        bounds for $(a_{\beta}, b_{\beta})$ keep
        $\beta = a_{\beta} + b_{\beta}\ln P$ in the
        empirically observed range across membranes.
  \item \textit{Initial population:} 10 candidates per
        parameter (\texttt{popsize}~$=10$, total $40$
        vectors) generated by Latin-hypercube sampling
        over the box constraints. The pooled-fit seed
        $\boldsymbol{\theta}_0 = (a_{\alpha},
        b_{\alpha}, a_{\beta}, b_{\beta}) =
        (5.06\times 10^{-3}, -0.652, 0.532, 0.056)$,
        obtained from a pooled fit across all
        membranes, is injected into the initial
        population and is also retained as the
        inference-time fallback parameter set
        (\texttt{alpha\_common}, \texttt{beta\_common})
        used for any membrane type absent from the
        calibration subset.
  \item \textit{Strategy and convergence:} default
        \texttt{best1bin} mutation/crossover;
        termination when either (i) the standard
        deviation of the population objective values
        falls below $\mathrm{tol} = 10^{-2}$ of the
        population mean, or (ii) the maximum number of
        generations $\mathrm{maxiter} = 200$ is
        reached, yielding an upper bound of
        $200 \times 10 \times 4 = 8{,}000$ objective
        evaluations per membrane.
  \item \textit{Reproducibility:} the random-number
        generator is seeded with \texttt{seed}~$=42$,
        so the entire IEP/FCP calibration is
        bit-for-bit reproducible.
\end{itemize}

\begin{algorithm}[H]
\caption{Differential-evolution calibration of
$(a_{\alpha}, b_{\alpha}, a_{\beta}, b_{\beta})$ for
membrane $k$ (identical for Protocol-IEP and
Protocol-FCP; only the calibration subset
$\mathcal{D}_k$ differs).}
\label{alg:de_calibration}
\begin{algorithmic}[1]
\Require Calibration subset
  $\mathcal{D}_k = \{(T_j, P_j, i_j, t_{m,j},
  \Phi_{\mathrm{H_2}, j})\}_{j=1}^{n_k}$ for membrane $k$
\Require Box constraints
  $\mathcal{B} = [10^{-5}, 10^{-1}] \times [-2, 0]
   \times [-1, 2] \times [-1, 1]$
\Require Fallback seed
  $\boldsymbol{\theta}_0 = (5.06\times 10^{-3},
  -0.652, 0.532, 0.056)$
\Require Hyperparameters
  $\mathrm{popsize}=10$, $\mathrm{maxiter}=200$,
  $\mathrm{tol}=10^{-2}$, $\mathrm{seed}=42$,
  strategy $=$ \texttt{best1bin}
\State Initialise population
  $\mathcal{P}^{(0)}$ of size $40$ by Latin-hypercube
  sampling over $\mathcal{B}$; inject
  $\boldsymbol{\theta}_0$ into $\mathcal{P}^{(0)}$
\For{$g = 1, \dots, \mathrm{maxiter}$}
  \For{each candidate
       $\boldsymbol{\theta}^{(g)} \in \mathcal{P}^{(g)}$}
    \State Mutate and crossover via \texttt{best1bin}
    \State Evaluate
      $\mathrm{MSE}(\boldsymbol{\theta}^{(g)}) =
       \frac{1}{n_k} \sum_{j=1}^{n_k}
       \bigl(\Phi_{\mathrm{H_2}, j} -
        \Phi_{\mathrm{H_2}, j}^{\mathrm{phys}}
        (T_j, P_j, i_j, t_{m,j};
         \boldsymbol{\theta}^{(g)})\bigr)^2$
    \State Select survivor by greedy comparison
  \EndFor
  \If{$\mathrm{std}(\mathrm{MSE})/|\mathrm{mean}
      (\mathrm{MSE})| < \mathrm{tol}$}
    \State \textbf{break}
  \EndIf
\EndFor
\State \Return
  $\boldsymbol{\theta}_k^{\star}
   = \arg\min_{\boldsymbol{\theta} \in \mathcal{P}^{(g)}}
     \mathrm{MSE}(\boldsymbol{\theta})$
\end{algorithmic}
\end{algorithm}

\clearpage

\subsection{Computational Framework and Hyperparameters}
\label{suppl:computational_framework}

For reproducibility, this section details the
computational hardware benchmarking and ensemble training
configurations. Table~\ref{tab:s_hardware} outlines the three
platforms used to validate the millisecond-level edge deployment
capability. Table~\ref{tab:s_hyperparameters} details the reproducible
training protocols used across the interpolation and extrapolation
pipelines, while Table~\ref{tab:s_hyperparams_search} summarises the
hyperparameter search space and interpolation-optimal settings.

\begin{table}[htbp]
\centering
\caption{Inference performance benchmarking and hardware
specifications. Each platform executed 1,000 forward passes
(discarding the first 100 as warm-up) using high-resolution
performance counters with microsecond precision.}
\label{tab:s_hardware}
\begin{adjustbox}{max width=\textwidth}
\begin{tabular}{llccccc}
\toprule
\textbf{Platform}
  & \textbf{CPU / GPU}
  & \textbf{RAM}
  & \textbf{Model Load (ms)}
  & \textbf{Single Inference (ms)}
  & \textbf{Batch-100 (ms)}
  & \textbf{Power (W)} \\
\midrule
Desktop PC
  & Intel Core i7-13700 + RTX-4060
  & 48~GB & 125 & $0.31 \pm 0.04$ & 28.7  & 125 \\
Edge GPU (Jetson AGX Orin)
  & ARM A78AE + GPU
  & 64~GB & 340 & $1.36 \pm 0.03$ & 122.1 & 40  \\
Edge CPU (Raspberry Pi 5)
  & ARM Cortex-A76
  & 8~GB  & 1850 & $1.08 \pm 0.34$ & 99.8  & 16  \\
\bottomrule
\end{tabular}
\end{adjustbox}
\end{table}

\begin{table}[htbp]
\centering
\caption{Detailed cross-validation and deep ensemble training
configurations ensuring reproducibility across both validation
protocols.}
\label{tab:s_hyperparameters}
\begin{tabular}{@{}p{5cm}p{5cm}p{5cm}@{}}
\toprule
\textbf{Parameter}
  & \textbf{5-Fold CV (Interpolation)}
  & \textbf{Deep Ensemble (Extrapolation)} \\
\midrule
Number of models
  & 100 (5 folds $\times$ 20 repeats)
  & 100 independent models \\
Random seed strategy
  & $42 + (\mathrm{rep\_idx} \times 5)$
  & $42 + \mathrm{model\_idx}$ \\
Training data scope
  & All 184 points (stratified 80/20)
  & $\leq$80~bar only ($n = 42$) \\
Test data scope
  & 20\% held-out per fold (OOF)
  & 120, 160, 200~bar ($n = 24$) \\
Stratification
  & By membrane type (6 classes)
  & N/A \\
Training epochs (max)
  & 700
  & 700 \\
Early stopping patience
  & 250 epochs (delta: $10^{-6}$)
  & 250 epochs (delta: $10^{-6}$) \\
Learning rate
  & PR-Net: $1.5 \times 10^{-3}$; Soft PINN/NN: $2.5 \times 10^{-3}$
  & PR-Net: $1.5 \times 10^{-3}$; Soft PINN/NN: $2.5 \times 10^{-3}$ \\
Optimiser
  & Adam
  & Adam \\
Batch size
  & 32
  & 32 \\
Physics weight $\beta$ (soft-constraint PINN)
  & Linear decay: $0.7 \rightarrow 0.01$
  & Linear decay: $0.7 \rightarrow 0.01$ \\
Residual weight $\lambda$ (PR-Net)
  & 2.0 (fixed)
  & 2.0 (fixed) \\
Purpose
  & Interpolation accuracy \& training stability
  & Uncertainty quantification \& extrapolation robustness \\
\bottomrule
\end{tabular}
\end{table}

\begin{table}[htbp]
\caption{Hyperparameter search space and selected configuration
for the PR-Net, soft-constraint PINN, and data-driven NN architectures. Grid search
was conducted via stratified 5-fold cross-validation
on the full training dataset ($n = 184$). Shared architecture
settings were fixed across models unless otherwise noted.}
\label{tab:s_hyperparams_search}
\scriptsize
\begin{adjustbox}{max width=\textwidth}
\begin{tabular*}{\textwidth}{@{}p{2.7cm}p{5.6cm}p{3.1cm}p{3.4cm}@{}}
\toprule
\textbf{Hyperparameter} & \textbf{Search Space}
  & \textbf{Optimal Value} & \textbf{Notes} \\
\midrule
\multicolumn{4}{@{}l}{\textit{Shared architectural settings}} \\
\midrule
Learning rate $\eta$
  & \makecell[l]{PR-Net / Soft PINN / Data-driven NN:\\
    $\{5.5\times10^{-3},\ 2.5\times10^{-3},\ 1.5\times10^{-3},\ 0.5\times10^{-3}\}$}
  & \makecell[l]{PR-Net: $1.5\times10^{-3}$\\
    Soft PINN: $2.5\times10^{-3}$\\
    Data-driven NN: $2.5\times10^{-3}$}
  & Adam optimiser \\
Hidden layer width
  & $\{32,\ 64,\ 128,\ 256\}$
  & 128
  & neurons per layer \\
Number of hidden layers
  & $\{2,\ 3,\ 4,\ 5\}$
  & 3
  & shared across models \\
Activation function
  & \{tanh, ReLU, SiLU\}
  & tanh
  & smooth gradient \\
Batch size
  & $\{16,\ 32,\ 64\}$
  & 32
  & shared across models \\
Max training epochs
  & fixed
  & 700
  & shared across models \\
Early stopping patience
  & fixed
  & 250 epochs
  & $\delta = 10^{-6}$ \\
\midrule
\multicolumn{4}{@{}l}{\textit{PR-Net specific}} \\
\midrule
Residual weight $\lambda$
  & $\{20,\ 10,\ 5,\ 2,\ 1,\ 0.5,\ 0.1\}$
  & 2.0
  & interp.--extrap.\ trade-off \\
\midrule
\multicolumn{4}{@{}l}{\textit{Soft PINN specific}} \\
\midrule
Physics weight $\beta$
  & \makecell[l]{$\{(0.9,0.5),\ (0.9,0.1),\ (0.9,0.01),\ (0.7,0.5),$\\
    $(0.7,0.1),\ (0.7,0.01),\ (0.5,0.1),\ (0.5,0.01)\}$}
  & $(0.7,\ 0.01)$
  & \makecell[l]{initial-to-final\\ linear decay;\\ interp.-optimal\\ Pareto setting} \\
\bottomrule
\end{tabular*}
\end{adjustbox}

\vspace{4pt}
\noindent\footnotesize
Preliminary experiments with 4--5 hidden layers showed no
statistically significant accuracy improvement
($\Delta R^2 < 0.05\%$) while increasing training instability
and inference latency. The 128-neuron width provides
sufficient capacity to capture non-linear transport phenomena
while maintaining millisecond-level edge inference
(Table~\ref{tab:s_hardware}). Model-specific learning-rate optima differed across PR-Net, soft-constraint PINN, and the data-driven NN, whereas the shared architecture depth and width were retained for controlled comparison.

\end{table}

\begin{figure}
\centering
\includegraphics[width=0.75\linewidth]{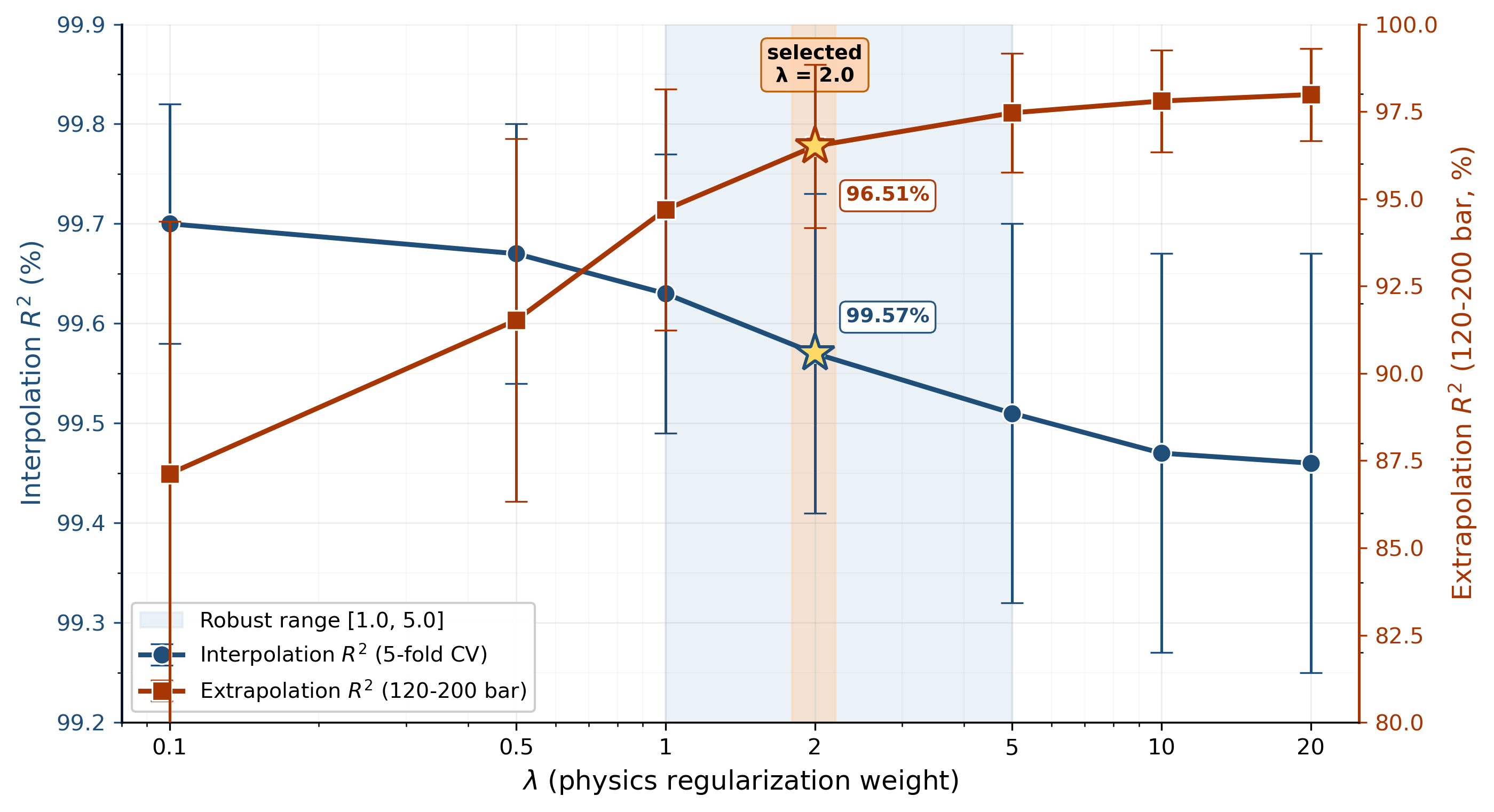}
\caption{Interpolation--extrapolation trade-off as a function of the
residual regularisation weight $\lambda$. Left axis (blue):
interpolation CV $R^2$ over a stratified 5-fold cross-validation
with 20 repeats; error bars denote $\pm 1$ standard deviation across
the 100 fold-level estimates. Right axis (red): extrapolation $R^2$
on the 120--200~bar test set from a $M = 100$ Deep Ensemble trained
under Protocol-IEP (train $\leq 80$~bar). Extrapolation $R^2$
saturates at large $\lambda$ ($97.99\%$ at $\lambda = 20$, the upper
end of the swept range); the shaded band marks the empirically
balanced range $[1.0, 5.0]$ and the highlighted band marks the
selected operating point $\lambda = 2.0$.}
\label{fig:exp4_lambda_tradeoff}
\end{figure}

\paragraph{Residual regularisation weight selection.}
The residual regularisation weight $\lambda$ in
Eq.~\ref{eq:prnet_loss} was selected by a controlled
empirical sweep over
$\lambda \in \{0.1,\,0.5,\,1.0,\,2.0,\,5.0,\,10.0,\,20.0\}$
under the two validation protocols described in
Section~\ref{sec:training_validation}: (i) interpolation
accuracy quantified by stratified 5-fold cross-validation
with 20 repeats (100 trained models per $\lambda$,
fold-level CV $R^2$); and (ii) extrapolation accuracy on the 120--200~bar
test set evaluated with a $M = 100$ Deep Ensemble trained
under Protocol-IEP. Aggregate metrics are reported in
Table~\ref{tab:exp4_lambda} and visualised in
Fig.~\ref{fig:exp4_lambda_tradeoff}; the corresponding mean
residual magnitude
$\langle |\Delta y_{\mathrm{NN}}| \rangle$ characterises
how strongly each setting suppresses the learned correction.

Two regimes emerge consistently across the swept range. At
small $\lambda$ the residual branch is essentially
unconstrained and over-fits to features of the $\leq 80$~bar
training subset, so extrapolation $R^2$ at 120--200~bar
collapses to $87.12\%$ at $\lambda = 0.1$ even though
interpolation remains nominally high ($R^2 = 99.70\%$).
Conversely at large $\lambda$ the residual is driven towards
zero ($\langle|\Delta y_{\mathrm{NN}}|\rangle$ falls from
$1.147$ at $\lambda = 0.1$ to $0.048$\%~H$_2$ at
$\lambda = 20.0$); extrapolation accuracy saturates,
reaching $97.99\%$ at $\lambda = 20$, but the residual
branch loses the capacity to capture the unmodelled
interfacial phenomena that the in-domain experiments
support, and interpolation $R^2$ degrades monotonically to
$99.46\%$ at $\lambda = 20$. Theoretically, as $\lambda$
grows without bound the $L_2$ residual penalty drives
$\mathrm{NN}_{\mathrm{residual}} \to 0$ and the prediction
converges to the deterministic backbone alone, consistent
with the observed empirical saturation.
The selected operating point $\lambda = 2.0$ retains a
non-trivial residual signal
($\langle|\Delta y_{\mathrm{NN}}|\rangle = 0.294$\%~H$_2$ at
200~bar; $0.213$\%~H$_2$ averaged across 120--200~bar), reaches an extrapolation $R^2$ of
$96.51\%$ in this regime, and gives up only $0.13$
percentage points of interpolation $R^2$ relative to
$\lambda = 0.1$.

\begin{table}[ht]
\centering
\caption{Empirical justification of the residual regularisation
weight $\lambda$. Interpolation column reports stratified 5-fold
cross-validation $R^2$ (20 repeats; mean $\pm$ standard deviation
over $5 \times 20 = 100$ fold-level estimates); extrapolation
columns report Deep Ensemble ($M = 100$) $R^2$ on the unseen
120--200~bar test set under Protocol-IEP, and the corresponding
mean absolute NN-residual at 200~bar. The selected operating point
$\lambda = 2.0$ lies at the elbow of the interpolation--extrapolation
trade-off within the displayed $[1.0,\,5.0]$ band, balancing high
interpolation accuracy, improved extrapolation accuracy, and
preservation of a non-trivial residual correction.}
\label{tab:exp4_lambda}
\begin{tabular*}{\textwidth}{@{\extracolsep{\fill}}lllll@{}}
\toprule
$\lambda$ & Interp.\ $R^2$ (\%) & Extrap.\ $R^2$ at & Extrap.\ $R^2$ at & Mean $|\Delta y_{\mathrm{NN}}|$ \\
          & (5-fold $\times$ 20 CV) & 200~bar (\%)      & 120--200~bar (\%)  & at 200~bar (\%H$_2$) \\
\midrule
0.1  & $99.70 \pm 0.12$ & 79.33 & 87.12 & 1.147 \\
0.5  & $99.67 \pm 0.13$ & 85.97 & 91.53 & 0.824 \\
1.0  & $99.63 \pm 0.14$ & 91.01 & 94.69 & 0.522 \\
\textbf{2.0 (selected)} & $\mathbf{99.57 \pm 0.16}$ & $\mathbf{94.02}$ & $\mathbf{96.51}$ & $\mathbf{0.294}$ \\
5.0  & $99.51 \pm 0.19$ & 95.63 & 97.46 & 0.148 \\
10.0 & $99.47 \pm 0.20$ & 96.22 & 97.80 & 0.085 \\
20.0 & $99.46 \pm 0.21$ & 96.55 & 97.99 & 0.048 \\
\bottomrule
\end{tabular*}
\end{table}

\paragraph{Soft-constraint PINN physics-weight selection.}
The soft-constraint PINN
physics-weight schedule $\beta$ was swept over a family of
start--end annealing schedules and evaluated for interpolation and
high-pressure extrapolation accuracy under the two validation protocols described in
Section~\ref{sec:training_validation}. The
selected schedule $0.7\rightarrow0.01$ attains the highest
interpolation accuracy among the tested schedules
($R^2 = 98.86 \pm 1.43\%$), confirming that it is an
interpolation-favourable operating point rather than a deliberately
weakened baseline. Even the most extrapolation-favouring schedule
($0.9\rightarrow0.5$) reaches only $R^2 = 75.70\%$ at 200~bar and
$85.14\%$ across 120--200~bar---well below the PR-Net performance of
($94.02\%$ and $96.51\%$, respectively; Table~\ref{tab:extrapolation_r2}).

\begin{table}[htbp]
\centering
\caption{Soft-constraint PINN physics-weight sensitivity used as a
baseline-fairness check. Interpolation performance is reported using the
stratified 5-fold cross-validation $R^2$ (20 repeats; mean $\pm$ s.d.).
Extrapolation columns report Deep Ensemble mean $R^2$ under
Protocol-IEP at 200~bar and over the pooled 120--200~bar hold-out. The
selected schedule $0.7\rightarrow0.01$ is the soft-constraint PINN
baseline benchmarked in Table~\ref{tab:extrapolation_r2}; $0.9\rightarrow0.5$
is included as an extrapolation-favouring sensitivity bound.}
\label{tab:s_soft_pinn_beta_sensitivity}
\begin{adjustbox}{max width=\textwidth}
\begin{tabular*}{\textwidth}{@{\extracolsep{\fill}}llll@{}}
\toprule
\textbf{Physics-weight} & \textbf{Interp.\ $R^2$ (\%)} & \textbf{Extrap.\ $R^2$ at} & \textbf{Extrap.\ $R^2$ at} \\
\textbf{schedule $\beta$} & \textbf{(5-fold $\times$ 20 CV)} & \textbf{200~bar (\%)} & \textbf{120--200~bar (\%)} \\
\midrule
$0.9 \rightarrow 0.5$  & $98.76 \pm 1.36$ & $75.70$ & $85.14$ \\
$0.9 \rightarrow 0.1$  & $98.83 \pm 1.45$ & $72.17$ & $82.93$ \\
$0.9 \rightarrow 0.01$ & $98.83 \pm 1.50$ & $71.29$ & $82.37$ \\
$0.7 \rightarrow 0.5$  & $98.71 \pm 1.49$ & $73.36$ & $83.73$ \\
$0.7 \rightarrow 0.1$  & $98.84 \pm 1.41$ & $69.00$ & $80.97$ \\
$\mathbf{0.7 \rightarrow 0.01}$ (selected) & $\mathbf{98.86 \pm 1.43}$ & $68.06$ & $80.35$ \\
$0.5 \rightarrow 0.1$  & $98.81 \pm 1.49$ & $65.41$ & $78.65$ \\
$0.5 \rightarrow 0.01$ & $98.80 \pm 1.55$ & $64.21$ & $77.88$ \\
\bottomrule
\end{tabular*}
\end{adjustbox}
\end{table}

\begin{figure}
\centering
\includegraphics[width=0.95\linewidth]{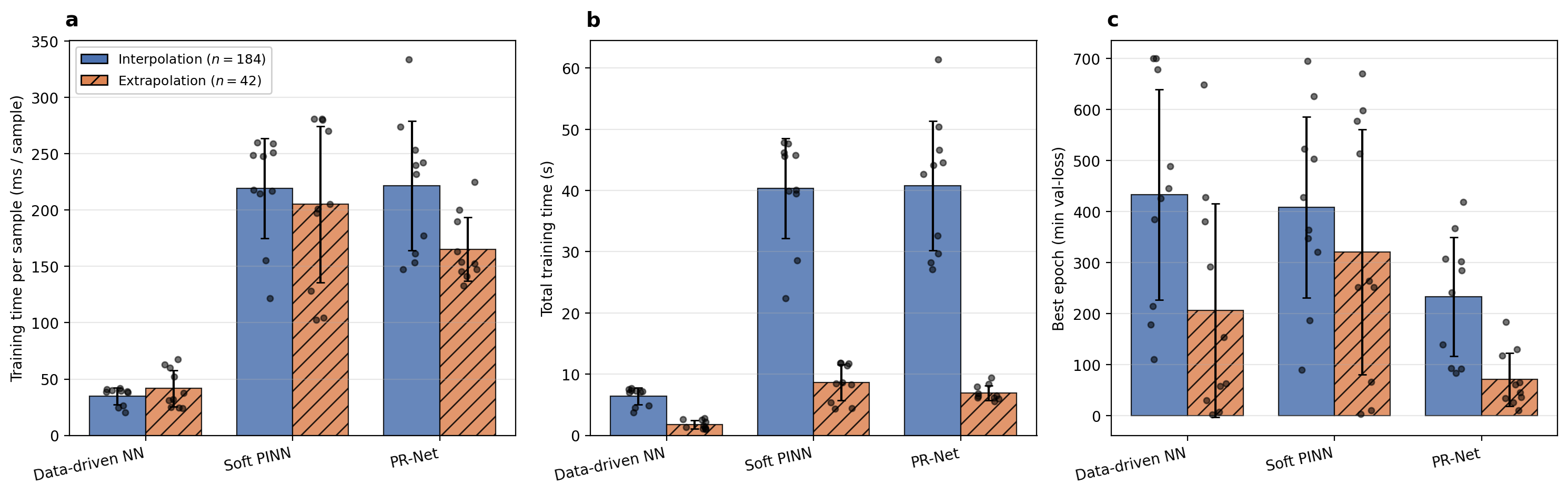}
\caption{Training-cost comparison across architectures in the
interpolation ($n = 184$) and high-pressure extrapolation
($n = 42$) regimes (10 random seeds, CPU; configuration per
Table~\ref{tab:s_hyperparams_search}).
(a)~Training time per sample (total time$\,/\,n$), (b)~total training
time, (c)~best epoch (minimum validation loss). Bars show means;
whiskers one standard deviation; overlaid dots individual
repetitions. Once normalised by dataset size, the per-sample training
cost is comparable across regimes; the physics backbone raises it
several-fold over the bare NN in (b), while PR-Net's hard constraint lowers
the epochs to convergence in (c).}
\label{fig:s_training_cost}
\end{figure}

\paragraph{Training-cost comparison across architectures.}
For an apples-to-apples comparison we trained the three architectures
(data-driven NN, soft-constraint PINN, and PR-Net) following the
shared network architecture of Section~\ref{sec:prnet} and the
training configuration of Table~\ref{tab:s_hyperparams_search} (each
model at its configured learning rate), repeating each experiment with
10 random seeds, in \emph{both} training regimes addressed in this
study: interpolation (the full compiled dataset, $n = 184$) and
high-pressure extrapolation (the $\leq 80$~bar partition, $n = 42$).
Total training wall-clock time and the early-stopping epoch were
measured with \texttt{time.perf\_counter()} on CPU; for these dataset
sizes ($n \leq 184$) the host--device transfer overhead exceeds the
GPU compute saving, so CPU execution is wall-clock-faster than CUDA on
the same hardware. The per-sample training time reported below is the
total time normalised by the dataset size $n$. The extrapolation pipeline additionally requires a
one-time differential-evolution calibration of the per-membrane
backbone coefficients ($\alpha, \beta$)---the IEP fit of
Section~\ref{sec:training_validation}---which the two physics-informed
architectures (PR-Net and the soft-constraint PINN) both rely on and
which therefore precedes their training; this preprocessing step is
timed separately, whereas the interpolation regime reuses the globally
pre-calibrated backbone and incurs no such fit.
Table~\ref{tab:s_training_cost} reports both regimes and
Fig.~\ref{fig:s_training_cost} visualises the per-repetition spread.

\begin{table}[htbp]
\centering
\caption{Training-cost comparison across architectures in the
interpolation ($n = 184$) and high-pressure extrapolation
($n = 42$, $\leq 80$~bar) training regimes, following the network
architecture of Section~\ref{sec:prnet} and the training configuration
of Table~\ref{tab:s_hyperparams_search} (each model at its configured
learning rate). Values are mean $\pm$ standard deviation across 10
random seeds, measured on CPU (24 threads; CUDA disabled, CPU being
faster than GPU at these dataset sizes). Time per sample $=$ total
time$\,/\,n$. The final row is the one-time IEP backbone calibration
shared by the two physics-informed architectures (PR-Net and the
soft-constraint PINN) in the extrapolation pipeline only.}
\label{tab:s_training_cost}
\begin{tabular}{lcccc}
\toprule
\textbf{Model}
  & \textbf{Total $n$}
  & \textbf{Time/sample (ms)}
  & \textbf{Total time (s)}
  & \textbf{Best epoch\textsuperscript{\dag}} \\
\midrule
\multicolumn{5}{l}{\textit{Interpolation regime}} \\
Data-driven NN       & 184 & $34.8 \pm 7.8$   & $6.41 \pm 1.44$  & $433 \pm 217$ \\
Soft-constraint PINN & 184 & $219.4 \pm 46.7$ & $40.37 \pm 8.59$ & $409 \pm 187$ \\
PR-Net               & 184 & $221.5 \pm 60.5$ & $40.75 \pm 11.13$ & $233 \pm 123$ \\
\midrule
\multicolumn{5}{l}{\textit{Extrapolation regime ($\leq 80$~bar)}} \\
Data-driven NN       & 42 & $41.6 \pm 17.2$  & $1.75 \pm 0.72$ & $207 \pm 221$ \\
Soft-constraint PINN & 42 & $205.1 \pm 73.1$ & $8.61 \pm 3.07$ & $321 \pm 253$ \\
PR-Net               & 42 & $165.3 \pm 29.9$ & $6.94 \pm 1.25$ & $71 \pm 55$ \\
\addlinespace
IEP calibration (one-time, shared) & 42 & --- & $24.92 \pm 0.22$ & --- \\
\bottomrule
\end{tabular}

\vspace{4pt}
\noindent\footnotesize
\textsuperscript{\dag}Best epoch $=$ epoch of minimum validation loss
(a convergence indicator, not a time).
\end{table}

Three points follow from these measurements. First, once normalised by
dataset size the per-sample training cost is comparable across the two
regimes for each architecture (data-driven NN $35$ versus
$42$~ms/sample; soft-PINN $219$ versus $205$; PR-Net $221$ versus
$165$), confirming that the large total-time gap chiefly reflects the
difference in $n$ rather than any regime-specific per-update cost; the
physics backbone raises the per-sample cost to roughly $4$--$6\times$
that of the bare NN. Second, PR-Net's hard constraint lowers the
early-stopping epoch---most strongly in extrapolation ($71$ versus
$207$ for the NN and $321$ for the soft-PINN)---so despite the higher
per-sample cost its total training time tracks the soft-constraint PINN
($6.9$ versus $8.6$~s in extrapolation; $40.8$ versus $40.4$~s in
interpolation) and is $4$--$6\times$ that of the bare NN. Third, the
extrapolation pipeline carries a one-time IEP backbone calibration of
$24.9$~s shared by the two physics-informed architectures (PR-Net and
the soft-constraint PINN); in that regime it exceeds either model's own
NN training and is the largest measured cost of preparing the
extrapolation models, but is incurred once and reused across all
subsequent inferences and deployments. The early-stopping epoch shows
high run-to-run variability (relative SD up to ${\sim}100\%$), but the
architecture ordering of cost is preserved across seeds.

\subsection{Online Residual-Update Protocol}
\label{suppl:online_update}
The online-update experiment summarised in
Section~\ref{sec:edge_deployment} uses the held-out Martin et al.\
(2022) dataset~\cite{Martin2022} (digitised from their Fig.~9), which is
not part of the compiled training corpus and probes a novel operating
axis---porous-transport-layer (PTL) compression---absent from the other
sources. It comprises $n = 104$ Nafion~212 measurements
($\approx 51~\mu$m) at a single temperature ($80^{\circ}$C) and anode
pressure ($1$~bar), spanning four PTL compression levels
(10, 35, 60, 85~$\mu$m), three cathode pressures (5, 10, 15~bar), and
nine current densities (0.25--3.5~A~cm$^{-2}$); the measured hydrogen
crossover spans 1.05--9.9\% (Table~\ref{tab:s_martin2022}). With the
calibrated backbone $(\alpha_k, \beta_k)$ frozen, the residual network
is fine-tuned on $k$ newly available measurements at
physics-regularisation weight $\lambda$. For $k = 5$--$60$, the updated
model is evaluated on the held-out remainder; the $k = 104$ column uses
all measurements and is reported only as an apparent in-sample
full-data calibration check (Table~\ref{tab:s_online_update}).

\begin{table}[htbp]
\centering
\caption{Operating-condition coverage of the held-out Martin et al.\
(2022) compression dataset (Nafion~212, $n = 104$)~\cite{Martin2022}
used for the online-update experiment.}
\label{tab:s_martin2022}
\begin{tabular}{ll}
\toprule
\textbf{Variable} & \textbf{Range / values} \\
\midrule
Membrane                 & Nafion~212 ($\approx 51~\mu$m) \\
Temperature              & $80^{\circ}$C \\
Cathode pressure         & 5, 10, 15~bar \\
Anode pressure           & 1~bar \\
PTL compression          & 10, 35, 60, 85~$\mu$m \\
Current density          & 0.25--3.5~A~cm$^{-2}$ (9 levels) \\
H$_2$ crossover (target) & 1.05--9.9\% \\
Samples                  & $n = 104$ \\
\bottomrule
\end{tabular}
\end{table}

\begin{table}[htbp]
\centering
\caption{Online residual update on the Martin et al.~(2022)
compression-varied dataset (Nafion~212, $n = 104$). With the physics
backbone $(\alpha_k, \beta_k)$ frozen, the residual network is
fine-tuned on $k$ new measurements at physics-regularisation weight
$\lambda$. For $k = 5$--$60$, entries are held-out test $R^2$ (\%),
mean over 10 repetitions; the $k = 104$ column uses all available
measurements and is therefore an apparent in-sample full-data
calibration check, not a held-out test estimate. Without any
fine-tuning on the new data, the deployed PR-Net (zero-shot transfer)
gives $R^2 \approx 82.9\%$.}
\label{tab:s_online_update}
\begin{tabular}{lcccccc}
\toprule
 & \multicolumn{6}{c}{\textbf{few-shot size $k$}} \\
\cmidrule(lr){2-7}
\textbf{$\lambda$} & \textbf{5} & \textbf{10} & \textbf{20}
  & \textbf{40} & \textbf{60} & \textbf{104$^\dagger$} \\
\midrule
no update (zero-shot) & $82.9$ & $83.0$ & $82.7$ & $85.0$ & $85.8$ & $83.0$ \\
\addlinespace
$0$     & $89.6$ & $96.2$ & $98.4$ & $98.8$ & $99.2$ & $99.6$ \\
$0.001$ & $89.0$ & $96.3$ & $98.5$ & $98.9$ & $99.2$ & $99.6$ \\
$0.01$  & $89.1$ & $96.2$ & $98.5$ & $98.9$ & $99.2$ & $99.6$ \\
$0.1$   & $89.6$ & $96.4$ & $98.6$ & $98.8$ & $99.1$ & $99.5$ \\
$0.5$   & $90.9$ & $95.8$ & $97.7$ & $97.6$ & $97.7$ & $98.1$ \\
$2.0$   & $89.6$ & $92.1$ & $92.9$ & $92.7$ & $93.0$ & $93.1$ \\
$5.0$   & $87.7$ & $89.0$ & $89.2$ & $88.6$ & $89.3$ & $89.3$ \\
\addlinespace
\textbf{best $\lambda$} & $0.5$ & $0.1$ & $0.1$ & $0.01$ & $0.001$ & $0.001$ \\
\bottomrule
\end{tabular}

\vspace{4pt}
\noindent\footnotesize
The ``no update (zero-shot)'' row is the deployed PR-Net (calibrated
backbone plus its existing residual head) applied with no fine-tuning on
the new data---i.e., zero-shot transfer---evaluated on each held-out
remainder for $k = 5$--$60$, hence its mild $k$-dependence.
The optimal $\lambda$ decreases with $k$: a moderate penalty regularises
the scarce-data updates while a vanishing penalty improves accuracy once
data are ample; for $k \gtrsim 20$ all $\lambda \le 0.1$ fall within
${\sim}0.2$~pp of one another (within run-to-run SD). The penalty also
governs stability---run-to-run SD at $k = 5$ is $2.5\%$ ($\lambda = 2.0$)
versus $6.0\%$ ($\lambda = 0$), and at $k = 60$ is $0.3\%$ ($\lambda = 0$)
versus $3.3\%$ ($\lambda = 5.0$); for $k = 104$, SD~$< 0.1\%$ at all
$\lambda$, as expected for repeated in-sample full-data fitting.
$^\dagger$The $k = 104$ column has no held-out remainder; it uses all
available measurements and is reported as an apparent in-sample
calibration endpoint rather than a held-out test estimate.
\end{table}

\clearpage

\subsection{Uncertainty Quantification Protocol}

Prediction uncertainty was quantified via a Deep Ensemble
approach~\cite{bib49} comprising $M = 100$
independently trained models. Each ensemble member $m$ was
initialised with a sequential random seed
(base seed $42 + m$, $m = 0, 1, \ldots, 99$) to provide
full reproducibility while capturing epistemic uncertainty
across the ensemble.

\paragraph{Ensemble statistics.}
For a given input $\mathbf{x}$, the ensemble mean
(point prediction) and epistemic variance are computed as:

\begin{equation}
\hat{\mu}(\mathbf{x}) = \frac{1}{M} \sum_{m=1}^{M}
  \hat{y}_m(\mathbf{x})
\label{eq:ensemble_mean}
\end{equation}

\begin{equation}
\hat{\sigma}^2(\mathbf{x}) = \frac{1}{M-1} \sum_{m=1}^{M}
  \bigl(\hat{y}_m(\mathbf{x}) - \hat{\mu}(\mathbf{x})\bigr)^2
\label{eq:ensemble_var}
\end{equation}

\noindent where $\hat{y}_m(\mathbf{x})$ denotes the
prediction of the $m$-th ensemble member.

\paragraph{Confidence interval construction.}
Approximate 95\% confidence intervals are constructed
under the Gaussian approximation of the ensemble
predictive distribution:

\begin{equation}
\mathrm{CI}_{95}(\mathbf{x}) =
  \hat{\mu}(\mathbf{x}) \pm 1.96\,\hat{\sigma}(\mathbf{x})
\label{eq:CI}
\end{equation}

\paragraph{Calibration assessment.}
Calibration was evaluated by computing the empirical
coverage probability (ECP)---the fraction of experimental
observations falling within the predicted 95\% confidence
bounds---over the extrapolation test set
($n = 24$; 120--200~bar):

\begin{equation}
\mathrm{ECP} = \frac{1}{n}\sum_{k=1}^{n}
  \mathbf{1}\!\left[
    y^{(k)}_{\mathrm{true}} \in
    \mathrm{CI}_{95}\!\left(\mathbf{x}^{(k)}\right)
  \right]
\label{eq:ECP}
\end{equation}

\noindent A well-calibrated ensemble yields
$\mathrm{ECP} \approx 0.95$. The observed calibration
result for the 100-model PR-Net Deep Ensemble under
Protocol-IEP is reported in
Section~\ref{sec:uncertainty}.

\paragraph{Sensitivity analysis perturbations.}
Operational sensitivity was assessed by independently
perturbing each input variable $x_j$ by its typical
measurement uncertainty $\delta x_j$ and computing
the induced change in $\hat{\mu}$:

\begin{equation}
S_j = \frac{\Delta\hat{\mu}}{\delta x_j}
    = \frac{\hat{\mu}(\mathbf{x} + \delta x_j
      \mathbf{e}_j) -
      \hat{\mu}(\mathbf{x} - \delta x_j
      \mathbf{e}_j)}
      {2\,\delta x_j}
\label{eq:sensitivity}
\end{equation}

\noindent where $\mathbf{e}_j$ is the unit vector along
dimension $j$. Perturbation magnitudes were:
$\delta T = \pm 1\,^{\circ}$C,
$\delta P = \pm 1$~bar,
$\delta i = \pm 0.1$~A~cm$^{-2}$,
$\delta t_{\mathrm{mem}} = \pm 1\,\mu$m.

\clearpage


\printcredits

\section*{Declaration of Competing Interest}
The authors declare that they have no known competing
financial interests or personal relationships that could
have appeared to influence the work reported in this paper.

\section*{Data Availability Statement}
The experimental data analysed in this study were obtained
from the peer-reviewed sources cited herein. The curated
dataset and the model code generated in the present work are
available from the corresponding author upon reasonable request.

\section*{Acknowledgements}
This work was supported by the National Research Foundation of Korea (NRF)
grant funded by the Korea government (MSIT) (No. RS-2024-00405278).
This work was also supported by the National Research Foundation of Korea (NRF)
grant funded by the Korea government (MSIT) (No. RS-2026-25470261).

\bibliographystyle{model1-num-names}

\bibliography{cas-refs}


\end{document}